\pdfoutput=1

\PassOptionsToPackage{table}{xcolor}
\documentclass[11pt,table]{article}

\usepackage[utf8]{inputenc}
\usepackage{arabtex}
\usepackage{utf8}
\setcode{utf8}

\usepackage[preprint]{acl}

\usepackage{times}
\usepackage{latexsym}
\usepackage[table]{xcolor}  
\usepackage[T1]{fontenc}

\usepackage{microtype}

\usepackage{inconsolata}

\usepackage{graphicx}
\usepackage{subcaption} 
\usepackage{booktabs}
\usepackage{tikz}
\usepackage{enumitem}
\usetikzlibrary{tikzmark, arrows.meta}
\usetikzlibrary{positioning}
\usepackage{amsmath}
\usepackage{cleveref}
\usepackage{hyperref}

\usetikzlibrary{calc,arrows.meta}

\usepackage{upquote}   
\usepackage{textcomp} 
\usepackage{tcolorbox}
\tcbuselibrary{skins}
\tcbuselibrary{breakable}
\tcbuselibrary{raster}
\tcbuselibrary{hooks}
\tcbuselibrary{theorems}
\tcbuselibrary{listings}
\tcbuselibrary{documentation}
\usepackage{enumitem}
\definecolor{blue}{HTML}{5989cf}
\definecolor{sub}{HTML}{cde4ff}
\newtcolorbox{todobox}{
    colback = sub, 
    colframe = blue, 
    boxrule = 0pt, 
    leftrule = 6pt 
}
\tcbset{
  findingbox/.style={
    colback=gray!5,             
    colframe=black!30,          
    coltitle=white,             
    colbacktitle=black!60,      
    fonttitle=\small\bfseries,      
    boxrule=0.3pt,              
    arc=2pt,                    
    left=6pt, right=6pt, top=2pt, bottom=2pt, 
    enhanced,
    boxed title style={size=small, colframe=black!60, colback=black!60, sharp corners,
    top=1pt, bottom=1pt, left=2pt, right=2pt
    },
  }
}

\usepackage{CJKutf8}
\usepackage{anyfontsize}

\usepackage{lipsum}

\definecolor{headerbg}{RGB}{0, 121, 124} 
\definecolor{bodybg}{RGB}{217, 237, 241} 

\definecolor{myBlue}{HTML}{5989cf}

\definecolor{myPink}{HTML}{f568be}
\newcommand{\pink}{\textcolor{myPink}}
\definecolor{myPurple}{HTML}{b68af3}
\newcommand{\purple}{\textcolor{myPurple}}
\definecolor{myFigureBlue}{HTML}{639df3}
\newcommand{\figureblue}{\textcolor{myFigureBlue}}

\usepackage{array}
\usepackage{multirow}
\usepackage{booktabs}
\usepackage{svg}
\definecolor{darkred}{rgb}{1.0,0.4,0.4}
\definecolor{mediumred}{rgb}{1.0,0.6,0.6}
\definecolor{lightred}{rgb}{1.0,0.8,0.8}

\definecolor{white}{rgb}{1.0,1.0,1.0}
\definecolor{lightgreen}{rgb}{0.8,1.0,0.8}
\definecolor{mediumgreen}{rgb}{0.6,1.0,0.6}
\definecolor{darkgreen}{rgb}{0.4,1.0,0.4}

\newcommand{\cellequal}[1]{\cellcolor{white}#1}
\newcommand{\celldash}{-}

\newcommand{\celllightred}[1]{\cellcolor{lightred}#1}
\newcommand{\cellmediumred}[1]{\cellcolor{mediumred}#1}
\newcommand{\celldarkred}[1]{\cellcolor{darkred}#1}

\newcommand{\celllightgreen}[1]{\cellcolor{lightgreen}#1}
\newcommand{\cellmediumgreen}[1]{\cellcolor{mediumgreen}#1}
\newcommand{\celldarkgreen}[1]{\cellcolor{darkgreen}#1}

\newcommand{\gobbledot}{}
\def\gobbledot#1.#2\relax{#1#2}
\title{Language Mixing in Reasoning Language Models:\\Patterns, Impact, and Internal Causes}

\author{
Mingyang Wang$^{1,2,3}$ \hspace*{0.2cm}
{Lukas Lange$^{1}$} \hspace*{0.2cm} {Heike Adel$^{4}$} 
 \\ 
{\bf Yunpu Ma$^{2,3}$} \hspace*{0.2cm} 
 {\bf Jannik Str\"{o}tgen$^{5}$ \hspace*{0.2cm} Hinrich Sch\"{u}tze$^{2,3}$} \\
  $^1$Bosch Center for Artificial Intelligence, Renningen, Germany \\
  $^2$LMU Munich, Germany \hspace*{0.2cm}
  $^3$Munich Center for Machine Learning (MCML) \\
  $^4$Hochschule der Medien, Stuttgart, Germany \\
  $^5$Karlsruhe University of Applied Sciences, Germany \\
  \texttt{mingyang@cis.lmu.de} 
  }

\newcounter{notecounter}
\newcommand{\enotesoff}{\long\gdef\enote##1##2{}}

\enotesoff

\begin{document}
\maketitle

\begin{abstract}
Reasoning language models (RLMs) excel at complex tasks by leveraging a chain-of-thought process to generate structured intermediate steps. However, \textit{language mixing}, i.e., reasoning steps containing tokens from languages other than the prompt, has been observed in their outputs and shown to affect performance, though its impact remains debated. 
We present the first systematic study of language mixing in RLMs, examining its patterns, impact, and internal causes across 15 languages, 7 task difficulty levels, and 18 subject areas, and show how all three factors influence language mixing. 
Moreover, we demonstrate that the choice of reasoning language significantly affects performance: forcing models to reason in Latin or Han scripts via constrained decoding notably improves accuracy.
Finally, we show that the script composition of reasoning traces closely aligns with that of the model's internal representations, indicating that language mixing reflects latent processing preferences in RLMs.
Our findings provide actionable insights for optimizing multilingual reasoning and open new directions for controlling reasoning languages to build more interpretable and adaptable RLMs.\footnote{We make our \href{https://github.com/cisnlp/Language-Mixing}{data} and \href{https://github.com/boschresearch/Language-Mixing}{code} publicly available.}
\end{abstract}



\section{Introduction}
Reasoning language model (RLMs)\footnote{The terms ``Reasoning language models'' (RLMs) and ``Large reasoning models'' (LRMs) are both used in prior works \citep[e.g.,][]{xu2025towards, chen2025towards}. In this paper, we adopt the term \textit{Reasoning language models}.}, such as Open\-AI’s o1 and o3 \citep{jaech2024openai, openai2024o3o4}, and the DeepSeek-R1 series \citep{guo2025deepseek}, have demonstrated impressive capabilities in solving complex tasks through structured chain-of-thought reasoning. These models generate intermediate reasoning steps before answering the input prompt. This not only improves task performance, but also enhances the transparency and interpretability of their decision-making processes.
However, the phenomenon of language mixing has emerged: when prompted in one language, RLMs have been observed to produce reasoning steps that include a mixture of languages \citep{qwq-32b-preview, guo2025deepseek}, as illustrated in Figure~\ref{fig:teaser}.
This phenomenon has been shown to affect reasoning performance, though prior work offers conflicting views on whether its impact is beneficial or detrimental \citep{guo2025deepseek, xie2025logicrlunleashingllmreasoning}. Moreover, it may hinder the readability and usability of outputs in multilingual contexts. 

\definecolor{greenbox}{HTML}{d6eadf}
\definecolor{greenborder}{HTML}{b8e0d4}
\definecolor{bluebox}{HTML}{d7e3fc}
\definecolor{blueborder}{HTML}{C5D3E8}
\definecolor{grayborder}{HTML}{B7B7B7}
\definecolor{redtext}{HTML}{BE5B50}

\definecolor{greentext}{HTML}{537D5D}
\begin{figure}[!t]
\begin{tcolorbox}[
    enhanced,
    colback=bluebox,
    colframe=blueborder,
    arc=6pt,
    boxrule=2pt,
    top=2pt,
    bottom=2pt,
    before upper={\linespread{1.1}\selectfont},
    fontupper=\small,
    overlay={
        \node[anchor=north east, xshift=0pt, yshift=24pt] at (frame.north east) {
            \includegraphics[height=1cm]{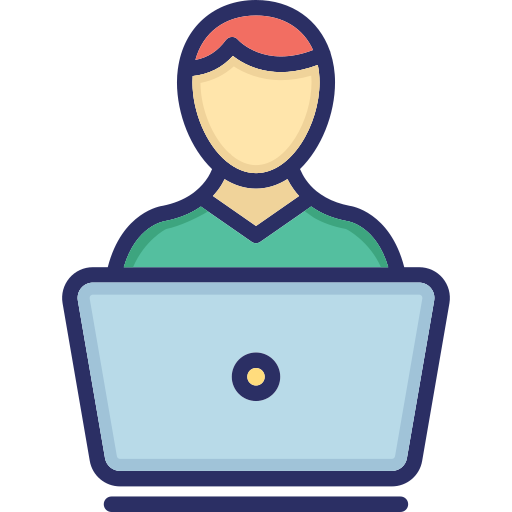} 
        };
    }
]
 {\fontsize{8.5pt}{9pt}\selectfont \textcolor{blue}{\textbf{{Question:}}} How many of the first one hundred\\positive integers are divisible by 3, 4, and 5?}
\end{tcolorbox}

\begin{tcolorbox}[
    enhanced,
    colback=gray!5,
    colframe=grayborder,
    arc=6pt,
    boxrule=2pt,
    before upper={\linespread{1.1}\selectfont},
    fontupper=\small, 
    overlay={
        \node[anchor=north east, xshift=32pt, yshift=13pt] at (frame.north west) {
            \includegraphics[height=0.75cm]{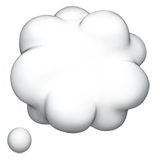}
        };
    }
]
{\fontsize{8.5pt}{9pt} \textcolor{darkgray}{\textbf{Thinking:}} Hmm, let me think about the question. Well, if a number is divisible by multiple numbers, it's called a common multiple.

\vspace{5pt}

So maybe I should start by finding the least common multiple (LCM) of 3, 4, and 5…

\vspace{5pt}

So, the LCM of 3, 4, and 5 is 60.

\vspace{5pt}

\textcolor{redtext}{\begin{CJK}{UTF8}{gbsn}\footnotesize接下来，我需要找出在1到100之间有多少个数是60的倍数。也就是说,我可以列举可能的倍数。比如下一个是 $60 \times 2 = 120$，超过了100 \end{CJK}}\textcolor{redtext}{...}

\vspace{5pt}

So yes, only 60. Therefore, the answer is 1.}
\end{tcolorbox}
\begin{tcolorbox}[
    enhanced,
    colback=greenbox,
    colframe=greenborder,
    arc=6pt,
    boxrule=2pt,
    top=2pt,
    bottom=2pt,
    before upper={\linespread{1.1}\selectfont},
    fontupper=\small,
    overlay={
        \node[anchor=north east, xshift=32pt, yshift=15pt] at (frame.north west) {
            \includegraphics[height=0.6cm]{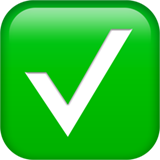} 
        };
        }
]
{\fontsize{8.5pt}{9pt} \textcolor{greentext}{\textbf{Answer}}: There is only one number, \textbf{60}, that is divisible by 3, 4, and 5 in the first 100 positive integers.}
\end{tcolorbox}
  \caption{An illustrative example of language mixing in reasoning where the reasoning model switches from English to Chinese mid-reasoning, then back to English.\protect\footnotemark}

  \label{fig:teaser}
\end{figure}

\footnotetext{The Chinese text translates to: "Next, I need to find how many numbers between 1 and 100 are multiples of 60. That is to say, I can list the possible multiples. For example, the next one is 60 × 2 = 120, which exceeds 100..."}

To shed light on the phenomenon of language mixing in state-of-the-art RLMs, we present a systematic investigation aimed at answering three key questions: 
\textbf{(1) When does language mixing happen} and what factors influence its occurrence and degree?
\\ \noindent \textbf{(2) Is language mixing an issue} and what impact does it have on the model's reasoning performance? 
\\ \noindent \textbf{(3) Why does language mixing happen} and how is it related to the model's internal thinking process?

First, we analyze occurrence patterns of language mixing across 15 input languages, 7 task difficulty levels, and 18 subject areas. We observe that language mixing is most prevalent when the input language is neither English nor Chinese, suggesting that English, and to a lesser extent, Chinese, serve as internal pivot languages during reasoning.
Moreover, we find that the degree of language mixing, measured by the entropy of language distribution in reasoning traces, increases with task difficulty across models and languages. 
Furthermore, subject-wise analysis on the multilingual MMLU dataset \citep{hendrycks2020measuring} reveals that language mixing entropy is significantly higher in STEM subjects compared to  other domains.

Second, although language mixing has been observed in prior works, its impact on model performance remains unclear and is under debate \citep{qwq-32b-preview, guo2025deepseek, xie2025logicrlunleashingllmreasoning}. While switching languages mid-reasoning may reduce readability, it may be undesirable to suppress it entirely if such mixing enhances the model's reasoning ability.
To assess its impact, we introduce a script control method that constrains models to reason in specific script(s) via constrained decoding.
We find that constraining reasoning to Latin or Han (Chinese) scripts significantly improves reasoning performance, by up to 110\% in some cases, indicating that the choice of reasoning script or language has a substantial impact on model performance.

Finally, to understand the underlying cause of language
mixing, we analyze the internal processing of RLMs using
mechanistic interpretability. Applying Logit
Lens \citep{nostal2022logitlens} at the script level, we
find that the script composition of hidden representations
closely mirrors that of reasoning traces, revealing that
language mixing reflects the model’s latent processing
preferences, particularly a consistent bias towards Latin.

In summary, we present the first systematic analysis of language mixing in RLMs, analyzing its occurrence patterns, performance impact, and internal causes. Our analysis offers practical guidance for improving multilingual reasoning and opens up new opportunities to control and adapt the reasoning language, supporting the development of more robust, interpretable, and user-aligned RLMs.






\section{Related Work}

\paragraph{Reasoning Language Models.}
\label{sec:related-work-rlms}
Reasoning language models, such as OpenAI’s o1/o3 \citep{jaech2024openai, openai2024o3o4} and the DeepSeek-R1 series \citep{guo2025deepseek}, have demonstrated strong capabilities on complex tasks by generating structured intermediate steps \citep{chen2025towards}.
While prior work has largely focused on improving reasoning quality via prompting or training, little attention has been paid to RLM behavior in multilingual settings. One underexplored aspect is language mixing, where the language used in reasoning steps differs from that of the input prompt. DeepSeek-R1 introduces a language consistency reward to reduce such mixing \citep{guo2025deepseek}, but reports a drop in performance, suggesting potential benefits of language mixing. In contrast, \citet{xie2025logicrlunleashingllmreasoning} show that responses without language mixing yield higher accuracy in logical reasoning tasks.

Given these contradictory findings, our work aims at a first systematic study of language mixing in RLMs, analyzing its patterns, performance impact, and internal causes in multilingual contexts.

\paragraph{Inner Workings of Multilingual LLMs.}
A growing body of work has explored how multilingual models internally represent and process information across languages. Works by \citet{wendler-etal-2024-llamas}, \citet{dumas2024llamas}, \citet{wang-etal-2025-lost-multilinguality}, and \citet{liu2025tracing} reveal that models like Llama tend to rely on English representations internally, even when operating in other languages, highlighting a strong bias toward high-resource languages. While these works focus on internal activations using translation or knowledge probing tasks, our analysis links internal processing to external reasoning traces, showing that language mixing aligns with the model’s latent language preferences during reasoning.

\paragraph{Code-Switching in Language Models.}
Code-switching, the natural alternation between languages within text, has been widely studied in NLP tasks like sentiment analysis, machine translation, and summarization \citep{dogruoz-etal-2021-survey}. 

Prior work shows that current language models still struggle to understand and generate code-switched text, particularly in low-resource settings \citep{khanuja-etal-2020-gluecos, winata-etal-2023-decades, yong-etal-2023-prompting, zhang-etal-2023-multilingual,li2024distinct, li2024towards}.
Recent studies have highlighted unnatural language confusion in multilingual models as another type of code-switching \citep{marchisio-etal-2024-understanding, nie2025mechanistic}, such as source-language hallucination or off-target translation, especially in English-centric models under zero-shot conditions.

In this work, we investigate language mixing, a code-switching–like behavior that arises during intermediate reasoning steps in RLMs. 
We examine its occurrence patterns across languages,
difficulty levels, and subject domains, assess its impact on performance, and link it to models' internal representations, offering new insights into the reasoning behavior of RLMs.

\begin{figure}[t]
\centering
\fontsize{9.5pt}{11.5pt}\selectfont
\begin{tcolorbox}[colback=blue!5!white, colframe=blue!75!black, title=Example of a Knights-and-Knaves puzzle, width=\linewidth, top=3pt, bottom=3pt]
\textbf{Problem:} A very special island is inhabited only by knights and knaves. Knights always tell the truth, and knaves always lie. You meet 2 inhabitants: Zoey, and Oliver. Zoey remarked, "Oliver is not a knight". Oliver stated, "Oliver is a knight if and only if Zoey is a knave". So who is a knight and who is a knave?

\vspace{0.2cm}
\textbf{Solution:} (1) Zoey is a knave (2) Oliver is a knight
\end{tcolorbox}
\caption{An illustrative example from the K\&K dataset.}
\label{fig:kk-example}
\end{figure}

\section{Experimental Setup}
\label{sec:exp-setup}
\paragraph{Models.}
\label{sec:exp-setup-models}
We evaluate a broad range of reasoning language models, including \texttt{DeepSeek-R1} \citep{guo2025deepseek} and its distilled variants, \texttt{DeepSeek-R1-Distill-Qwen-\{1.5B, 7B, 14B, 32B\}} and \texttt{DeepSeek-R1-Distill-Llama-\{8B, 70B\}}. We also include \texttt{QwQ-32B} \citep{qwq-32b}, \texttt{Qwen3-\{4B, 30B-A3B, 32B\}} \citep{qwen3}, and \texttt{Gemini 2.0 Flash Thinking} \citep{gemini} models to broaden coverage. For comparing language mixing between RLMs and their backbones, we include \texttt{Qwen2.5-\{14B, 32B\}} and \texttt{Llama3.3-70B-Instruct}, to compare \texttt{DeepSeek-R1-Distill-Qwen-\{14B, 32B\}} and \texttt{DeepSeek-R1-Distill-Llama-70B} with their distillation backbones.
Table~\ref{tab:model-list} in Appendix~\ref{sec:appendix-model} summarizes all models used in our evaluation. We follow the official hyperparameter settings for each model (see Table~\ref{tab:model-hp} in Appendix~\ref{sec:appendix-model-hp}). Some reasoning models occasionally exhibit endless reasoning --- continue to generate reasoning steps without reaching a final answer.\footnote{This overthinking behavior is also observed in prior work such as \citet{cuadron2025danger}.} As this occurs only in specific cases for most models and does not reflect a consistent pattern, our analysis focuses on valid reasoning traces that conclude with a final answer.\footnote{See Appendix~\ref{sec:appendix-model-performance} for validity statistics and accuracies.}

\paragraph{Datasets.}
\label{sec:exp-setup-dataset}
The \textbf{Knights-and-Knaves} (K\&K) dataset \citep{xie2024memorization} contains logical reasoning puzzles where each character is either a \textit{knight}, who always tells the truth, or a \textit{knave}, who always lies. The goal is to infer each character’s identity based on their statements (see example in Figure~\ref{fig:kk-example}). Difficulty is controlled by varying the number of characters (2–8). The original dataset is in English. To enable multilingual evaluation, we translate it into five additional languages: Arabic, French, Hindi, Japanese, and Chinese using \texttt{gpt-4o-mini}, resulting in six languages and seven difficulty levels. 

\begin{table}[t]
\footnotesize
  \centering
    \begin{tabular}{p{1.8cm}p{4.5cm}} 
    \midrule
    \textbf{Supercategory} & \textbf{Subject} \\
    \midrule
    Humanities & HS World History, Moral Disputes, Philosophy, World Religions \\
    \midrule
    Social Science & HS Macroeconomics, Sociology \\
    \midrule
    STEM (Science, Technology, Engineering, Mathematics) & HS Computer Science, Col Computer Science, Elem Mathematics, HS Mathematics, Col Mathematics, HS Chemistry, Col Chemistry, HS Physics, Col Physics \\
    \midrule
    Other & Global Facts, Management, Professional Medicine \\
    \midrule
    \end{tabular}%
  \caption{Overview of the 18 subjects in m-MMLU included in our evaluation. Abbreviations: HS = High School, Col = College, Elem = Elementary.}
  \label{tab:m-mmlu-stats}%
\end{table}%


To evaluate language mixing across broader domains, we also use the \textbf{multilingual MMLU} (m-MMLU) dataset \citep{hendrycks2020measuring}, a large-scale benchmark of multiple-choice questions covering 15 languages and 57 subjects across Humanities, Social Sciences, STEM, and Other domains. We select 18 representative subjects for RLM evaluation, as summarized in Table~\ref{tab:m-mmlu-stats}.

Further details on the datasets, including the translation process and language coverage, are provided in Appendix~\ref{sec:appendix-dataset}.

\section{Language Mixing Patterns}
\label{sec:pattern-res}

\begin{figure*}[h]
    \centering
    \includegraphics[width=0.96\linewidth]{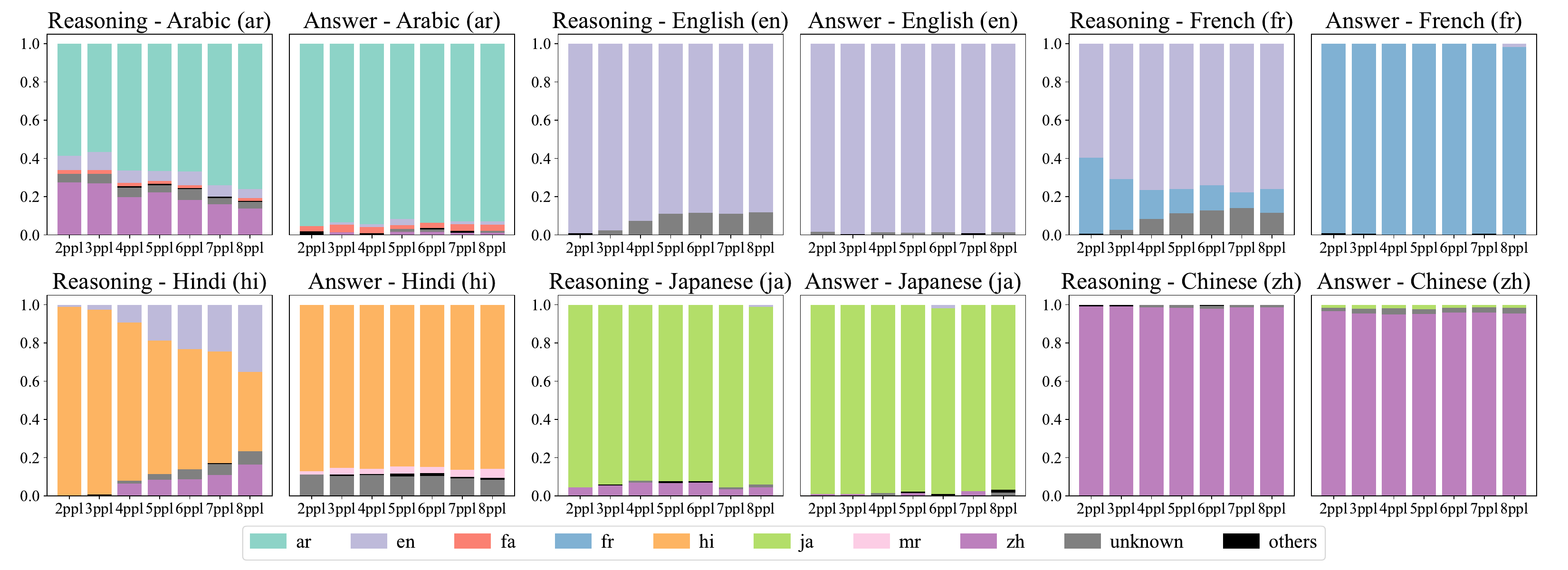}
    \caption{Language composition across difficulty levels in the K\&K dataset for \texttt{DeepSeek-R1-Distill-Llama-70B}.}
    \label{fig:showcase-pattern-DS-70B}
\end{figure*}

\begin{figure*}[h]
    \centering
    \includegraphics[width=0.96\linewidth]{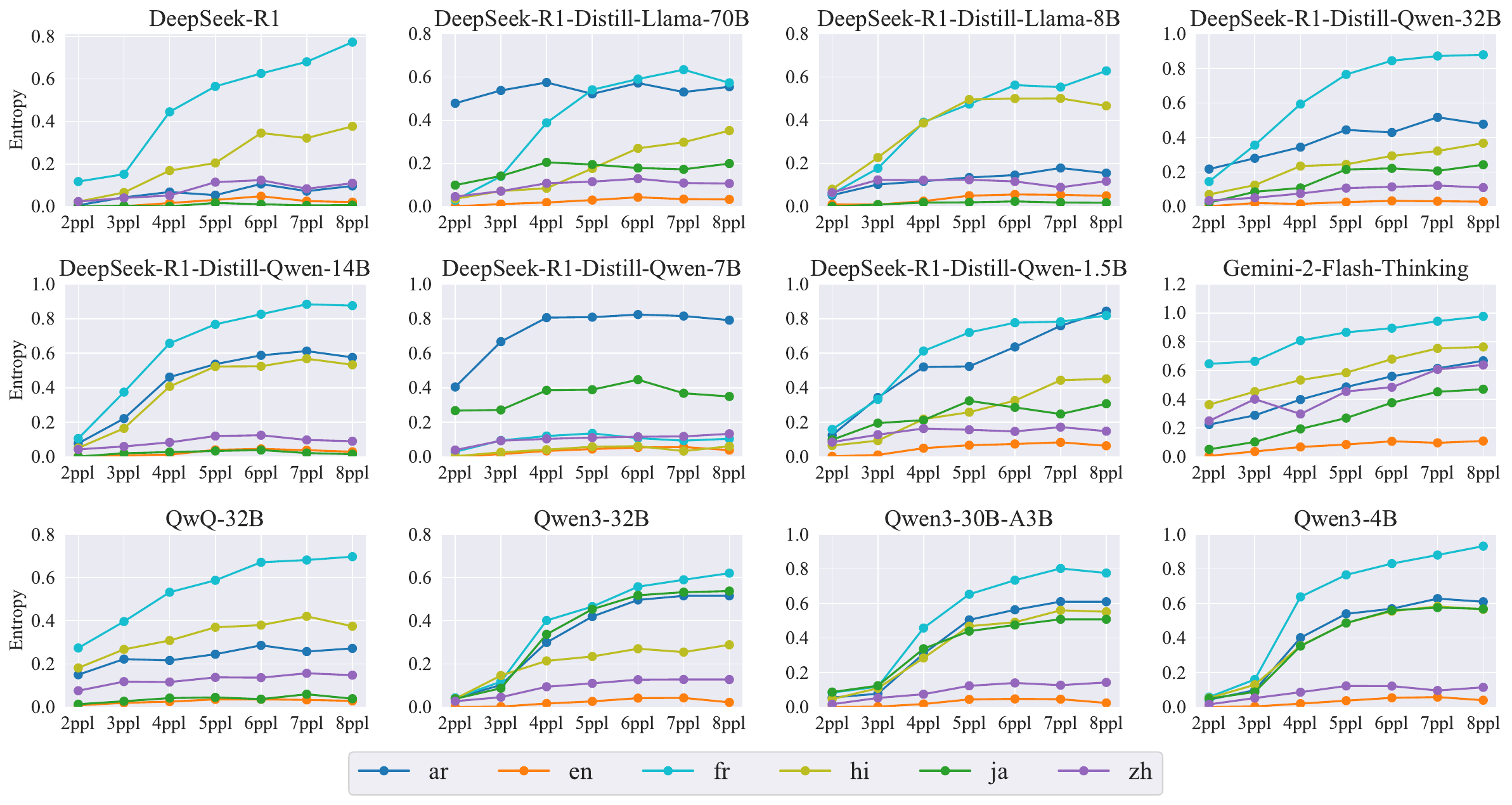}
    \caption{Language mixing entropy across task difficulty levels for six input languages and twelve models. Entropy mostly increases with difficulty, indicating harder tasks generally induce more language mixing in reasoning.}
    \label{fig:kk-entropy}
\end{figure*}


\subsection{Method}
\label{sec:pattern-eval-intro}
To investigate the occurrence patterns of language mixing in RLMs, we first collect the reasoning traces and final answers generated by each model for queries in the K\&K and m-MMLU datasets. Following \citet{marchisio-etal-2024-understanding}, we perform line-level language detection by splitting each output on newline characters and identifying the language of each line using fastText \citep{joulin-etal-2017-bag}. This yields a per-line language distribution for each reasoning trace and answer.\footnote{We consider results with confidence scores below 0.5 as \textit{unknown}. They are usually lines mixing multiple languages or are mostly symbols.}

We compute the language composition of each sample by
aggregating the detected languages across lines. Averaging
across all samples in a dataset yields an overall language
usage distribution, e.g., $\{\text{``en'': }
0.5, \text{``fr'': } 0.2, \text{``zh'': } 0.3\}$. To
quantify the degree of language mixing, we calculate the
entropy of the distribution, ($[0.5, 0.2, 0.3]$ in the example), where higher entropy indicates a higher degree of language mixing, and lower entropy reflects greater language consistency.

\begin{table*}[h]
\centering
\footnotesize
\scalebox{0.84}{
\begin{tikzpicture}
\node (table) {
\begin{tabular}{lcccccccccc}
\toprule
\textbf{m-MMLU Subject} & \textbf{\texttt{R1-70B}} & \textbf{\texttt{R1-32B}} & \textbf{\texttt{R1-14B}} & \textbf{\texttt{R1-8B}} & \textbf{\texttt{R1-7B}} & \textbf{\texttt{Gemini}} & \textbf{\texttt{QwQ-32B}} & \textbf{\texttt{Q3-32B}} & \textbf{\texttt{Q3-30B-A3B}} & \textbf{\texttt{Q3-4B}} \\
\midrule
    Elementary Mathematics & 0.11  & 0.09  & 0.08  & 0.12  & 0.12  & 0.51  & 0.09  & 0.04  & 0.08  & 0.09 \\
    High School Mathematics & 0.32  & 0.31  & 0.28  & 0.31  & 0.32  & 0.74  & 0.34  & 0.27  & 0.32  & 0.31 \\
    College Mathematics & 0.39  & 0.35  & 0.35  & 0.39  & 0.39  & 0.84  & 0.35  & 0.35  & 0.38  & 0.40 \\
    \midrule
    High School Chemistry & 0.20  & 0.17  & 0.18  & 0.22  & 0.21  & 0.55  & 0.20  & 0.11  & 0.17  & 0.19 \\
    College Chemistry & 0.27  & 0.28  & 0.29  & 0.25  & 0.21  & 0.66  & 0.44  & 0.20  & 0.25  & 0.22 \\
    \midrule
    High School Physics & 0.27  & 0.22  & 0.22  & 0.24  & 0.26  & 0.74  & 0.26  & 0.15  & 0.21  & 0.24 \\
    College Physics & 0.32  & 0.27  & 0.28  & 0.26  & 0.30  & 0.84  & 0.33  & 0.20  & 0.27  & 0.28 \\
    \midrule
    High School Computer Science & 0.15  & 0.14  & 0.14  & 0.17  & 0.19  & 0.47  & 0.14  & 0.10  & 0.15  & 0.15 \\
    College Computer Science & 0.20  & 0.19  & 0.22  & 0.22  & 0.23  & 0.54  & 0.21  & 0.13  & 0.19  & 0.19 \\

\bottomrule
\end{tabular}
};

\draw[-{Stealth[length=2mm,width=2mm]},gray,line width=0.4pt] ($(table.north)+(-3.6,-0.85)$) -- ($(table.north)+(-3.6,-1.8)$);
\draw[-{Stealth[length=2mm,width=2mm]},gray,line width=0.4pt] ($(table.north)+(-2.25,-0.85)$) -- ($(table.north)+(-2.25,-1.8)$);
\draw[-{Stealth[length=2mm,width=2mm]},gray,line width=0.4pt] ($(table.north)+(-0.9,-0.85)$) -- ($(table.north)+(-0.9,-1.8)$);
\draw[-{Stealth[length=2mm,width=2mm]},gray,line width=0.4pt] ($(table.north)+(0.4,-0.85)$) -- ($(table.north)+(0.4,-1.8)$);
\draw[-{Stealth[length=2mm,width=2mm]},gray,line width=0.4pt] ($(table.north)+(1.6,-0.85)$) -- ($(table.north)+(1.6,-1.8)$);
\draw[-{Stealth[length=2mm,width=2mm]},gray,line width=0.4pt] ($(table.north)+(2.9,-0.85)$) -- ($(table.north)+(2.9,-1.8)$);
\draw[-{Stealth[length=2mm,width=2mm]},gray,line width=0.4pt] ($(table.north)+(4.35,-0.85)$) -- ($(table.north)+(4.35,-1.8)$);
\draw[-{Stealth[length=2mm,width=2mm]},gray,line width=0.4pt] ($(table.north)+(5.8,-0.85)$) -- ($(table.north)+(5.8,-1.8)$);
\draw[-{Stealth[length=2mm,width=2mm]},gray,line width=0.4pt] ($(table.north)+(7.5,-0.85)$) -- ($(table.north)+(7.5,-1.8)$);
\draw[-{Stealth[length=2mm,width=2mm]},gray,line width=0.4pt] ($(table.north)+(9.1,-0.85)$) -- ($(table.north)+(9.1,-1.8)$);

\draw[-{Stealth[length=2mm,width=2mm]},gray,line width=0.4pt] ($(table.north)+(-3.6,-2.1)$) -- ($(table.north)+(-3.6,-2.7)$);
\draw[-{Stealth[length=2mm,width=2mm]},gray,line width=0.4pt] ($(table.north)+(-2.25,-2.1)$) -- ($(table.north)+(-2.25,-2.7)$);
\draw[-{Stealth[length=2mm,width=2mm]},gray,line width=0.4pt] ($(table.north)+(-0.9,-2.1)$) -- ($(table.north)+(-0.9,-2.7)$);
\draw[-{Stealth[length=2mm,width=2mm]},gray,line width=0.4pt] ($(table.north)+(0.4,-2.1)$) -- ($(table.north)+(0.4,-2.7)$);
\draw[dashed,gray,line width=0.2pt] ($(table.north)+(1.6,-2.21)$) -- ($(table.north)+(1.8,-2.21)$);
\draw[dashed,gray,line width=0.2pt] ($(table.north)+(1.6,-2.56)$) -- ($(table.north)+(1.8,-2.56)$);
\draw[-{Stealth[length=2mm,width=2mm]},gray,line width=0.4pt] ($(table.north)+(2.9,-2.1)$) -- ($(table.north)+(2.9,-2.7)$);
\draw[-{Stealth[length=2mm,width=2mm]},gray,line width=0.4pt] ($(table.north)+(4.35,-2.1)$) -- ($(table.north)+(4.35,-2.7)$);
\draw[-{Stealth[length=2mm,width=2mm]},gray,line width=0.4pt] ($(table.north)+(5.8,-2.1)$) -- ($(table.north)+(5.8,-2.7)$);
\draw[-{Stealth[length=2mm,width=2mm]},gray,line width=0.4pt] ($(table.north)+(7.5,-2.1)$) -- ($(table.north)+(7.5,-2.7)$);
\draw[-{Stealth[length=2mm,width=2mm]},gray,line width=0.4pt] ($(table.north)+(9.1,-2.1)$) -- ($(table.north)+(9.1,-2.7)$);

\draw[-{Stealth[length=2mm,width=2mm]},gray,line width=0.4pt] ($(table.north)+(-3.6,-3)$) -- ($(table.north)+(-3.6,-3.6)$);
\draw[-{Stealth[length=2mm,width=2mm]},gray,line width=0.4pt] ($(table.north)+(-2.25,-3)$) -- ($(table.north)+(-2.25,-3.6)$);
\draw[-{Stealth[length=2mm,width=2mm]},gray,line width=0.4pt] ($(table.north)+(-0.9,-3)$) -- ($(table.north)+(-0.9,-3.6)$);
\draw[-{Stealth[length=2mm,width=2mm]},gray,line width=0.4pt] ($(table.north)+(0.4,-3)$) -- ($(table.north)+(0.4,-3.6)$);
\draw[-{Stealth[length=2mm,width=2mm]},gray,line width=0.4pt] ($(table.north)+(1.6,-3)$) -- ($(table.north)+(1.6,-3.6)$);
\draw[-{Stealth[length=2mm,width=2mm]},gray,line width=0.4pt] ($(table.north)+(2.9,-3)$) -- ($(table.north)+(2.9,-3.6)$);
\draw[-{Stealth[length=2mm,width=2mm]},gray,line width=0.4pt] ($(table.north)+(4.35,-3)$) -- ($(table.north)+(4.35,-3.6)$);
\draw[-{Stealth[length=2mm,width=2mm]},gray,line width=0.4pt] ($(table.north)+(5.8,-3)$) -- ($(table.north)+(5.8,-3.6)$);
\draw[-{Stealth[length=2mm,width=2mm]},gray,line width=0.4pt] ($(table.north)+(7.5,-3)$) -- ($(table.north)+(7.5,-3.6)$);
\draw[-{Stealth[length=2mm,width=2mm]},gray,line width=0.4pt] ($(table.north)+(9.1,-3)$) -- ($(table.north)+(9.1,-3.6)$);

\draw[-{Stealth[length=2mm,width=2mm]},gray,line width=0.4pt] ($(table.north)+(-3.6,-3.9)$) -- ($(table.north)+(-3.6,-4.5)$);
\draw[-{Stealth[length=2mm,width=2mm]},gray,line width=0.4pt] ($(table.north)+(-2.25,-3.9)$) -- ($(table.north)+(-2.25,-4.5)$);
\draw[-{Stealth[length=2mm,width=2mm]},gray,line width=0.4pt] ($(table.north)+(-0.9,-3.9)$) -- ($(table.north)+(-0.9,-4.5)$);
\draw[-{Stealth[length=2mm,width=2mm]},gray,line width=0.4pt] ($(table.north)+(0.4,-3.9)$) -- ($(table.north)+(0.4,-4.5)$);
\draw[-{Stealth[length=2mm,width=2mm]},gray,line width=0.4pt] ($(table.north)+(1.6,-3.9)$) -- ($(table.north)+(1.6,-4.5)$);
\draw[-{Stealth[length=2mm,width=2mm]},gray,line width=0.4pt] ($(table.north)+(2.9,-3.9)$) -- ($(table.north)+(2.9,-4.5)$);
\draw[-{Stealth[length=2mm,width=2mm]},gray,line width=0.4pt] ($(table.north)+(4.35,-3.9)$) -- ($(table.north)+(4.35,-4.5)$);
\draw[-{Stealth[length=2mm,width=2mm]},gray,line width=0.4pt] ($(table.north)+(5.8,-3.9)$) -- ($(table.north)+(5.8,-4.5)$);
\draw[-{Stealth[length=2mm,width=2mm]},gray,line width=0.4pt] ($(table.north)+(7.5,-3.9)$) -- ($(table.north)+(7.5,-4.5)$);
\draw[-{Stealth[length=2mm,width=2mm]},gray,line width=0.4pt] ($(table.north)+(9.1,-3.9)$) -- ($(table.north)+(9.1,-4.5)$);

\end{tikzpicture}
}
\caption{Language mixing entropy across STEM subjects in
  m-MMLU for various reasoning models. Arrows indicate
  increasing entropy trends with subject difficulty (Elementary $\rightarrow$ High school $\rightarrow$ College)  in mathematics, chemistry, physics, and computer science subjects.}
\label{tab:mmlu-entropy}
\end{table*}


\begin{figure*}
    \centering
    \includegraphics[width=0.9\linewidth]{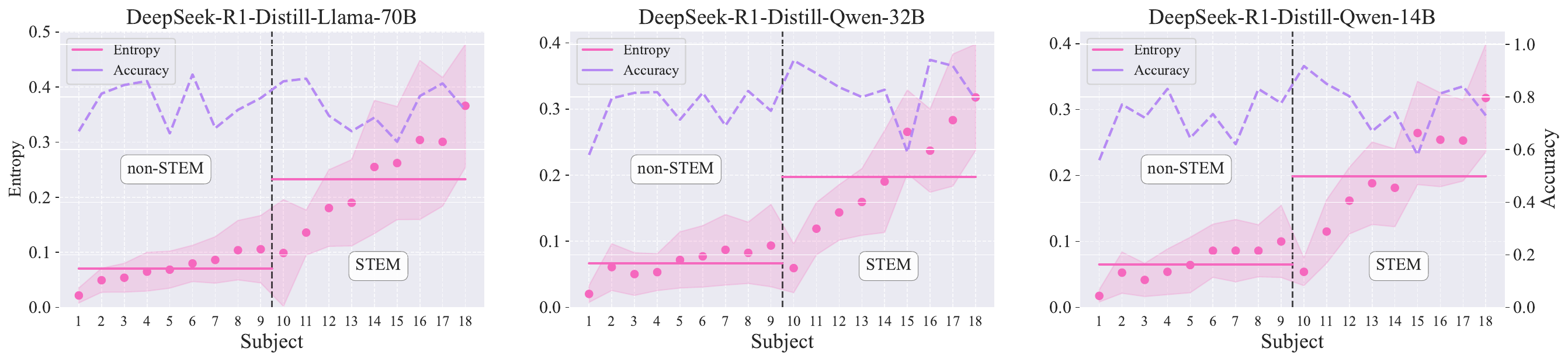}
    \caption{Language mixing entropy (\pink{pink}) and task accuracy (\purple{purple}) across 18 m-MMLU subjects for various reasoning models. Language mixing is notably more pronounced in STEM subjects.}
    \label{fig:mmlu-entropy}
\end{figure*}

\subsection{Results}
We analyze the language mixing patterns in RLMs across different input languages, task difficulty levels (in K\&K), and subject areas (in m-MMLU). Our key observations are summarized below.

\begin{tcolorbox}[findingbox, title=Finding 1]
Language mixing is most prevalent when the input is neither English nor Chinese.
\end{tcolorbox}

Figure~\ref{fig:showcase-pattern-DS-70B} shows the language
usage composition in reasoning traces and final answers
across input languages and difficulty levels in the K\&K
dataset for \texttt{DeepSeek-R1-Distill-Llama-70B}. Results
of other models and on the m-MMLU dataset are provided in
Appendix~\ref{sec:appendix-pattern-res}. We observe that
when the input is English or Chinese, the reasoning remains
mostly in the input language. In contrast,
Arabic, French, Hindi, and Japanese yield more
mixed-language reasoning, often incorporating English and/or
Chinese in the intermediate steps.
In some cases,
the composition is even more complex. For instance, with Arabic prompts, \texttt{R1-70B} produces reasoning traces that mix Arabic, English, Persian, and Chinese (Figure~\ref{fig:showcase-pattern-DS-70B}), while \texttt{R1-32B} generates traces involving Arabic, English, Spanish, and Chinese (Figure~\ref{fig:r1_32b_kk}).

\begin{tcolorbox}[findingbox, title=Finding 2]
RLMs mix languages during reasoning, yet tend to generate answers in the input language.
\end{tcolorbox}
\label{subsec:finding-difficulty}

While reasoning traces exhibit language mixing, the final answers remain more aligned with the input language. This indicates that language mixing occurs primarily in the intermediate reasoning phase, not in the final output. 
This behavior is unsurprising, as reasoning models are generally trained with supervision or reward signals focused on the final answer, encouraging alignment with the input language, while leaving the reasoning steps unconstrained. As a result, models are free to mix languages during reasoning if it helps them reach a correct answer more effectively.

\begin{tcolorbox}[findingbox, title=Finding 3]
The degree of language mixing increases with task difficulty.
\end{tcolorbox}

As shown in Figure~\ref{fig:kk-entropy}, language mixing entropy in reasoning traces rises with task difficulty across languages and models. 
This trend is also evident in the m-MMLU dataset, which includes subjects with varying difficulty levels, e.g., mathematics at the elementary, high school, and college levels. As shown in Table~\ref{tab:mmlu-entropy}, the average entropy\footnote{The entropy is averaged across languages for each subject.} consistently increases with subject difficulty across models, further confirming that task difficulty is an important trigger of language mixing behavior in RLMs.

\begin{tcolorbox}[findingbox, title=Finding 4]
Language mixing is more pronounced in STEM subjects.
\end{tcolorbox}

Figure~\ref{fig:mmlu-entropy} shows language mixing entropy across 18 m-MMLU subjects\footnote{The subject order is provided in Appendix~\ref{sec:appendix-pattern-res}.} for \texttt{R1-70B}, \texttt{R1-32B} and \texttt{R1-14B}. Entropy values are averaged across all evaluated languages.
Additional results for other models and per-language
breakdowns are shown in
Figure~\ref{fig:mmlu-entropy-others}
and Figure~\ref{fig:mmlu-entropy-per-language} in
Appendix~\ref{sec:appendix-pattern-res}. STEM subjects,
starting from tick mark 10 in the figure, consistently exhibit higher entropy than Humanities, Social Sciences, or Other domains, suggesting that technical content tends to induce more language mixing during reasoning.
While entropy still increases with task difficulty within individual STEM domains (Table~\ref{tab:mmlu-entropy}), the overall entropy gap between STEM and non-STEM subjects appears more related to subject type, as they exhibit no clear difference in difficulty based on accuracy (the purple curve in Figure~\ref{fig:mmlu-entropy}).

\enote{hs}{the last bit above reads like a contradiction to
me. first you say that language mixing increases with task
difficulty. (finding 3)
then you say that language mixing is more tied to subject
type than to task difficulty. or am i misunderstanding something?}

\begin{tcolorbox}[findingbox, title=Finding 5]
Distillation amplifies language mixing.
\end{tcolorbox}

Table~\ref{tab:base-model-entropy} compares the language mixing entropy for three DeepSeek backbone models and their corresponding distilled variants. Across all model pairs, the distilled versions consistently exhibit higher average entropy values, particularly when the input language is neither English nor Chinese. Although \citet{guo2025deepseek} do not provide details on the language composition of the distillation data, the observed trend suggests that an English- and Chinese-heavy training distribution may lead the model to rely more on these high-resource language features during reasoning, thereby amplifying language mixing in multilingual settings.

\begin{figure*}[h]
    \centering
    \includegraphics[width=0.935\linewidth]{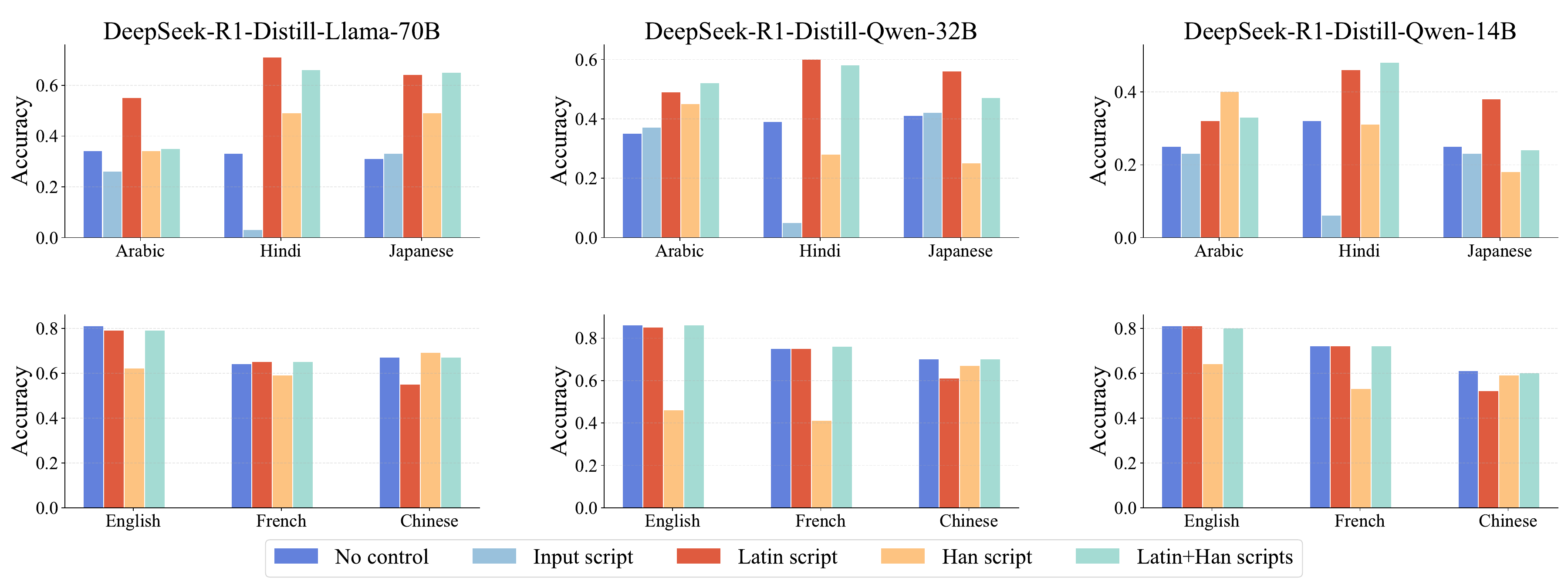}
    \caption{Accuracy on the K\&K dataset under script-controlled reasoning. Latin or Han script control boosts performance for Arabic, Hindi, and Japanese, while native scripts yield the best results for English, French, and Chinese, highlighting the impact of script choice on reasoning efficacy.}
    \label{fig:script-control-res}
\end{figure*}

\begin{table}[t]
  \centering
  \footnotesize
  \scalebox{0.82}{
    \begin{tabular}{lccccccc}
    \toprule
          & \textbf{ar} & \textbf{en} & \textbf{fr} & \textbf{hi} & \textbf{ja} & \textbf{zh} & \textbf{AVG} \\
    \midrule
    \textit{R1} & 0.06  & 0.02  & 0.48  & 0.22  & 0.01  & 0.08  & 0.15 \\
    \midrule
    \textit{Qwen2.5-14B} & \textbf{0.07}  & \textbf{0.01}  & 0.32  & 0.05  & \textbf{0.04}  & 0.10  & 0.10 \\
    \textit{R1-14B} & 0.06  & 0.00  & \textbf{0.59}  & \textbf{0.32}  & 0.01  & \textbf{0.11}  & \textbf{0.18} \\
    \midrule
    \textit{Qwen2.5-32B} & 0.25  & 0.01  & 0.25  & 0.14  & 0.12  & \textbf{0.26}  & 0.17 \\
    \textit{R1-32B} & \textbf{0.39}  & \textbf{0.02}  & \textbf{0.64}  & \textbf{0.24}  & \textbf{0.16}  & 0.09  & \textbf{0.26 }\\
    \midrule
    \textit{Llama3.3-70B} & 0.08  & 0.00  & 0.12  & 0.14  & 0.00  & 0.07  & 0.07 \\
    \textit{R1-70B} & \textbf{0.51}  & \textbf{0.01}  & \textbf{0.43}  & \textbf{0.18}  &\textbf{ 0.14}  & \textbf{0.12}  & \textbf{0.23} \\
    \bottomrule
    \end{tabular}%
    }
  \caption{Language mixing entropy for backbone models and their distilled variants on the K\&K dataset. Distilled reasoning models (bottom row in each pair) exhibit higher entropy than their corresponding base models.}
  \label{tab:base-model-entropy}%
\end{table}%

\section{Performance Impact of Reasoning Languages}
\label{sec:performance-impact}

\subsection{Method: Script-controlled Generation}
\label{sec:script-control-intro}

As discussed in Section~\ref{sec:related-work-rlms}, the impact of language mixing on RLM performance remains debated. To address this, we adopt a direct script control approach to evaluate how reasoning language influences model performance.

We constrain models to generate reasoning steps using only specific script(s) by masking the logits of all other scripts during decoding. 
While language-level control is difficult due to token overlap, script-level control offers a clean separation based on Unicode, as tokens from different scripts do not overlap.
We apply this script control during the reasoning phase (though it could also be applied to the answer phase); the final answer is generated freely after the ``</think>'' token, as models typically default to the input language for answers (as shown in Section~\ref{sec:pattern-res}). For examples of model generation under different script control settings, see \Cref{fig:script-control-exp-free,fig:script-control-exp-latin,fig:script-control-exp-han,fig:script-control-exp-arabic} in Appendix~\ref{sec:appendix-script-control-examples}.

\subsection{Results}


We compare model performance across various reasoning modes:
unconstrained, single-script, and multi-script control to
directly measure the impact of script (and implicitly,
language) choice on reasoning
performance. Figure~\ref{fig:script-control-res} shows
results for six input languages from the K\&K dataset: three
written in non-Latin/Han scripts (Arabic, Hindi, Japanese),
two in Latin (English, French) and one in Han
(Chinese), on three reasoning models: \texttt{R1-70B}, \texttt{R1-32B}, and \texttt{R1-14B}.\footnote{These models are selected to cover a range of model sizes and base architectures.} 

\begin{tcolorbox}[findingbox, title=Finding 6]
Non-Latin/Han-script languages benefit significantly from reasoning in Latin or Han scripts.
\end{tcolorbox}

For Arabic, Hindi, and Japanese, forcing reasoning in Latin or Han scripts significantly improves performance. In Hindi, for example, switching to Latin  yields gains ranging from 44\% to 115\%. In contrast, using the native script consistently results in the lowest accuracy, indicating that models struggle with scripts underrepresented in their training data. For multi-script control (see Table~\ref{tab:script-control-all-1} and Table~\ref{tab:script-control-all-2} in Appendix~\ref{sec:appendix-script-control-res}), reasoning in both Latin and Han scripts does not outperform single-script control, and including the input script alongside Latin and/or Han leads to suboptimal results.

\begin{tcolorbox}[findingbox, title=Finding 7]
Latin and Han-script languages favor their native script.
\end{tcolorbox}

For English, French, and Chinese, reasoning in the native script yields performance comparable to the unconstrained setting. However, switching scripts (e.g., Han for English/French or Latin for Chinese) leads to notable performance drops, suggesting that mismatched scripts disrupt alignment with the model’s internal reasoning patterns.

These results highlight a strong connection between script choice and reasoning performance. Models internally favor Latin and Han scripts, which benefits non-Latin/Han inputs when reasoning is constrained to those scripts. For Latin or Han-script languages, maintaining script consistency is optimal, while mismatches hurt the reasoning performance. This suggests that language mixing reflects the models' learned association between dominant scripts and reasoning competence. 

\section{Internal Causes of Language Mixing}
\label{sec:internal-causes-res}

\subsection{Method: Logit Lens Analysis}
\label{sec:logit-lens-intro}
Inspired by prior work on latent language dynamics in LLMs \citep{wendler-etal-2024-llamas, wang-etal-2025-lost-multilinguality}, we use Logit Lens \citep{nostal2022logitlens} to examine the script composition of internal representations of RLMs and connect it to the scripts used in the models' external reasoning output. 

We use Logit Lens to project intermediate layer representations onto the vocabulary space and identify the script of the top-ranked token at each layer.\footnote{We operate at the script-level (rather than language-level) to avoid ambiguity due to token overlap across languages.} By averaging predictions across all K\&K samples, we obtain a layer-wise distribution of script usage. We then compare these internal patterns to the script usage in reasoning traces, tracking how both evolve with task difficulty. This allows us to directly link the model’s external reasoning behavior with its internal processing dynamics.

To quantify this connection, we compute the Pearson correlation between script usage across difficulty levels in hidden layers and reasoning traces. Specifically: (1) for each difficulty level (ranging from 2 to 8 ppl), we calculate the proportion of tokens belonging to different scripts (e.g., Latin, Devanagari) in both the latent space (estimated with the Logit Lens) and the reasoning trace (measured via script-level detection); and (2) for each script, we compute the Pearson correlation coefficient between its percentage across difficulty levels in the latent space and its corresponding percentage in the reasoning trace. A more detailed description of the Pearson correlation calculation is provided in Appendix~\ref{sec:pearson_details}.
\subsection{Results}

\begin{figure}[h]
    \centering

    \begin{subfigure}[t]{0.48\linewidth}
        \centering
        \includegraphics[width=\linewidth]{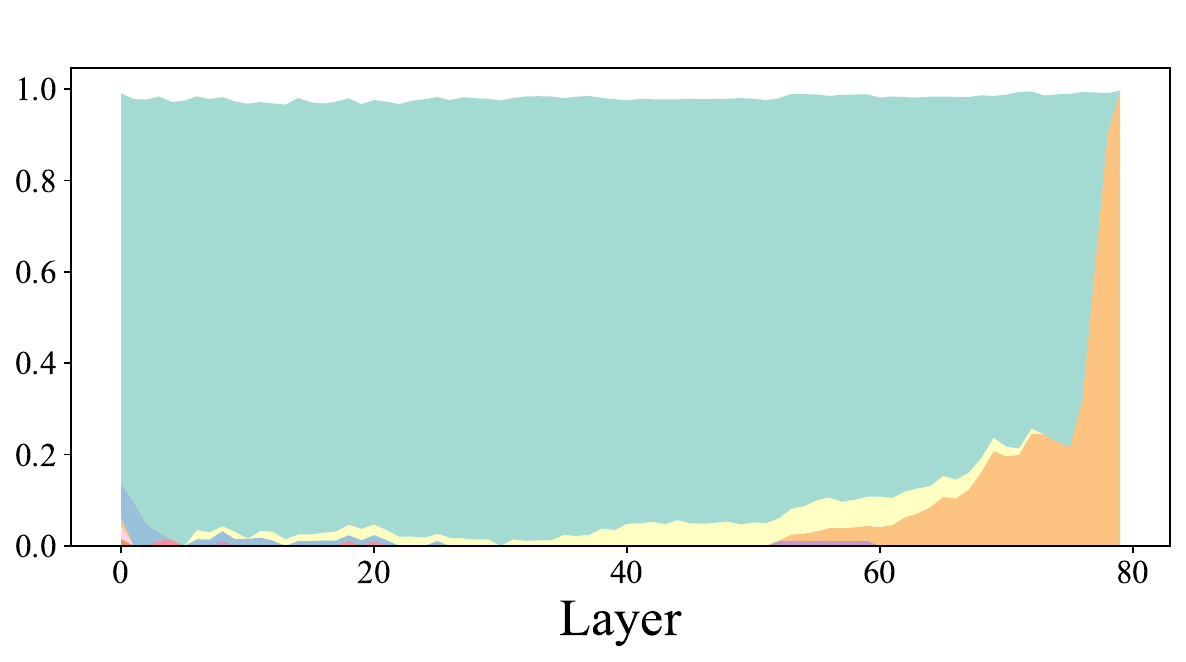}
        \caption{Hidden layer script composition (2ppl).}
        \label{fig:showcase-internal-2ppl}
    \end{subfigure}
    \hfill
    \begin{subfigure}[t]{0.48\linewidth}
        \centering
        \includegraphics[width=\linewidth]{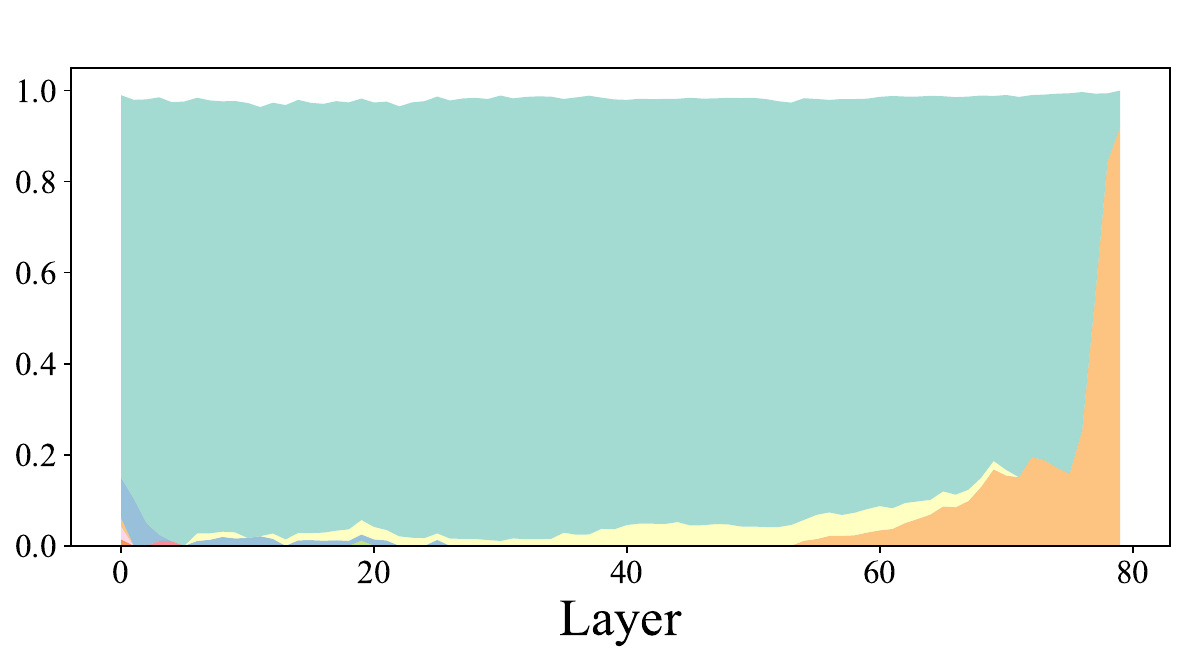}
        \caption{Hidden layer script composition (5ppl).}
        \label{fig:showcase-internal-5ppl}
    \end{subfigure}

    \begin{subfigure}[t]{0.48\linewidth}
        \centering
        \includegraphics[width=\linewidth]{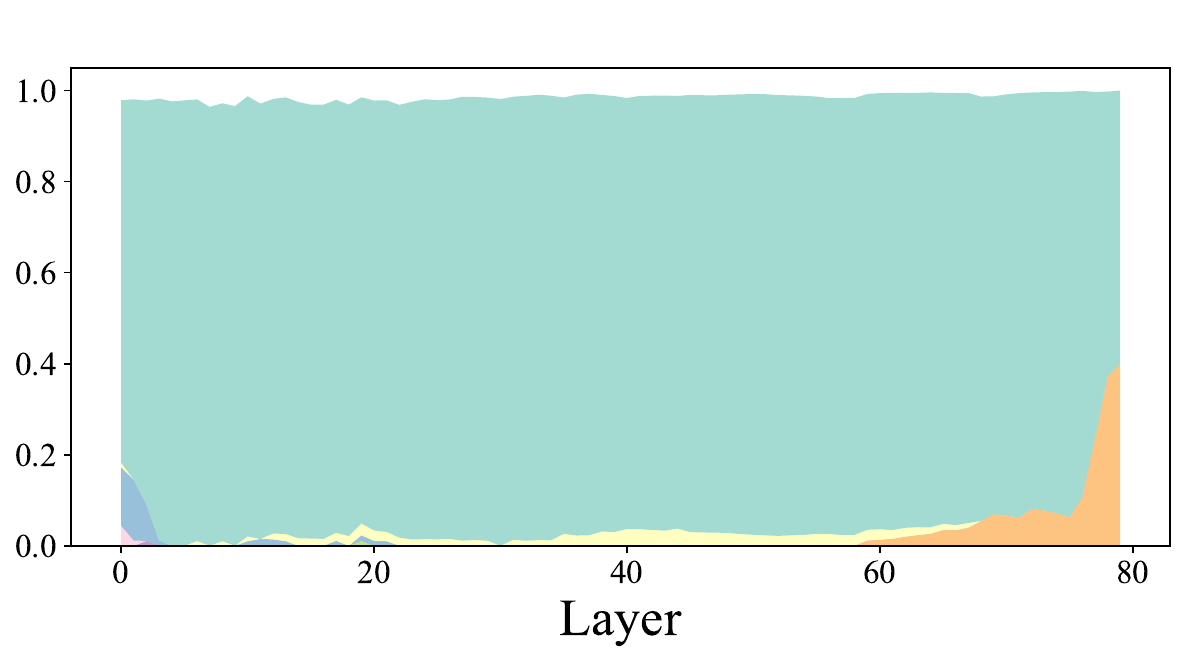}
        \caption{Hidden layer script composition (8ppl).}
        \label{fig:showcase-internal-8ppl}
    \end{subfigure}
    \hfill
    \begin{subfigure}[t]{0.48\linewidth}
        \centering
        \includegraphics[width=\linewidth]{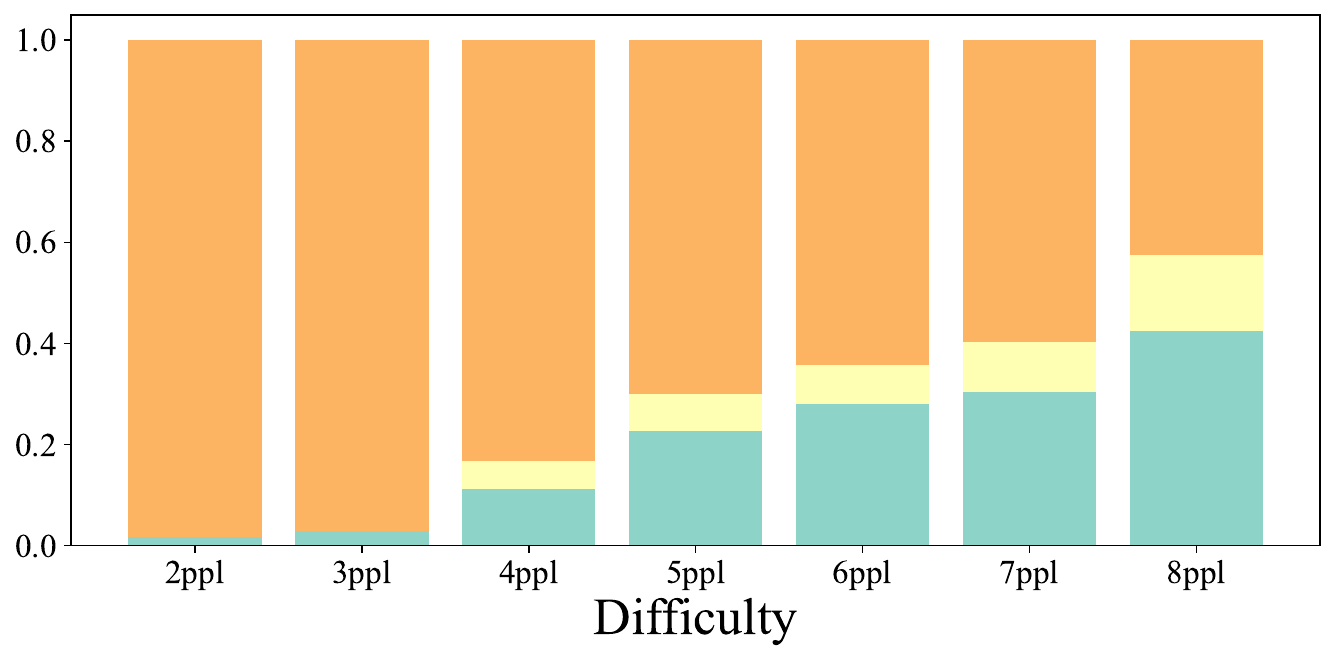}
        \caption{Reasoning trace script composition across difficulty levels.}
        \label{fig:showcase-external}
    \end{subfigure}

    \begin{center}
        \includegraphics[width=\linewidth]{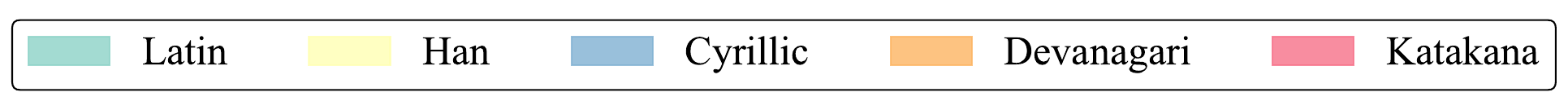}
    \end{center}
    \caption{Script composition of internal representations and reasoning traces for Hindi inputs in \texttt{R1-70B}. (a–c) show hidden layer script distributions via Logit Lens across three difficulty levels. 
    (d) shows reasoning trace script composition mirroring internal trends: Latin usage increases and Devanagari decreases with difficulty.}
    \label{fig:showcase-internal-external}
\end{figure}

Figure~\ref{fig:showcase-internal-external} shows that the
hidden layer script composition
(i.e., the internal representation)
for Hindi inputs
at
difficulty levels 2ppl, 5ppl, and 8ppl in \texttt{R1-70B},
is consistently dominated by the Latin script.
Devanagari (Hindi’s script) appears only in
final layers. This aligns with prior findings that Llama
models primarily "think" in English (i.e., Latin
script) \citep{wendler-etal-2024-llamas, wang-etal-2025-lost-multilinguality}.

\begin{tcolorbox}[findingbox, title=Finding 8]
Language mixing reflects the internal processing patterns of RLMs, as reasoning traces mirror the model’s preferred scripts.
\end{tcolorbox}

\begin{table}[h]
  \centering
  \footnotesize
  \scalebox{1}{
    \begin{tabular}{lccc}
    \toprule
    \textbf{Correlation} & \textbf{\texttt{R1-70B}} & \textbf{\texttt{R1-32B}} & \textbf{\texttt{R1-14B}} \\
    \midrule
    \textbf{\textit{Arabic}} &       &        & \\
    Arabic script   & 0.7411 & 0.9960 & 0.8986 \\
    Latin script    & 0.7168 & 0.7751 & 0.7774 \\
    \midrule
    \textbf{\textit{Hindi}} &       &        & \\
    Devanagari script & 0.9019 & 0.9352 & 0.8840  \\
    Latin script     & 0.8875 & 0.9173 &  0.8969 \\
    \bottomrule
    \end{tabular}%
  }
  \caption{Pearson correlation between script usage
  in hidden layers and reasoning traces across task difficulty levels, showing strong alignment between internal representations and external reasoning outputs.}
  \label{tab:internal-external-correlation-res}%
\end{table}%


As task difficulty increases, Latin usage rises while Devanagari decreases. The same trend appears in the reasoning traces (Figure~\ref{fig:showcase-external}), suggesting a strong link between internal processing and external reasoning behavior.
The Pearson correlations between script usage across difficulty levels in hidden layers and reasoning traces are reported in Table~\ref{tab:internal-external-correlation-res}. 
We observe consistently high correlations for both Arabic and Hindi inputs in \texttt{R1-70B}, \texttt{R1-32B} and \texttt{R1-14B} models.
These results provide evidence that language mixing in reasoning traces reflects internal processing patterns. In particular, the internal preference for the Latin script explains why RLMs tend to mix into Latin-script reasoning, especially when processing underrepresented scripts.

\section{Discussion and Future Works}
\paragraph{Is language mixing a solved problem in RLMs?} 
Although \citet{guo2025deepseek} attempt to mitigate language mixing by incorporating a language consistency reward during reinforcement learning, our evaluation reveals that language mixing persists in \texttt{DeepSeek-R1}, as shown, for instance, in Figure~\ref{fig:kk-entropy} and Figure~\ref{fig:r1_kk}. Regardless of whether language mixing should be removed (which we discuss in the following), our findings indicate that eliminating it with targeted optimization during training remains technically challenging.

\paragraph{Should language mixing be eliminated?} 
Our investigation into the impact of language mixing on reasoning performance (Section~\ref{sec:performance-impact}) shows that constraining reasoning to high-resource language scripts, such as Latin or Han, significantly improves performance on under-resourced language inputs like Arabic, Hindi, and Japanese. This raises questions about whether language mixing is truly undesirable. Notably, \citet{guo2025deepseek} also report that enforcing language consistency leads to performance degradation, suggesting that language mixing may serve as a functional adaptation that supports effective multilingual reasoning.


\paragraph{Reasoning language control.}
Our study shows that script control significantly boosts performance for under-resourced languages, highlighting the potential of reasoning language control in RLMs. Our findings in Section~\ref{sec:performance-impact} and Section~\ref{sec:internal-causes-res} point to two potential future directions: (1) developing fine-grained \emph{language-level} control through constrained decoding, extending beyond \emph{script-level} masking, and (2) steering representations through representation engineering \citep{zou2023representation} or latent space reasoning methods \citep{hao2024training} to align the model’s latent processing with the desired reasoning language. These approaches may offer more flexible control over language mixing and enable more controllable and performant RLMs.

\paragraph{Impact of distillation on language mixing.}
Table~\ref{tab:base-model-entropy} shows that DeepSeek-R1 distilled models consistently exhibit higher language mixing entropy than their original backbones, suggesting distillation amplifies language mixing when the input language is neither English nor Chinese.
This highlights the need to better understand how distillation affects language mixing \citep{yong2025crosslingual}, which is a promising future work direction. Moreover, multilingual distillation strategies may help reduce unnecessary mixing while preserving or improving performance.

\paragraph{Adaptive reasoning language selection.}
Our results show that switching to high-resource languages during reasoning can substantially improve model performance on under-resourced inputs. This suggests a promising direction for future work: training RLMs to adaptively determine which language to use at different stages of reasoning, rather than relying on static language constraints. Such adaptive control could allow the model to dynamically leverage the strengths of high-resource languages while still accommodating the input language, ultimately realizing more robust multilingual reasoning. This idea is aligned with recent advances in adaptive memory usage for reasoning \citep{yan2025memory}, which highlight the benefits of enabling models to flexibly select internal strategies based on task demands.

\section{Conclusions}
In this work, we presented the first systematic investigation of language mixing in reasoning language models, analyzing its occurrence patterns, performance impact, and internal causes. Our findings reveal that language mixing is most likely to occur when the input is neither English nor Chinese, and becomes more pronounced with higher task difficulty and in STEM subjects. 
We further demonstrated that constraining reasoning to Latin or Han scripts significantly improves performance for non-Latin/Han inputs. Finally, our interpretability analysis shows that the language composition of reasoning traces mirrors that of the models' internal representations, suggesting language mixing reflects underlying processing preferences. These insights offer a deeper understanding of multilingual reasoning behavior and provide guidance for developing more controllable and interpretable RLMs.
\section*{Limitations}


While our analysis provides a broad view of language mixing in reasoning language models, it has several limitations.

First, we use script-level control to examine the impact of reasoning language. This approach, while effective, remains relatively coarse. More fine-grained methods, such as language-level constrained decoding, could enable more precise control and allow detailed comparisons of the performance across different reasoning languages given a specific input language.

Second, while we analyze the internal causes of language mixing through mechanistic interpretability, we do not trace its origins in the training process. Future work could investigate the role of training data composition (including pretraining, reinforcement learning, and distillation) and optimization objectives to better understand the roots of this phenomenon.

Third, our study covers 15 languages, but many other languages—particularly those with low resources or unique scripts—remain unexamined. Extending the analysis to a broader set of languages would help validate the generality of our findings.

Lastly, we do not include \texttt{DeepSeek-R1-Zero} \citep{guo2025deepseek} in our evaluation due to resource constraints. As a key variant of reasoning language models with a different training setup, comparing it with \texttt{DeepSeek-R1} could offer further insight into the effect of reinforcement learning training on language mixing.

\section*{Acknowledgments}
This work was partially supported by Deutsche Forschungsgemeinschaft (project SCHU 2246/14-1) and the EU Project SMARTY (GA 101140087).
We thank JUREAP 73 for supporting this work with computing resources. We also thank Sebastian Gerstner, Ahmad Dawar Hakimi, Lea Hirlimann, Valentin Knappich, Felicia Körner, Yihong Liu, Ali Modarressi, Philipp Mondorf, Timo Pierre Schrader, Leonor Veloso, Wei Zhou, Yuqicheng Zhu for their valuable discussions.

\bibliography{anthology_0, anthology_1,custom}

\begin{thebibliography}{32}
\providecommand{\natexlab}[1]{#1}

\bibitem[{Chen et~al.(2025)Chen, Qin, Liu, Peng, Guan, Wang, Hu, Zhou, Gao, and Che}]{chen2025towards}
Qiguang Chen, Libo Qin, Jinhao Liu, Dengyun Peng, Jiannan Guan, Peng Wang, Mengkang Hu, Yuhang Zhou, Te~Gao, and Wanxiang Che. 2025.
\newblock Towards reasoning era: A survey of long chain-of-thought for reasoning large language models.
\newblock \emph{arXiv preprint arXiv:2503.09567}.

\bibitem[{Cuadron et~al.(2025)Cuadron, Li, Ma, Wang, Wang, Zhuang, Liu, Schroeder, Xia, Mao et~al.}]{cuadron2025danger}
Alejandro Cuadron, Dacheng Li, Wenjie Ma, Xingyao Wang, Yichuan Wang, Siyuan Zhuang, Shu Liu, Luis~Gaspar Schroeder, Tian Xia, Huanzhi Mao, and 1 others. 2025.
\newblock The danger of overthinking: Examining the reasoning-action dilemma in agentic tasks.
\newblock \emph{arXiv preprint arXiv:2502.08235}.

\bibitem[{Do{\u{g}}ru{\"o}z et~al.(2021)Do{\u{g}}ru{\"o}z, Sitaram, Bullock, and Toribio}]{dogruoz-etal-2021-survey}
A.~Seza Do{\u{g}}ru{\"o}z, Sunayana Sitaram, Barbara~E. Bullock, and Almeida~Jacqueline Toribio. 2021.
\newblock \href {https://doi.org/10.18653/v1/2021.acl-long.131} {A survey of code-switching: Linguistic and social perspectives for language technologies}.
\newblock In \emph{Proceedings of the 59th Annual Meeting of the Association for Computational Linguistics and the 11th International Joint Conference on Natural Language Processing (Volume 1: Long Papers)}, pages 1654--1666, Online. Association for Computational Linguistics.

\bibitem[{Dumas et~al.(2024)Dumas, Veselovsky, Monea, West, and Wendler}]{dumas2024llamas}
Cl{\'e}ment Dumas, Veniamin Veselovsky, Giovanni Monea, Robert West, and Chris Wendler. 2024.
\newblock \href {https://openreview.net/forum?id=0ku2hIm4BS} {How do llamas process multilingual text? a latent exploration through activation patching}.
\newblock In \emph{ICML 2024 Workshop on Mechanistic Interpretability}.

\bibitem[{Google(2025)}]{gemini}
Google. 2025.
\newblock \href {https://ai.google.dev/gemini-api/docs/models} {Gemini 2.0 flash thinking.}

\bibitem[{Guo et~al.(2025)Guo, Yang, Zhang, Song, Zhang, Xu, Zhu, Ma, Wang, Bi et~al.}]{guo2025deepseek}
Daya Guo, Dejian Yang, Haowei Zhang, Junxiao Song, Ruoyu Zhang, Runxin Xu, Qihao Zhu, Shirong Ma, Peiyi Wang, Xiao Bi, and 1 others. 2025.
\newblock \href {https://arxiv.org/abs/2501.12948} {Deepseek-r1: Incentivizing reasoning capability in llms via reinforcement learning}.
\newblock \emph{arXiv preprint arXiv:2501.12948}.

\bibitem[{Hao et~al.(2024)Hao, Sukhbaatar, Su, Li, Hu, Weston, and Tian}]{hao2024training}
Shibo Hao, Sainbayar Sukhbaatar, DiJia Su, Xian Li, Zhiting Hu, Jason Weston, and Yuandong Tian. 2024.
\newblock Training large language models to reason in a continuous latent space.
\newblock \emph{arXiv preprint arXiv:2412.06769}.

\bibitem[{Hendrycks et~al.(2020)Hendrycks, Burns, Basart, Zou, Mazeika, Song, and Steinhardt}]{hendrycks2020measuring}
Dan Hendrycks, Collin Burns, Steven Basart, Andy Zou, Mantas Mazeika, Dawn Song, and Jacob Steinhardt. 2020.
\newblock \href {https://arxiv.org/abs/2009.03300} {Measuring massive multitask language understanding}.
\newblock \emph{arXiv preprint arXiv:2009.03300}.

\bibitem[{Jaech et~al.(2024)Jaech, Kalai, Lerer, Richardson, El-Kishky, Low, Helyar, Madry, Beutel, Carney et~al.}]{jaech2024openai}
Aaron Jaech, Adam Kalai, Adam Lerer, Adam Richardson, Ahmed El-Kishky, Aiden Low, Alec Helyar, Aleksander Madry, Alex Beutel, Alex Carney, and 1 others. 2024.
\newblock Openai o1 system card.
\newblock \emph{arXiv preprint arXiv:2412.16720}.

\bibitem[{Joulin et~al.(2017)Joulin, Grave, Bojanowski, and Mikolov}]{joulin-etal-2017-bag}
Armand Joulin, Edouard Grave, Piotr Bojanowski, and Tomas Mikolov. 2017.
\newblock \href {https://aclanthology.org/E17-2068/} {Bag of tricks for efficient text classification}.
\newblock In \emph{Proceedings of the 15th Conference of the {E}uropean Chapter of the Association for Computational Linguistics: Volume 2, Short Papers}, pages 427--431, Valencia, Spain. Association for Computational Linguistics.

\bibitem[{Khanuja et~al.(2020)Khanuja, Dandapat, Srinivasan, Sitaram, and Choudhury}]{khanuja-etal-2020-gluecos}
Simran Khanuja, Sandipan Dandapat, Anirudh Srinivasan, Sunayana Sitaram, and Monojit Choudhury. 2020.
\newblock \href {https://doi.org/10.18653/v1/2020.acl-main.329} {{GLUEC}o{S}: An evaluation benchmark for code-switched {NLP}}.
\newblock In \emph{Proceedings of the 58th Annual Meeting of the Association for Computational Linguistics}, pages 3575--3585, Online. Association for Computational Linguistics.

\bibitem[{Li et~al.(2024{\natexlab{a}})Li, Sun, Li, Chen, Liu, Weng, Bai, and Hu}]{li2024distinct}
Bin Li, Bin Sun, Shutao Li, Encheng Chen, Hongru Liu, Yixuan Weng, Yongping Bai, and Meiling Hu. 2024{\natexlab{a}}.
\newblock Distinct but correct: generating diversified and entity-revised medical response.
\newblock \emph{Science China Information Sciences}, 67(3):132106.

\bibitem[{Li et~al.(2024{\natexlab{b}})Li, Weng, Xia, and Deng}]{li2024towards}
Bin Li, Yixuan Weng, Fei Xia, and Hanjun Deng. 2024{\natexlab{b}}.
\newblock Towards better chinese-centric neural machine translation for low-resource languages.
\newblock \emph{Computer Speech \& Language}, 84:101566.

\bibitem[{Liu et~al.(2025)Liu, Wang, Kargaran, K{\"o}rner, Nie, Plank, Yvon, and Sch{\"u}tze}]{liu2025tracing}
Yihong Liu, Mingyang Wang, Amir~Hossein Kargaran, Felicia K{\"o}rner, Ercong Nie, Barbara Plank, Fran{\c{c}}ois Yvon, and Hinrich Sch{\"u}tze. 2025.
\newblock \href {https://arxiv.org/abs/2505.14824} {Tracing multilingual factual knowledge acquisition in pretraining}.
\newblock \emph{arXiv preprint arXiv:2505.14824}.

\bibitem[{Marchisio et~al.(2024)Marchisio, Ko, Berard, Dehaze, and Ruder}]{marchisio-etal-2024-understanding}
Kelly Marchisio, Wei-Yin Ko, Alexandre Berard, Th{\'e}o Dehaze, and Sebastian Ruder. 2024.
\newblock \href {https://doi.org/10.18653/v1/2024.emnlp-main.380} {Understanding and mitigating language confusion in {LLM}s}.
\newblock In \emph{Proceedings of the 2024 Conference on Empirical Methods in Natural Language Processing}, pages 6653--6677, Miami, Florida, USA. Association for Computational Linguistics.

\bibitem[{Nie et~al.(2025)Nie, Schmid, and Sch{\"u}tze}]{nie2025mechanistic}
Ercong Nie, Helmut Schmid, and Hinrich Sch{\"u}tze. 2025.
\newblock \href {https://arxiv.org/abs/2505.16538} {Mechanistic understanding and mitigation of language confusion in english-centric large language models}.
\newblock \emph{arXiv preprint arXiv:2505.16538}.

\bibitem[{Nostalgebraist(2022)}]{nostal2022logitlens}
Nostalgebraist. 2022.
\newblock \href {https://www.lesswrong.com/posts/AcKRB8wDpdaN6v6ru/interpreting-gpt-the-logit-lens} {Interpreting gpt: The logit lens}.

\bibitem[{OpenAI(2025)}]{openai2024o3o4}
OpenAI. 2025.
\newblock \href {https://openai.com/index/o3-o4-mini-system-card/} {Openai o3 and o4-mini system card.}

\bibitem[{QwenTeam(2024)}]{qwq-32b-preview}
QwenTeam. 2024.
\newblock \href {https://qwenlm.github.io/blog/qwq-32b-preview/} {Qwq: Reflect deeply on the boundaries of the unknown}.

\bibitem[{QwenTeam(2025{\natexlab{a}})}]{qwen3}
QwenTeam. 2025{\natexlab{a}}.
\newblock \href {https://qwenlm.github.io/blog/qwen3/} {Qwen3: Think deeper, act faster}.

\bibitem[{QwenTeam(2025{\natexlab{b}})}]{qwq-32b}
QwenTeam. 2025{\natexlab{b}}.
\newblock \href {https://qwenlm.github.io/blog/qwq-32b/} {Qwq-32b: Embracing the power of reinforcement learning}.

\bibitem[{Wang et~al.(2025)Wang, Adel, Lange, Liu, Nie, Str{\"o}tgen, and Schuetze}]{wang-etal-2025-lost-multilinguality}
Mingyang Wang, Heike Adel, Lukas Lange, Yihong Liu, Ercong Nie, Jannik Str{\"o}tgen, and Hinrich Schuetze. 2025.
\newblock \href {https://doi.org/10.18653/v1/2025.acl-long.253} {Lost in multilinguality: Dissecting cross-lingual factual inconsistency in transformer language models}.
\newblock In \emph{Proceedings of the 63rd Annual Meeting of the Association for Computational Linguistics (Volume 1: Long Papers)}, pages 5075--5094, Vienna, Austria. Association for Computational Linguistics.

\bibitem[{Wendler et~al.(2024)Wendler, Veselovsky, Monea, and West}]{wendler-etal-2024-llamas}
Chris Wendler, Veniamin Veselovsky, Giovanni Monea, and Robert West. 2024.
\newblock \href {https://doi.org/10.18653/v1/2024.acl-long.820} {Do llamas work in {E}nglish? on the latent language of multilingual transformers}.
\newblock In \emph{Proceedings of the 62nd Annual Meeting of the Association for Computational Linguistics (Volume 1: Long Papers)}, pages 15366--15394, Bangkok, Thailand. Association for Computational Linguistics.

\bibitem[{Winata et~al.(2023)Winata, Aji, Yong, and Solorio}]{winata-etal-2023-decades}
Genta Winata, Alham~Fikri Aji, Zheng~Xin Yong, and Thamar Solorio. 2023.
\newblock \href {https://doi.org/10.18653/v1/2023.findings-acl.185} {The decades progress on code-switching research in {NLP}: A systematic survey on trends and challenges}.
\newblock In \emph{Findings of the Association for Computational Linguistics: ACL 2023}, pages 2936--2978, Toronto, Canada. Association for Computational Linguistics.

\bibitem[{Xie et~al.(2024)Xie, Huang, Zhang, Yu, Chen, Lin, Li, Ghazi, and Kumar}]{xie2024memorization}
Chulin Xie, Yangsibo Huang, Chiyuan Zhang, Da~Yu, Xinyun Chen, Bill~Yuchen Lin, Bo~Li, Badih Ghazi, and Ravi Kumar. 2024.
\newblock On memorization of large language models in logical reasoning.
\newblock \emph{arXiv preprint arXiv:2410.23123}.

\bibitem[{Xie et~al.(2025)Xie, Gao, Ren, Luo, Hong, Dai, Zhou, Qiu, Wu, and Luo}]{xie2025logicrlunleashingllmreasoning}
Tian Xie, Zitian Gao, Qingnan Ren, Haoming Luo, Yuqian Hong, Bryan Dai, Joey Zhou, Kai Qiu, Zhirong Wu, and Chong Luo. 2025.
\newblock \href {https://arxiv.org/abs/2502.14768} {Logic-rl: Unleashing llm reasoning with rule-based reinforcement learning}.
\newblock \emph{Preprint}, arXiv:2502.14768.

\bibitem[{Xu et~al.(2025)Xu, Hao, Zong, Wang, Zhang, Wang, Lan, Gong, Ouyang, Meng et~al.}]{xu2025towards}
Fengli Xu, Qianyue Hao, Zefang Zong, Jingwei Wang, Yunke Zhang, Jingyi Wang, Xiaochong Lan, Jiahui Gong, Tianjian Ouyang, Fanjin Meng, and 1 others. 2025.
\newblock Towards large reasoning models: A survey of reinforced reasoning with large language models.
\newblock \emph{arXiv preprint arXiv:2501.09686}.

\bibitem[{Yan et~al.(2025)Yan, Yang, Huang, Nie, Ding, Li, Ma, Sch{\"u}tze, Tresp, and Ma}]{yan2025memory}
Sikuan Yan, Xiufeng Yang, Zuchao Huang, Ercong Nie, Zifeng Ding, Zonggen Li, Xiaowen Ma, Hinrich Sch{\"u}tze, Volker Tresp, and Yunpu Ma. 2025.
\newblock \href {https://arxiv.org/abs/2508.19828} {Memory-r1: Enhancing large language model agents to manage and utilize memories via reinforcement learning}.
\newblock \emph{arXiv preprint arXiv:2508.19828}.

\bibitem[{Yong et~al.(2025)Yong, Adilazuarda, Mansurov, Zhang, Muennighoff, Eickhoff, Winata, Kreutzer, Bach, and Aji}]{yong2025crosslingual}
Zheng-Xin Yong, M~Farid Adilazuarda, Jonibek Mansurov, Ruochen Zhang, Niklas Muennighoff, Carsten Eickhoff, Genta~Indra Winata, Julia Kreutzer, Stephen~H Bach, and Alham~Fikri Aji. 2025.
\newblock \href {https://arxiv.org/abs/2505.05408} {Crosslingual reasoning through test-time scaling}.
\newblock \emph{arXiv preprint arXiv:2505.05408}.

\bibitem[{Yong et~al.(2023)Yong, Zhang, Forde, Wang, Subramonian, Lovenia, Cahyawijaya, Winata, Sutawika, Cruz, Tan, Phan, Phan, Garcia, Solorio, and Aji}]{yong-etal-2023-prompting}
Zheng~Xin Yong, Ruochen Zhang, Jessica Forde, Skyler Wang, Arjun Subramonian, Holy Lovenia, Samuel Cahyawijaya, Genta Winata, Lintang Sutawika, Jan Christian~Blaise Cruz, Yin~Lin Tan, Long Phan, Long Phan, Rowena Garcia, Thamar Solorio, and Alham~Fikri Aji. 2023.
\newblock \href {https://aclanthology.org/2023.calcs-1.5/} {Prompting multilingual large language models to generate code-mixed texts: The case of south {E}ast {A}sian languages}.
\newblock In \emph{Proceedings of the 6th Workshop on Computational Approaches to Linguistic Code-Switching}, pages 43--63, Singapore. Association for Computational Linguistics.

\bibitem[{Zhang et~al.(2023)Zhang, Cahyawijaya, Cruz, Winata, and Aji}]{zhang-etal-2023-multilingual}
Ruochen Zhang, Samuel Cahyawijaya, Jan Christian~Blaise Cruz, Genta Winata, and Alham~Fikri Aji. 2023.
\newblock \href {https://doi.org/10.18653/v1/2023.emnlp-main.774} {Multilingual large language models are not (yet) code-switchers}.
\newblock In \emph{Proceedings of the 2023 Conference on Empirical Methods in Natural Language Processing}, pages 12567--12582, Singapore. Association for Computational Linguistics.

\bibitem[{Zou et~al.(2023)Zou, Phan, Chen, Campbell, Guo, Ren, Pan, Yin, Mazeika, Dombrowski et~al.}]{zou2023representation}
Andy Zou, Long Phan, Sarah Chen, James Campbell, Phillip Guo, Richard Ren, Alexander Pan, Xuwang Yin, Mantas Mazeika, Ann-Kathrin Dombrowski, and 1 others. 2023.
\newblock \href {https://arxiv.org/abs/2310.01405} {Representation engineering: A top-down approach to ai transparency}.
\newblock \emph{arXiv preprint arXiv:2310.01405}.

\end{thebibliography}
\appendix
\newpage

\section{Appendix}
\label{sec:appendix}

\subsection{Experimental Setup Details}
\subsubsection{Models}
\label{sec:appendix-model}

\paragraph{Model Lists.} As introduced in Section~\ref{sec:exp-setup}, we evaluate a diverse set of reasoning language models (RLMs) across developers. Table~\ref{tab:model-list} lists their full names, backbone models, and reference names used throughout this paper.
\label{sec:appendix-model-list}


\begin{table*}[h]
  \centering
  \footnotesize
  \begin{tabular}{llll}
    \toprule
    \textbf{Developer} & \textbf{Model Name} & \textbf{Backbone Model} & \textbf{Ref. Name} \\
    \midrule
    \multirow{7}{*}{DeepSeek} 
        & DeepSeek-R1-Distill-Qwen-1.5B & Qwen2.5-Math-1.5B & R1-1.5B \\
        & DeepSeek-R1-Distill-Qwen-7B   & Qwen2.5-Math-7B   & R1-7B \\
        & DeepSeek-R1-Distill-Llama-8B  & Llama-3.1-8B       & R1-8B \\
        & DeepSeek-R1-Distill-Qwen-14B  & Qwen2.5-14B        & R1-14B \\
        & DeepSeek-R1-Distill-Qwen-32B  & Qwen2.5-32B        & R1-32B \\
        & DeepSeek-R1-Distill-Llama-70B & Llama-3.3-70B-Instruct & R1-70B \\
        & DeepSeek-R1                   & DeepSeek-V3-Base   & R1 \\
    \midrule
    Google & gemini-2.0-flash-thinking-exp-01-21 & - & Gemini / Gemini 2.0 Flash Thinking \\
    \midrule
    \multirow{4}{*}{Qwen} 
        & QwQ-32B       & Qwen2.5-32B-Instruct & QwQ \\
        & Qwen3-4B      & Qwen3-4B-Base             & Q3-4B / Qwen3-4B \\
        & Qwen3-30B-A3B & Qwen3-30B-A3B-Base        & Q3-30B-A3B / Qwen3-30B-A3B \\
        & Qwen3-32B     & Qwen3-32B-Base            & Q3-32B / Qwen3-32B \\
    \bottomrule
  \end{tabular}
  \caption{Summary of reasoning models evaluated in this work, including their backbone models and the reference names used in the paper.}
  \label{tab:model-list}
\end{table*}

\begin{table*}[h]
  \centering
  \footnotesize
    \begin{tabular}{ll}
    \toprule
    \textbf{Model } & \textbf{Hyperparameters} \\
    \midrule
    DeepSeek R1 series & Temperature=0.6, TopP=0.95 \\
    gemini-2.0-flash-thinking & Temperature = 0.7, TopP = 0.95, TopK = 64 \\
    QwQ-32B & Temperature=0.6, TopP=0.95 \\
    Qwen3 series & Temperature=0.6, TopP=0.95, TopK=20 \\
    \bottomrule
    \end{tabular}%
  \caption{Hyperparameters used for different reasoning models in our evaluation. Settings follow the recommended configurations provided by model developers on HuggingFace to ensure fair comparison.}
  \label{tab:model-hp}%
\end{table*}%

\paragraph{Hyperparameters.} 
\label{sec:appendix-model-hp}
For reproducibility and consistency, we report the decoding hyperparameters used across all models evaluated in our experiments in Table~\ref{tab:model-hp}.

\subsubsection{Datasets}
\label{sec:appendix-dataset}
\paragraph{Knights-and-Knaves dataset translation.}
As introduced in Section~\ref{sec:exp-setup-dataset}, the Knights and Knaves (K\&K) dataset consists of algorithmically generated reasoning puzzles, with difficulty controlled by varying the number of characters (2–8) and the complexity of logical operations. In these puzzles, as illustrated in Figure~\ref{fig:kk-example}, each character is either a \textit{knight}, who always tells the truth, or a \textit{knave}, who always lies. The objective is to determine each character's identity based on their statements. All puzzles are constructed using formal logic rules, ensuring a unique, verifiable solution, which makes the dataset well-suited for analyzing the impact of task difficulty on reasoning in language models. 

To extend the dataset beyond English, we translate it into five additional languages, Arabic (ar), French (fr), Hindi (hi), Japanese (ja), and Chinese (zh), covering diverse scripts and linguistic families. To ensure consistent translations of identity terms (e.g., ``knight'', ``knave'') and character names (e.g., Zoey, Oliver),  we construct a fixed translation map of these identity and character names from English to each target language. We then use \texttt{gpt-4o-mini} to translate the puzzles and solutions while enforcing consistency with this map. The translation prompt is shown in Figure~\ref{fig:kk-translation-prompt}.

\begin{figure}[h]
\centering
\fontsize{9.2pt}{10pt}\selectfont
\begin{tcolorbox}[colback=blue!5!white, colframe=blue!75!black, title=K\&K Translation System Prompt]
You are a professional translator. Please translate the following English text into \texttt{\{target\_language\}} while following these rules:

\begin{itemize}[leftmargin=1.5em]
    \item Translate person names according to this mapping: \texttt{\{name\_map\}}.
    \item Translate ``Knights'' and ``Knaves'' always as follows:
    
    - ``Knights'' $\rightarrow$ 
        \texttt{\{identity\_map['Knights']\}}

    - ``Knight'' $\rightarrow$ \texttt{\{identity\_map['Knight']\}}
    
    - ``Knaves'' $\rightarrow$ \texttt{\{identity\_map['Knaves']\}}
    
    - ``Knave'' $\rightarrow$ \texttt{\{identity\_map['Knave']\}}

    \item Ensure the sentence remains grammatically correct and natural in \texttt{\{target\_language\}}.
    \item Do NOT translate placeholders (if any exist).
    \item Return only the translated text, no extra information.
\end{itemize}
\end{tcolorbox}

    \caption{System prompt used to translate K\&K puzzles into target languages with consistent identity and person name mapping.}
    \label{fig:kk-translation-prompt}
\end{figure}

\begin{table}[h]
\footnotesize
  \centering
    \begin{tabular}{p{1.5cm}p{5cm}} 
    \midrule
    \textbf{Dataset} & \textbf{Languages} \\
    \midrule
    K\&K & Arabic (ar), English (en), French (fr), Hindi (hi), Japenese (ja), Chinese (zh) \\
    \midrule
    m-MMLU & Arabic (AR-XY), Bengali (BN-BD), German (DE-DE), Spanish (ES-LA), French (FR-FR), Hindi (HI-IN), Indonesian (ID-ID), Italian (IT-IT), Japanese (JA-JP), Korean (KO-KR), Brazilian Portuguese (PT-BR), Swahili (SW-KE), Yoruba (YO-NG), Simplified Chinese (ZH-CN) \\
    \midrule
    \end{tabular}%
  \caption{Languages covered in the K\&K and m-MMLU datasets. }
  \label{tab:language-coverage}%
\end{table}%

\begin{table}[h]
\centering
\footnotesize
\begin{tabular}{lll}
\midrule
\textbf{Language}                                & \textbf{Abbreviation} & \textbf{Script}                  \\
\midrule
\multicolumn{3}{l}{\textit{Languages of investigation}} \\
\midrule
Arabic                                  & ar / AR-XY   & Arabic                  \\
Bengali                                 & BN-BD        & Bangla                  \\
Chinese                                 & zh / ZH-CN   & Han                     \\
English                                 & en / EN-US   & Latin                   \\
French                                  & fr           & Latin                   \\
German                                  & DE-DE        & Latin                   \\
Hindi                                   & hi / HI-IN   & Devanagari              \\
Indonesian                              & ID-ID        & Latin                   \\
Italian                                 & IT-IT        & Latin                   \\
Japanese                                & ja / JA-JP   & Han, Hiragana, Katakana \\
Korean                                  & KO-KR        & Hangul                  \\
Portuguese                              & PT-BR        & Latin                   \\
Spanish                                 & ES-LA        & Latin                   \\
Swahili                                 & SW-KE        & Latin                   \\
Yoruba                                  & YO-NG        & Latin                   \\
\midrule
\multicolumn{3}{l}{\textit{Languages emerge in the reasoning trace}} \\
\midrule
Persian                                 & fa           & Perso-Arabic            \\
Marathi                                 & mr           & Devanagari              \\
Russian                                 & ru           & Cyrillic    \\          
\midrule 
\end{tabular}
\caption{Languages examined in our study, with their abbreviations and writing scripts. We distinguish between the \emph{languages of investigation}, which serve as evaluation inputs covered by the K \& K and/or m-MMLU datasets, and the \emph{languages that emerge in the reasoning trace}, which appear spontaneously during the model’s reasoning process.}
\label{tab:lang-abbr-script}%
\end{table}

\paragraph{Further details of m-MMLU}
The m-MMLU dataset covers 15 languages across diverse linguistic and geographic regions as shown in Table~\ref{tab:language-coverage}. This broad coverage allows for a comprehensive assessment of multilingual reasoning abilities across a wide spectrum of languages.
We follow the prompts and the evaluation framework used in simple-evals library.\footnote{\url{https://github.com/openai/simple-evals}}

\subsubsection{Pearson correlation calculation}
\label{sec:pearson_details}
To clarify how the Pearson correlation coefficients in Table~\ref{tab:internal-external-correlation-res} are calculated, we provide a detailed description of the procedure.  

We take Figure~\ref{fig:showcase-internal-external} in the main text as an example (which shows the script composition of internal representations and reasoning traces for Hindi inputs in the \texttt{R1-70B} model). The Pearson correlation is computed as follows:

\begin{enumerate}
    \item For each difficulty level (from 2 to 8 ppl), we collect the percentage of tokens belonging to different scripts, such as Latin and Devanagari. These percentages are measured in (1) the latent space, obtained via Logit Lens analysis of hidden layer activations; and (2) the reasoning trace, obtained through script-level detection of the generated text.

    \item For each script, we then calculate the Pearson correlation coefficient between its sequence of percentages across difficulty levels in the latent space and the corresponding sequence in the reasoning trace. An illustrative example is provided in Table~\ref{tab:pearson-calc-example}. 
\end{enumerate}

These Pearson correlations indicate how closely the model’s internal script usage aligns with the script composition of the generated reasoning trace.

\begin{table*}[h]
\centering
\footnotesize
\begin{tabular}{c|c c|c c}
\hline
\textbf{Difficulty} & \textbf{Latin -- latent (\%)} & \textbf{Latin -- reasoning (\%)} & \textbf{Devanagari -- latent (\%)} & \textbf{Devanagari -- reasoning (\%)} \\
\hline
2 ppl & 77.01 & 0.76  & 17.82 & 99.19 \\
3 ppl & 77.00 & 4.22  & 18.34 & 95.75 \\
4 ppl & 78.54 & 15.05 & 16.96 & 78.11 \\
5 ppl & 80.93 & 31.68 & 15.11 & 61.90 \\
6 ppl & 84.78 & 48.02 & 12.18 & 48.02 \\
7 ppl & 91.46 & 44.71 & 6.60  & 44.33 \\
8 ppl & 91.79 & 63.26 & 6.40  & 21.54 \\
\hline
\end{tabular}
\caption{Example of script composition across difficulty levels for Hindi inputs in the \texttt{R1-70B} model. Percentages of tokens in the latent space (via Logit Lens) and reasoning trace (via script-level detection) are shown for Latin and Devanagari scripts. These values form the basis for computing Pearson correlations reported in the main text.}
\label{tab:pearson-calc-example}
\end{table*}

\subsection{Additional Experimental Results}

\subsubsection{Language Mixing Patterns}
\label{sec:appendix-pattern-res}
\paragraph{m-MMLU complete results.}

\begin{figure*}
    \centering
    \includegraphics[width=1\linewidth]{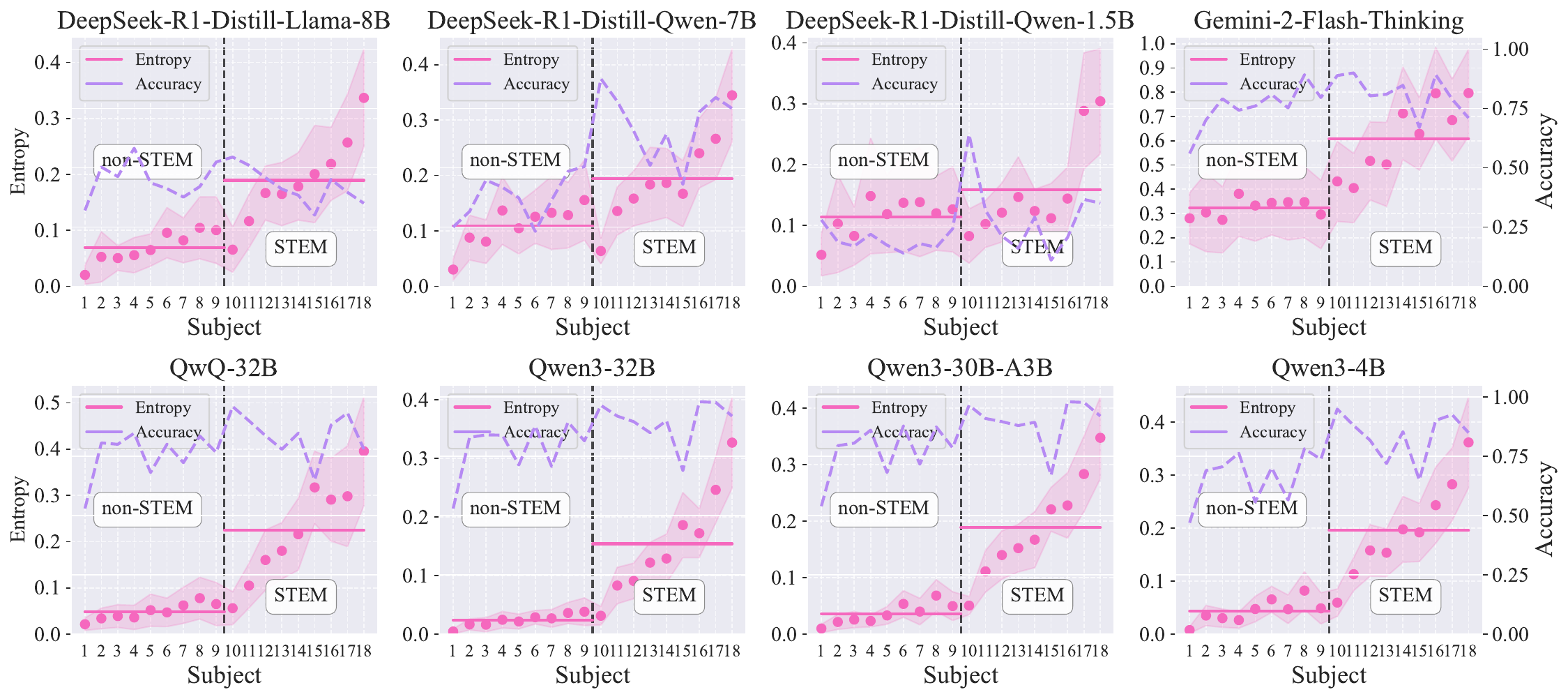}
    \caption{Language mixing entropy (\pink{pink}) and task accuracy (\figureblue{blue}) across 18 m-MMLU subjects for additional reasoning models. Entropy consistently increases in STEM subjects.}
    \label{fig:mmlu-entropy-others}
\end{figure*}

Figure~\ref{fig:mmlu-entropy-others} presents language mixing entropy across m-MMLU subjects for additional reasoning models beyond those shown Figure~\ref{fig:mmlu-entropy}. As in the main results, entropy is averaged across languages. The trend remains consistent: language mixing is markedly higher in STEM subjects (subjects 10–18), compared to non-STEM areas. This pattern reinforces our main finding that technical subjects tend to induce more language mixing during reasoning.

\begin{figure*}[h]
    \centering
    \includegraphics[width=1\linewidth]{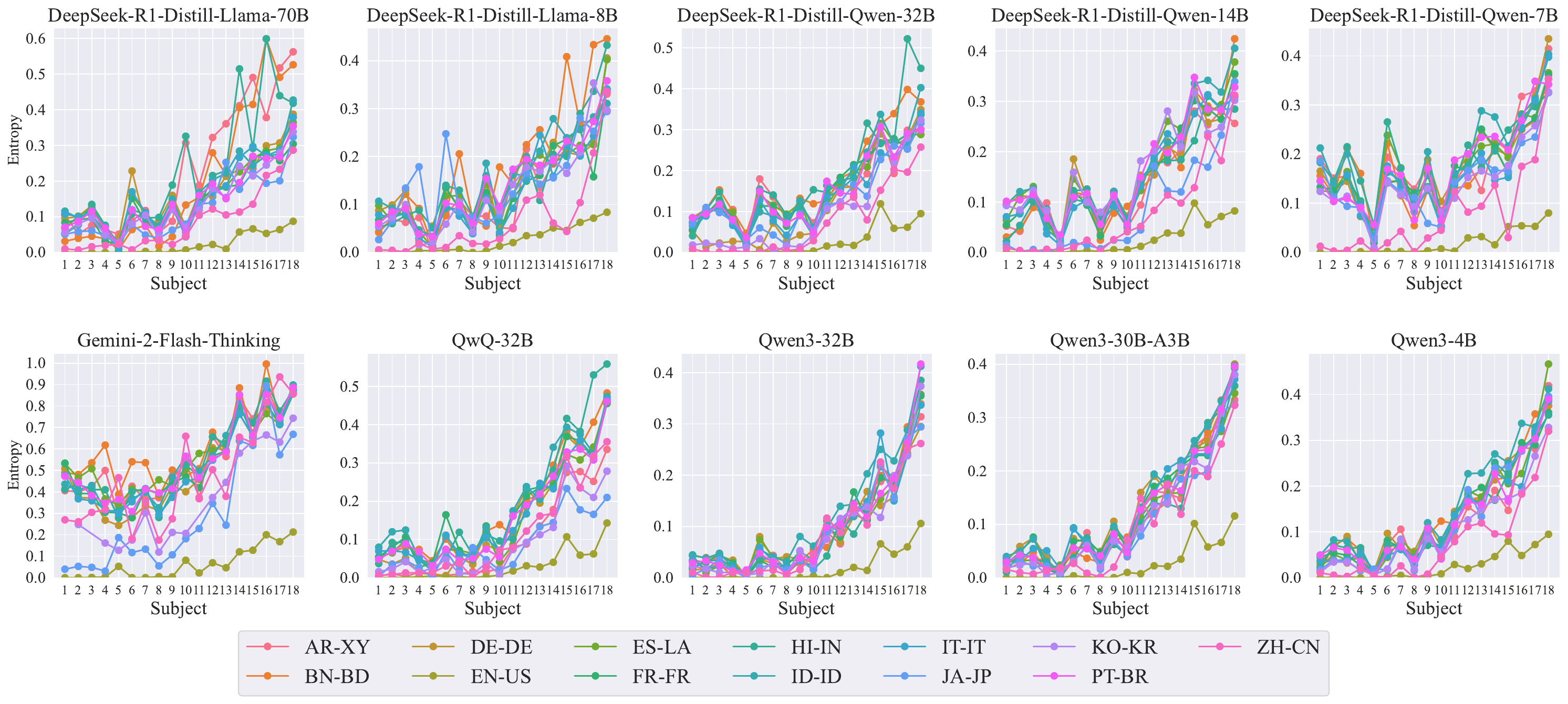}
    \caption{Per-language language mixing entropy across 18 m-MMLU subjects for various reasoning models. Each line represents the entropy trend for a specific input language. STEM subjects (subjects 10–18) consistently exhibit higher entropy across most languages, mirroring the pattern seen in the language-averaged results (Figure~\ref{fig:mmlu-entropy}).}
    \label{fig:mmlu-entropy-per-language}
\end{figure*}

\paragraph{m-MMLU per-language results.}
Figure~\ref{fig:mmlu-entropy-per-language} presents the per-language breakdown of language mixing entropy across m-MMLU subjects, complementing the averaged results in Figure~\ref{fig:mmlu-entropy}. The observed rise in entropy within STEM subjects (subjects 10–18) holds consistently across most languages, supporting the conclusion that STEM subjects tend to elicit more language mixing during reasoning. The specific subject order we use here is: global\_facts, world\_religions, sociology, high\_\\school\_world\_history, moral\_disputes,profession-\\al\_medicine, philosophy, high\_school\_macroeco-\\nomics, management, elementary\_mathematics, 
high\_school\_computer\_science, high\_school\_che-\\mistry, college\_computer\_science, high\_school\_\\physics, college\_chemistry, college\_physics, high\_\\school\_mathematics, college\_mathematics.

\begin{figure*}[h]
\centering
\fontsize{9.2pt}{12pt}\selectfont
\begin{tcolorbox}[colback=blue!5!white, colframe=blue!75!black, title=R1-70B reasoning \textbf{without} script control,
fonttitle=\fontsize{12pt}{12pt}\selectfont,
toptitle=0.2mm,
bottomtitle=0.2mm]
\textbf{Question:}
{\fontsize{6pt}{10pt}\selectfont \RL{جزيرة خاصة جدًا يسكنها فقط الفرسان والمحتالون. الفرسان دائمًا يقولون الحقيقة، والمحتالون دائمًا يكذبون. تلتقي بسكانين: أميليا ويعقوب. قالت أميليا إن يعقوب فارس وإن يعقوب محتال. علق يعقوب قائلاً: "إذا كانت أميليا محتالة، فإن أميليا فارس". فمن هو الفارس ومن هو المحتال؟}}

\vspace{2mm}
(A very special island is inhabited only by knights and knaves. Knights always tell the truth, and knaves always lie. You meet 2 inhabitants: Amelia and Jacob. Amelia said that Jacob is a knight and Jacob is a knave. Jacob commented: 'If Amelia is a knave, then Amelia is a knight.' So who is the knight and who is the knave?)

\vspace{3.5mm}
\textbf{Thinking:}
{\fontsize{6pt}{10pt}\selectfont \RL{حاولت فهم المشكلة التي واجهتني. هناك جزيرة تسكنها فرسان ومحتالون. الفرسان يقولون الحقيقة دائمًا، بينما المحتالون يكذبون دائمًا. التقي باميليا ويعقوب. قالت اميليا أن يعقوب هو فارس ومحتال في الوقت نفسه، وهو ما يبدو متناقضًا. ثم علق يعقوب قائلاً: "إذا كانت اميليا محتالة، فإن اميليا هي فارس".أولاً، ألاحظ أن اميليا تتهامس عن يعقوب بأنه فارس ومحتال، وهذا يبدو مستحيلًا لأن شخصًا لا يمكن أن يكون فارسًا ومحتالًا في الوقت نفسه. ربما كانت تريد التأكيد على أن يعقوب إما فارس أو محتال، أو ربما كانت تحاول الخداع.ثم، قال يعقوب: "إذا كانت اميليا محتالة، فإن اميليا هي فارس". هذا الكلام يبدو غامضًا، خاصة إذا كنا نعلم أن الفرسان يقولون الحقيقة والمحتالون يكذبون. إذا كان يعقوب فارسًا، ف}}
\begin{CJK}{UTF8}{gbsn}\footnotesize他的陈述应为真。如果\end{CJK}
{\fontsize{6pt}{10pt}\selectfont \RL{اميليا كانت محتالة،}}
then according to his statement, 
{\fontsize{6pt}{10pt}\selectfont \RL{اميليا}}
 should be a knight, which is a contradiction. So, if 
 {\fontsize{6pt}{10pt}\selectfont \RL{ يعقوب }}
 is a knight, 
 {\fontsize{6pt}{10pt}\selectfont \RL{اميليا}}
 cannot be a liar, which means 
 {\fontsize{6pt}{10pt}\selectfont \RL{اميليا}}
 must be a knight too, but that contradicts her initial statement. Alternatively, if 
 {\fontsize{6pt}{10pt}\selectfont \RL{ يعقوب }}
 is a liar, then his statement is false. That means that if 
 {\fontsize{6pt}{10pt}\selectfont \RL{اميليا}}
 is a liar, she is not a knight, which is consistent because liars cannot be knights. So, if 
 {\fontsize{6pt}{10pt}\selectfont \RL{ يعقوب }}
 is a liar, 
 {\fontsize{6pt}{10pt}\selectfont \RL{اميليا}}
 must be a knight, but that contradicts her initial statement that 
 {\fontsize{6pt}{10pt}\selectfont \RL{ يعقوب }}
 is both a knight and a liar. Wait, this seems confusing. Maybe I need to break it down step by step.Let me assume 
 {\fontsize{6pt}{10pt}\selectfont \RL{اميليا}}
 is a knight. Then her statements must be true. She says 
 {\fontsize{6pt}{10pt}\selectfont \RL{ يعقوب }}
 is a knight and a liar, which is impossible. So, 
 {\fontsize{6pt}{10pt}\selectfont \RL{اميليا}}
 cannot be a knight, which means she must be a liar. Therefore, her statements are false. So, it's not true that 
 {\fontsize{6pt}{10pt}\selectfont \RL{ يعقوب }}
 is a knight and a liar. That means either he is a knight or a liar, but not both. Now, let's consider 
 {\fontsize{6pt}{10pt}\selectfont \RL{ يعقوب }}
 's statement: "If 
 {\fontsize{6pt}{10pt}\selectfont \RL{اميليا}}
 is a liar, then 
 {\fontsize{6pt}{10pt}\selectfont \RL{اميليا}}
 is a knight." Since we've established that 
 {\fontsize{6pt}{10pt}\selectfont \RL{اميليا}}
 is a liar, the statement becomes: If true, then false. In logic, this is a false statement because a true antecedent leading to a false consequent makes the implication false. Since 
 {\fontsize{6pt}{10pt}\selectfont \RL{ يعقوب }}
 's statement is false, he must be a liar.So, putting it all together: 
 {\fontsize{6pt}{10pt}\selectfont \RL{اميليا}}
 is a liar, and 
 {\fontsize{6pt}{10pt}\selectfont \RL{ يعقوب }}
 is also a liar. But wait, can both be liars? The problem doesn't state that there must be at least one knight, so it's possible."

\vspace{3.5mm}
\textbf{Answer:} (1){\fontsize{6pt}{10pt}\selectfont\RL{ يعقوب هو محتال}} \   (2){\fontsize{6pt}{10pt}\RL{أميليا هي محتالة}}
\end{tcolorbox}

    \caption{Example output from the DeepSeek R1-70B model on a Knights and Knaves logic puzzle. The model processes Arabic input and generates its reasoning without script constraints. The results demonstrate language mixing behavior, with the model combining English, Chinese, and Arabic for reasoning, illustrating the model's natural tendency toward cross-lingual reasoning when solving complex logical problems.}
    \label{fig:script-control-exp-free}
\end{figure*}

\begin{figure*}[h]
\centering
\fontsize{9.2pt}{12pt}\selectfont
\begin{tcolorbox}[colback=blue!5!white, colframe=blue!75!black, title=R1-70B reasoning with script control in \textbf{Latin},
fonttitle=\fontsize{12pt}{12pt}\selectfont,
toptitle=0.2mm,
bottomtitle=0.2mm]
\textbf{Question:}
{\fontsize{6pt}{6pt}\selectfont \RL{جزيرة خاصة جدًا يسكنها فقط الفرسان والمحتالون. الفرسان دائمًا يقولون الحقيقة، والمحتالون دائمًا يكذبون. تلتقي بسكانين: أميليا ويعقوب. قالت أميليا إن يعقوب فارس وإن يعقوب محتال. علق يعقوب قائلاً: "إذا كانت أميليا محتالة، فإن أميليا فارس". فمن هو الفارس ومن هو المحتال؟}}

\vspace{2mm}
(A very special island is inhabited only by knights and knaves. Knights always tell the truth, and knaves always lie. You meet 2 inhabitants: Amelia and Jacob. Amelia said that Jacob is a knight and Jacob is a knave. Jacob commented: 'If Amelia is a knave, then Amelia is a knight.' So who is the knight and who is the knave?)

\vspace{3.5mm}
\textbf{Thinking:}
Alright, so I've got this logic puzzle here about knights and knaves on a special island. Knights always tell the truth, and knaves always lie. I meet two people, Amelia and Jacob. Amelia says that Jacob is a knight and that Jacob is a knave. Then Jacob says, "If Amelia is a knave, then Amelia is a knight." I need to figure out who's who.\textbackslash n\textbackslash nFirst, let me break down the statements. Amelia makes two claims: Jacob is a knight and Jacob is a knave. Wait, that's contradictory. If she's a knight, she can't lie, so both her statements must be true, but saying Jacob is both a knight and a knave can't be true at the same time. So that suggests Amelia might be a knave because she's making contradictory statements.\textbackslash n\textbackslash nBut hold on, if Amelia is a knave, then her statements are lies. So if she says Jacob is a knight, then Jacob must actually be a knave. And when she says Jacob is a knave, that would mean he's actually a knight. Wait, that's also a contradiction. Hmm, maybe I'm approaching this wrong.\textbackslash n\textbackslash nLet me try another angle. Let's assume Amelia is a knight. Then her statements must both be true. But she says Jacob is both a knight and a knave, which is impossible. So Amelia can't be a knight because that leads to a contradiction. Therefore, Amelia must be a knave.\textbackslash n\textbackslash nIf Amelia is a knave, her statements are false. So when she says Jacob is a knight, ... (Omitted here due to character limit) ... Wait, no, if she says J is both a knight and a knave, and she's lying, then J is neither? But that's impossible because everyone is either a knight or a knave.\textbackslash n\textbackslash nI think I'm getting tangled up here. Let me try again. If Amelia is a knave, her statements are false. So when she says Jacob is a knight, it's false, meaning Jacob is a knave. And when she says Jacob is a knave, it's false, meaning Jacob is a knight. Contradiction again.\textbackslash n\textbackslash nWait, maybe the only way this works is if Amelia is a knave and Jacob is a knave. Because if Jacob is a knave, his statement is a lie. His statement is "If Amelia is a knave, then Amelia is a knight," which is ¬A → A. Since Amelia is a knave (¬A), the implication is false, which makes Jacob's statement false, so he's a knave. That fits.\textbackslash n\textbackslash nSo, Amelia is a knave, and Jacob is a knave. But wait, earlier I thought Jacob had to be a knight. I'm confused.\textbackslash n\textbackslash nWait, no. If Amelia is a knave, her statements are false, so Jacob is not a knight and not a knave? That can't be. So maybe I need to accept that Jacob is a knave, and his statement is a lie, which makes sense because as a knave, he lies.\textbackslash n\textbackslash nSo, putting it all together: Amelia is a knave, Jacob is a knave. But does that satisfy all statements? Let's check.\textbackslash n\textbackslash nAmelia says Jacob is a knight (lie, so Jacob is a knave) and Jacob is a knave (lie, so Jacob is a knight). Contradiction again. Hmm, I'm stuck.\textbackslash n\textbackslash nMaybe I need to consider that both are knaves. If both are knaves, then Amelia's statements are lies, so Jacob is not a knight and not a knave, which is impossible. So that can't be.\textbackslash n\textbackslash nWait, perhaps the only solution is that Amelia is a knave and Jacob is a knight. Let's see:\textbackslash n\textbackslash nAmelia is a knave, so her statements are false. She says Jacob is a knight (false, so Jacob is a knave) and Jacob is a knave (false, so Jacob is a knight). Contradiction again.\textbackslash n\textbackslash nI'm going in circles. Maybe the answer is that Amelia is a knave and Jacob is a knave. Even though it seems contradictory, it's the only way to make their statements false.\textbackslash n\textbackslash nWait, no, because if Jacob is a knave, his statement is a lie, which fits. And Amelia's statements being lies fit if Jacob is a knave. So maybe that's the answer.\textbackslash n

\vspace{3.5mm}
\textbf{Answer:} (1){\fontsize{6pt}{6pt}\selectfont\RL{ يعقوب هو محتال}} \   (2){\fontsize{6pt}{6pt}\RL{أميليا هي محتالة}}
\end{tcolorbox}

    \caption{Example output from the DeepSeek R1-70B model on a Knights and Knaves logic puzzle. The model processes Arabic input, generates its reasoning in English via Latin script constrained decoding, and produces the final answer in Arabic without script constraints, demonstrating cross-lingual reasoning with selective script control.}
    \label{fig:script-control-exp-latin}
\end{figure*}

\begin{figure*}[h]
\centering
\fontsize{9.2pt}{12pt}\selectfont
\begin{tcolorbox}[colback=blue!5!white, colframe=blue!75!black, title=R1-70B reasoning with script control in \textbf{Han},
fonttitle=\fontsize{12pt}{12pt}\selectfont,
toptitle=0.2mm,
bottomtitle=0.2mm,
]
\textbf{Question:}
{\fontsize{6pt}{6pt}\selectfont \RL{جزيرة خاصة جدًا يسكنها فقط الفرسان والمحتالون. الفرسان دائمًا يقولون الحقيقة، والمحتالون دائمًا يكذبون. تلتقي بسكانين: أميليا ويعقوب. قالت أميليا إن يعقوب فارس وإن يعقوب محتال. علق يعقوب قائلاً: "إذا كانت أميليا محتالة، فإن أميليا فارس". فمن هو الفارس ومن هو المحتال؟}}

\vspace{2mm}
(A very special island is inhabited only by knights and knaves. Knights always tell the truth, and knaves always lie. You meet 2 inhabitants: Amelia and Jacob. Amelia said that Jacob is a knight and Jacob is a knave. Jacob commented: 'If Amelia is a knave, then Amelia is a knight.' So who is the knight and who is the knave?)

\vspace{3.5mm}
\textbf{Thinking:}
\begin{CJK}{UTF8}{gbsn}\footnotesize 嗯，这个问题看起来有点复杂，但我会慢慢理清楚。首先，我知道在骑士与无赖的问题中，骑士总是说真话，而无赖总是撒谎。现在有两个居民，阿梅莉亚和雅各布，他们各自发表了一些言论，我需要根据这些言论来判断谁是骑士，谁是无赖。\textbackslash n\textbackslash n首先，阿梅莉亚说雅各布是骑士。而雅各布则说：“如果阿梅莉亚是无赖，那么阿梅莉亚就是骑士。”这听起来有点绕，所以我需要一步步分析。\textbackslash n\textbackslash n假设阿梅莉亚是骑士，那么她的话就是真的，也就是说雅各布确实是骑士。那么雅各布作为骑士，他的话也必须是真的。雅各布的话是一个条件句：“如果阿梅莉亚是无赖，那么阿梅莉亚就是骑士。”这里有点矛盾，因为如果阿梅莉亚是骑士，那么前提“阿梅莉亚是无赖”就是假的。在逻辑学中，一个假命题作为条件句的前件，整个条件句就是真的，无论后件如何。所以雅各布的话在这种情况下是正确的，这与他是骑士的身份一致。所以这个假设下，两人都为骑士，但这样阿梅莉亚说雅各布是骑士是正确的，而雅各布的话也成立，所以这可能是一个可能的解。\ldots \textbackslash n\textbackslash n或者，可能只有当阿梅莉亚是无赖时，雅各布的话导致矛盾，因为如果阿梅莉亚是无赖，雅各布作为无赖，他的话是假的，所以“如果阿梅莉亚是无赖，那么阿梅莉亚就是骑士”为假，这意味着阿梅莉亚是无赖且阿梅莉亚不是骑士，这成立，所以两人都为无赖是可能的。\textbackslash n\textbackslash n但问题可能在于，如果两人都为无赖，那么阿梅莉亚说雅各布是骑士是假的，而雅各布作为无赖，他的条件句为假，这也成立，所以可能有两种解，但通常问题只有一个解，可能需要重新审视。\textbackslash n\textbackslash n或者，可能问题在于，如果雅各布是无赖，他的话为假，所以“如果阿梅莉亚是无赖，那么阿梅莉亚就是骑士”为假，这意味着阿梅莉亚是无赖且阿梅莉亚不是骑士，这成立，所以两人都为无赖是可能的。\textbackslash n\textbackslash n但通常，这类问题只有一个解，可能我需要重新考虑。\textbackslash n\textbackslash n另一种方法是使用真值表来分析。设A为阿梅莉亚是骑士，J为雅各布是骑士。那么：\textbackslash n\textbackslash nA → J（阿梅莉亚的话）\textbackslash nJ → (¬A → A)（雅各布的话，因为他说的是“如果A是无赖（¬A），那么A是骑士（A）”）\textbackslash n\textbackslash n现在，我们需要找出A和J的可能组合。\textbackslash n\textbackslash n情况1：A=真，J=真\textbackslash n阿梅莉亚的话：真→真，成立。\textbackslash n雅各布的话：真→（¬真→真）= 真→（假→真）= 真→真，成立。\textbackslash n所以情况1成立。\textbackslash n\textbackslash n情况2：A=真，J=假\textbackslash n阿梅莉亚的话：真→假，不成立，所以排除。\textbackslash n\textbackslash n情况3：A=假，J=真\textbackslash n阿梅莉亚的话：假→真，成立（因为假命题可以推出任何结论）。\textbackslash n雅各布的话：真→（¬假→假）= 真→（真→假）= 真→假，不成立，所以排除。\textbackslash n\textbackslash n情况4：A=假，J=假\textbackslash n阿梅莉亚的话：假→假，成立。\textbackslash n雅各布的话：假→（¬假→假）= 假→（真→假）= 假→假，这在逻辑中，当J为假时，雅各布的话为假，所以雅各布作为无赖，他的话应该是假的，所以这成立。\textbackslash n\textbackslash n因此，情况1和情况4都成立，这意味着有两种可能的解：两人都为骑士，或者两人都为无赖。但通常这类问题只有一个解，可能需要重新审视。\textbackslash n\textbackslash n或者，可能问题在于，如果两人都为无赖，那么雅各布的话为假，而“如果阿梅莉亚是无赖，那么阿梅莉亚就是骑士”为假，这意味着阿梅莉亚是无赖且阿梅莉亚不是骑士，这成立，所以两人都为无赖是可能的。\ldots另一种方法是考虑雅各布的话的逻辑结构。雅各布说：“如果阿梅莉亚是无赖，那么阿梅莉亚就是骑士。”这相当于说“¬A → A”，也就是“如果¬A，则A”。这在逻辑上是矛盾的，因为如果¬A为真，那么A必须为真，这导致矛盾，因为A不能同时为真和假。所以，雅各布的话在逻辑上是矛盾的，因此他的话只能是假的，所以雅各布必须是无赖。\textbackslash n\textbackslash n既然雅各布是无赖，那么他的话是假的，所以“¬A → A”是假的，这意味着¬A为真且A为假，即阿梅莉亚是无赖。因此，阿梅莉亚是无赖，雅各布也是无赖。\textbackslash n\textbackslash n但这样，阿梅莉亚作为无赖，她说雅各布是骑士是假的，所以雅各布是无赖，这与我们之前的结论一致。\textbackslash n\textbackslash n所以，最终结论是，阿梅莉亚和雅各布都是无赖。\end{CJK}

\vspace{3.5mm}
\textbf{Answer:} (1){\fontsize{6pt}{6pt}\selectfont\RL{ يعقوب هو محتال}} \   (2){\fontsize{6pt}{6pt}\RL{أميليا هي محتالة}}
\end{tcolorbox}

    \caption{Example output from the DeepSeek R1-70B model on a Knights and Knaves logic puzzle. The model processes Arabic input, generates its reasoning in Chinese via Han script constrained decoding, and produces the final answer in Arabic without script constraints, demonstrating cross-lingual reasoning with selective script control.}
    \label{fig:script-control-exp-han}
\end{figure*}

\begin{figure*}[h]
\centering
\fontsize{9.2pt}{12pt}\selectfont
\begin{tcolorbox}[colback=blue!5!white, colframe=blue!75!black, title=R1-70B reasoning with script control in \textbf{Arabic},
fonttitle=\fontsize{12pt}{12pt}\selectfont,
toptitle=0.2mm,
bottomtitle=0.2mm]
\textbf{Question:}
{\fontsize{6pt}{6pt}\selectfont \RL{جزيرة خاصة جدًا يسكنها فقط الفرسان والمحتالون. الفرسان دائمًا يقولون الحقيقة، والمحتالون دائمًا يكذبون. تلتقي بسكانين: أميليا ويعقوب. قالت أميليا إن يعقوب فارس وإن يعقوب محتال. علق يعقوب قائلاً: "إذا كانت أميليا محتالة، فإن أميليا فارس". فمن هو الفارس ومن هو المحتال؟}}

\vspace{2mm}
(A very special island is inhabited only by knights and knaves. Knights always tell the truth, and knaves always lie. You meet 2 inhabitants: Amelia and Jacob. Amelia said that Jacob is a knight and Jacob is a knave. Jacob commented: 'If Amelia is a knave, then Amelia is a knight.' So who is the knight and who is the knave?)

\vspace{3.5mm}
\textbf{Thinking:}
{\fontsize{6pt}{6pt}\selectfont \RL{
أولاً، سأبدأ بتحليل البيانات التي زودتني بها. هناك جزيرة يسكنها فرسان ومحتالون. الفرسان دائمًا يقولون الحقيقة، بينما المحتالون دائمًا يكذبون.التقيتم باميليا ويعقوب. قالت اميليا إن يعقوب فارس وإن يعقوب محتال. هذا البيان يبدو متناقضًا على أول نظر، مما يجعله مثيرًا للاهتمام. لذلك سأتحليله بدقة.ثم علق يعقوب قائلاً: "إذا كانت أميليا محتالة، فإن أميليا فارس". هذا الكلام يحتوي على تعليمات شرطية، حيث يربط بين كون اميليا محتالة وكونها فارسًا.سأبدأ بفحص كلام اميليا. قالت اميليا أن يعقوب فارس وإن يعقوب محتال. هذا يعني أن اميليا تتهم يعقوب بكونهم فارسًا ومحتالًا في الوقت نفسه، وهو ما يبدو مستحيلًا لأن شخصًا لا يمكن أن يكون فارسًا ومحتالًا في آن واحد. هذا يثير الشك في مصداقية اميليا.إذا كانت اميليا فارسًا، فإنها تقول الحقيقة، وبالتالي يجب أن يكون يعقوب فارسًا ومحتالًا في الوقت نفسه، وهو ما لا يمكن أن يكون. لذلك، هذا يؤكد أن اميليا لا يمكن أن تكون فارسًا، وبالتالي هي محتالة.بما أن اميليا محتالة، فإن تعليماتها الكاذبة تُظهر أن يعقوب ليس بالفارس ولا بالمحتال المزعومين. لكننا نعلم أن الجزيرة يسكنها فقط فرسان ومحتالون، لذلك يجب أن يكون يعقوب فارسًا.لتحديد ذلك بشكل أكثر دقة، سأستخدم المنطق الافتراضي. إذا كانت اميليا محتالة، فإن كلامها كاذب، وبالتالي يعقوب ليس فارسًا أو محتالًا، ولكن هذا يتعارض مع حقيقة أن الجزيرة يسكنها فقط هذان النوعان. لذلك، يجب أن يكون يعقوب فارسًا.أما تعليق يعقوب، فإن كلامه يبدو متناقضًا إذا كنا نعتقد أن اميليا محتالة. لكن بمجرد تحديد أن اميليا محتالة، يصبح كلام يعقوب منطقيًا، حيث إنه كفارس يقول الحقيقة.}}

\vspace{3.5mm}
\textbf{Answer:} (1){\fontsize{6pt}{6pt}\selectfont\RL{يعقوب هو فارس}} \   (2){\fontsize{6pt}{6pt}\RL{اميليا هي محتالة}}
\end{tcolorbox}

    \caption{Example output from the DeepSeek R1-70B model on a Knights and Knaves logic puzzle. The model processes Arabic input, generates its reasoning in Arabic via Arabic script constrained decoding, and produces the final answer in Arabic without script constraints, demonstrating cross-lingual reasoning with selective script control.}
    \label{fig:script-control-exp-arabic}
\end{figure*}

\begin{figure*}[h]
    \centering
    \includegraphics[width=\textwidth]{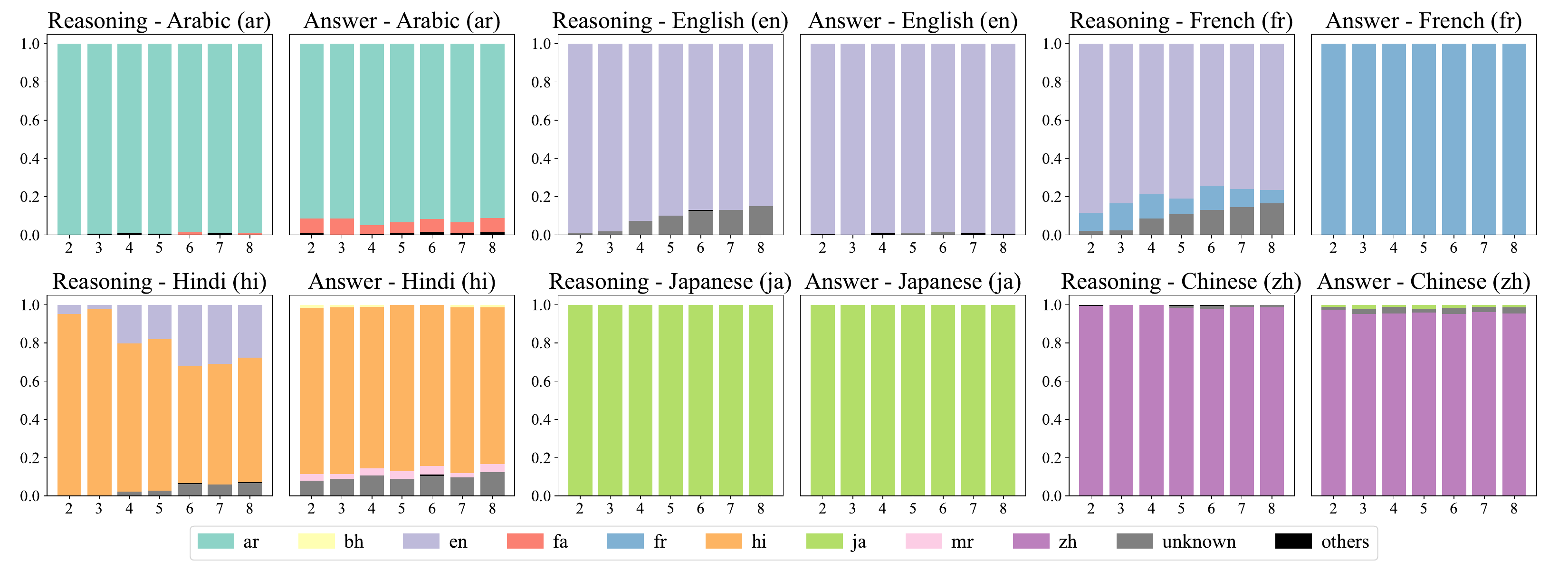}
    \caption{Language composition for \texttt{R1} (K\&K).}
    \label{fig:r1_kk}
\end{figure*}

\begin{figure*}[h]
    \centering
    \includegraphics[width=\textwidth]{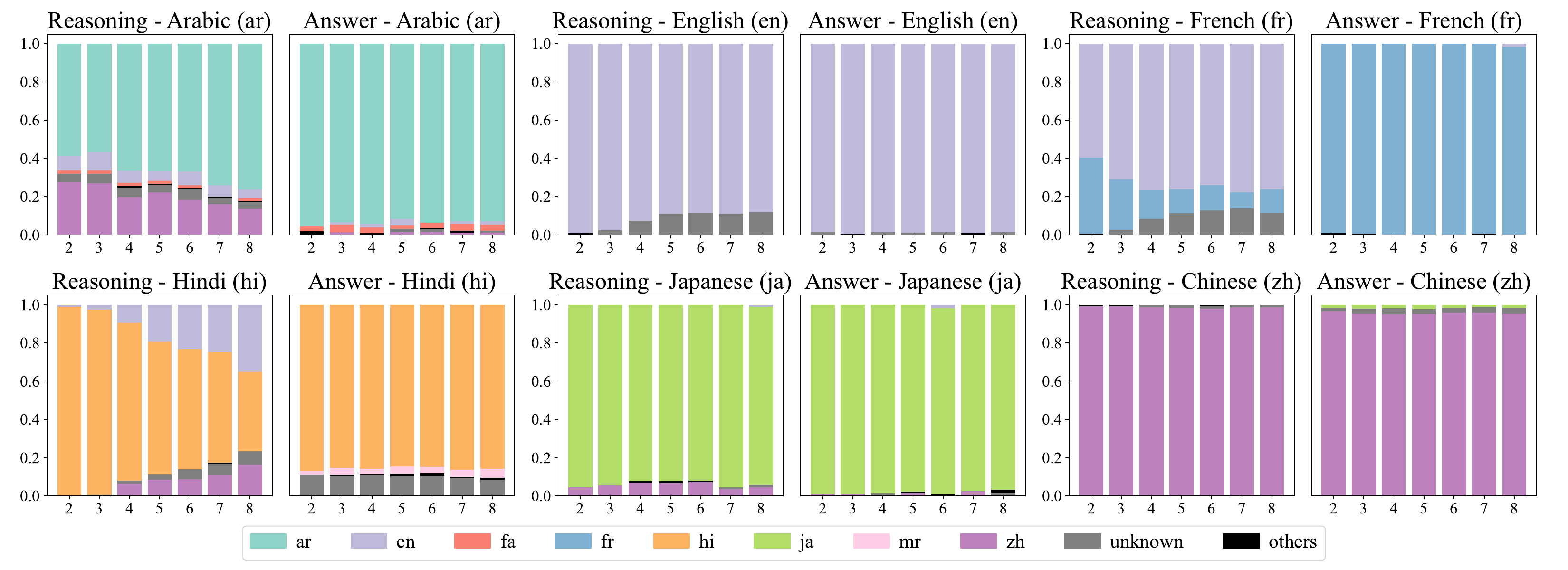}
    \caption{Language composition for \texttt{R1-70B} (K\&K).}
    \label{fig:r1_70b_kk}
\end{figure*}

\begin{figure*}[h]
    \centering
    \includegraphics[width=\textwidth]{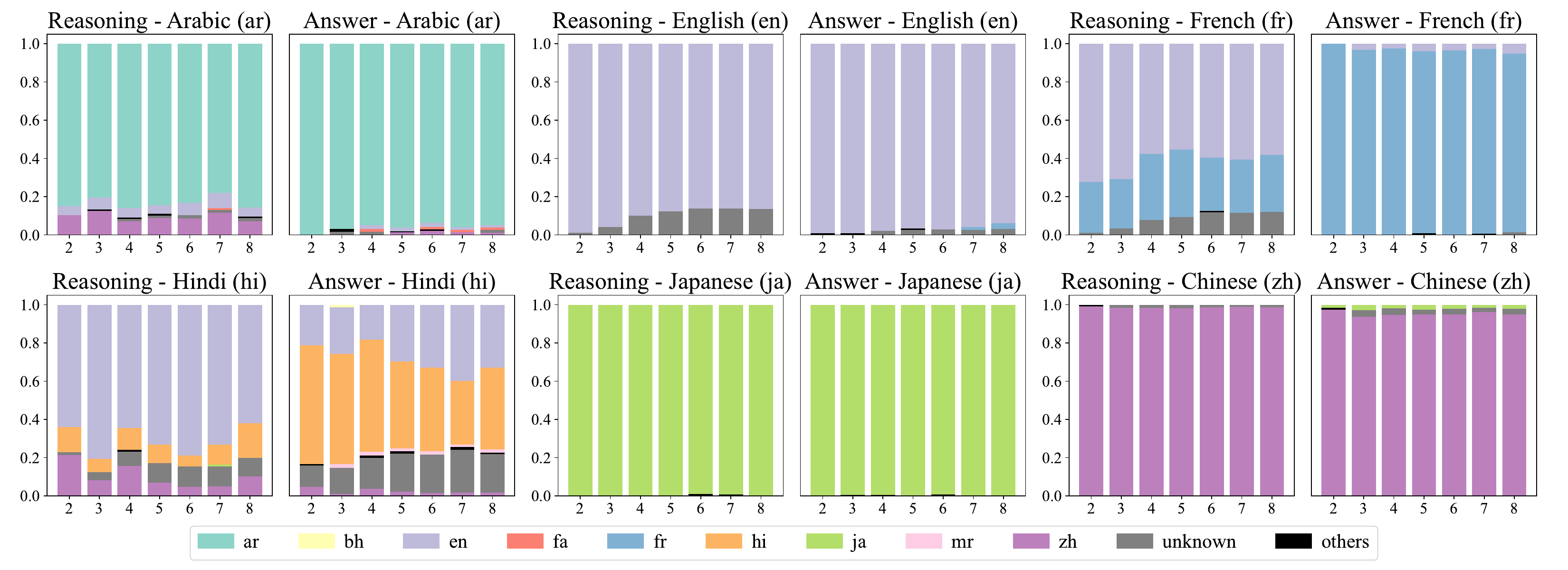}
    \caption{Language composition for \texttt{R1-8B} (K\&K).}
    \label{fig:r1_8b_kk}
\end{figure*}

\begin{figure*}[h]
    \centering
    \includegraphics[width=\textwidth]{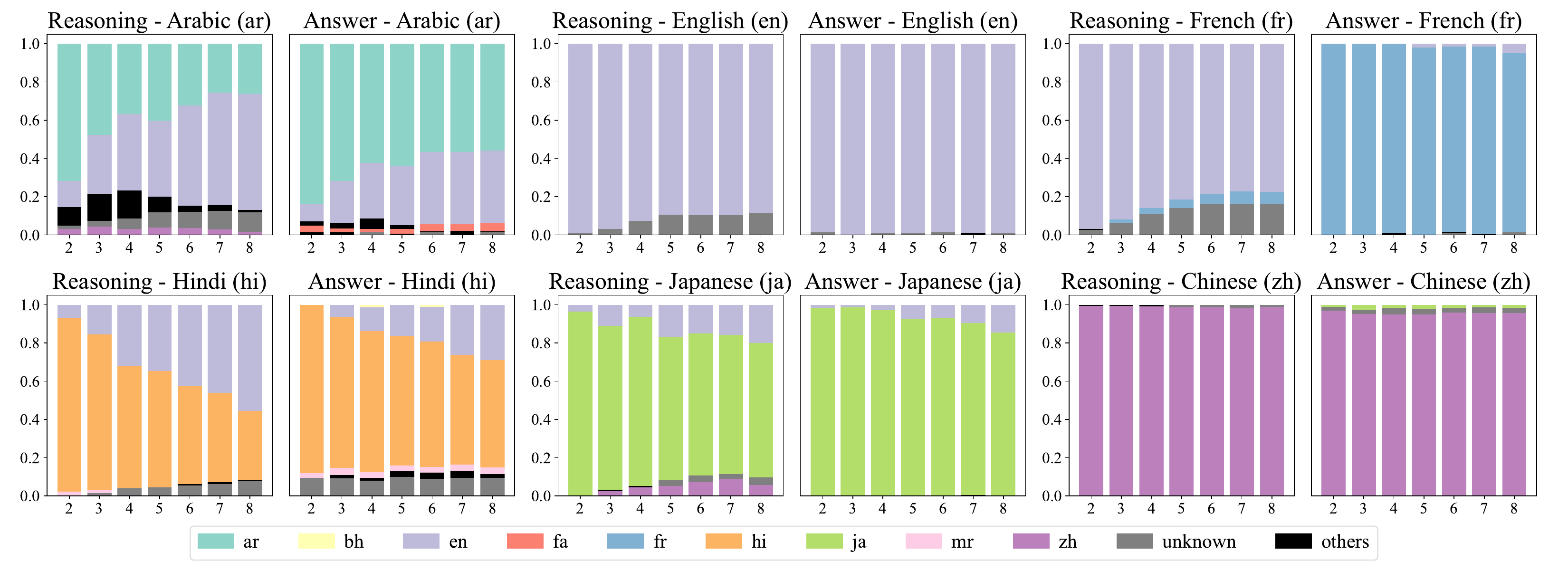}
    \caption{Language composition for \texttt{R1-32B} (K\&K).}
    \label{fig:r1_32b_kk}
\end{figure*}

\begin{figure*}[h]
    \centering
    \includegraphics[width=\textwidth]{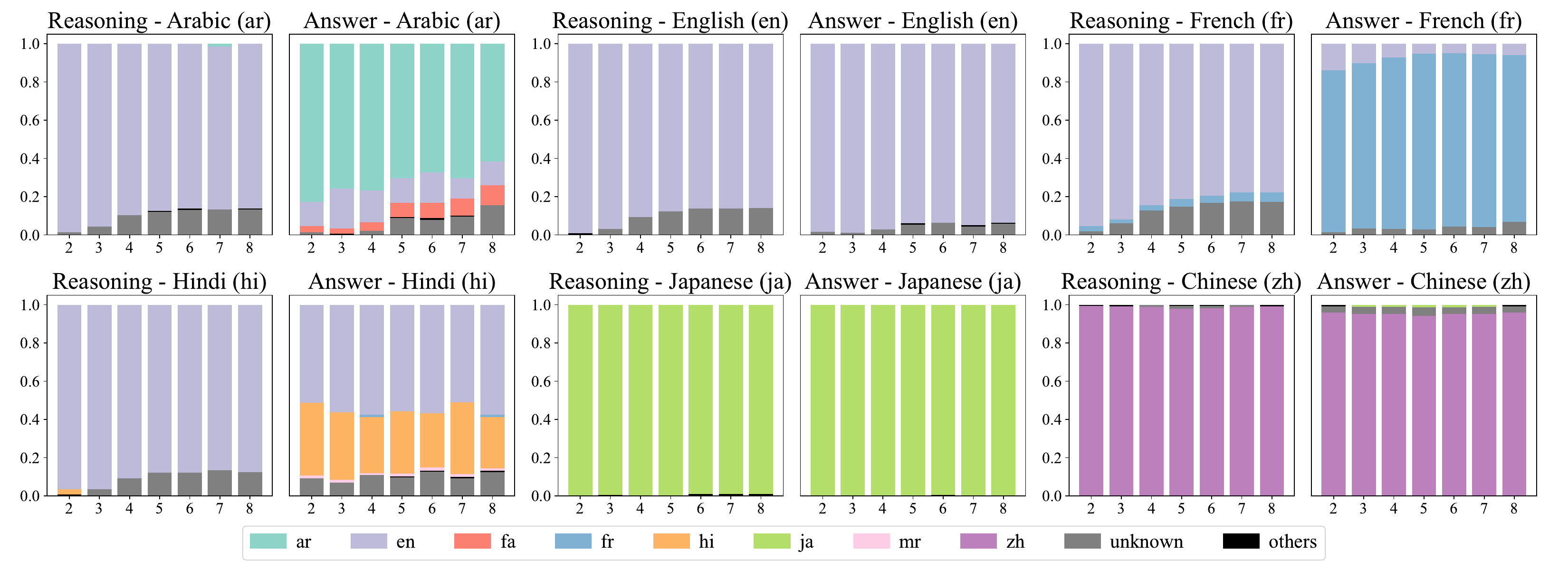}
    \caption{Language composition for \texttt{R1-14B} (K\&K).}
    \label{fig:r1_14b_kk}
\end{figure*}

\begin{figure*}[h]
    \centering
    \includegraphics[width=\textwidth]{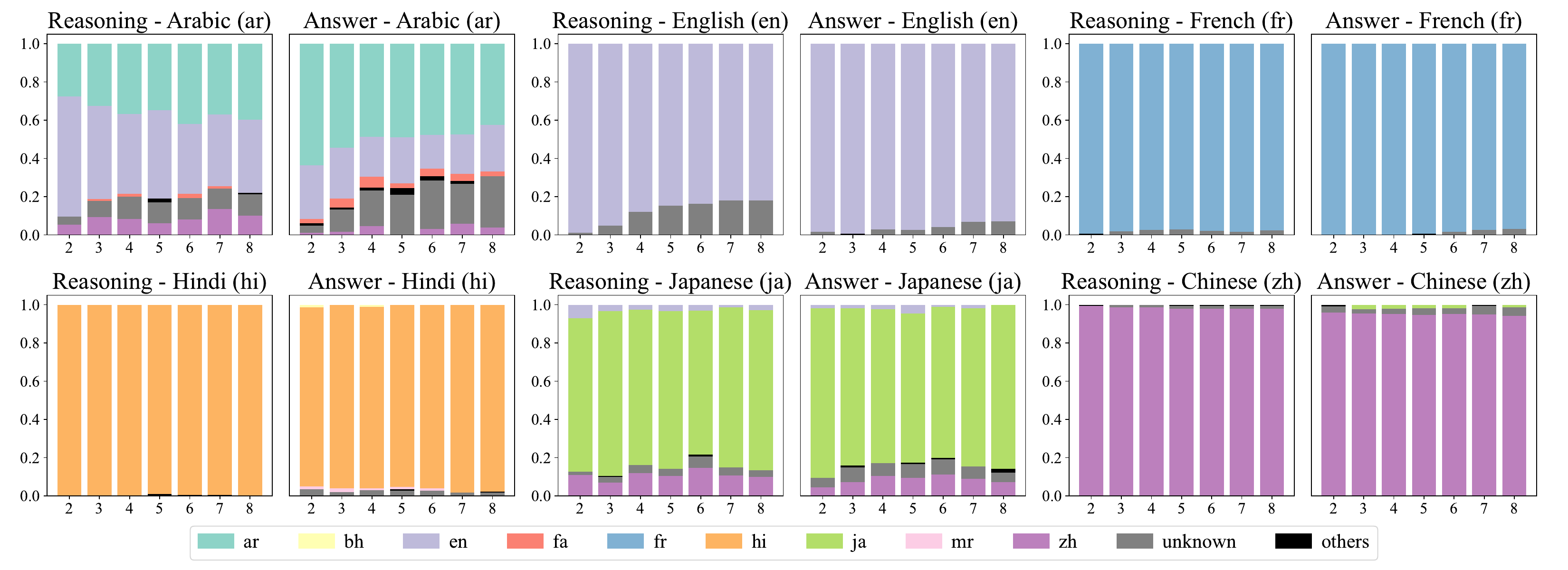}
    \caption{Language composition for \texttt{R1-7B} (K\&K).}
    \label{fig:r1_7b_kk}
\end{figure*}

\begin{figure*}[h]
    \centering
    \includegraphics[width=\textwidth]{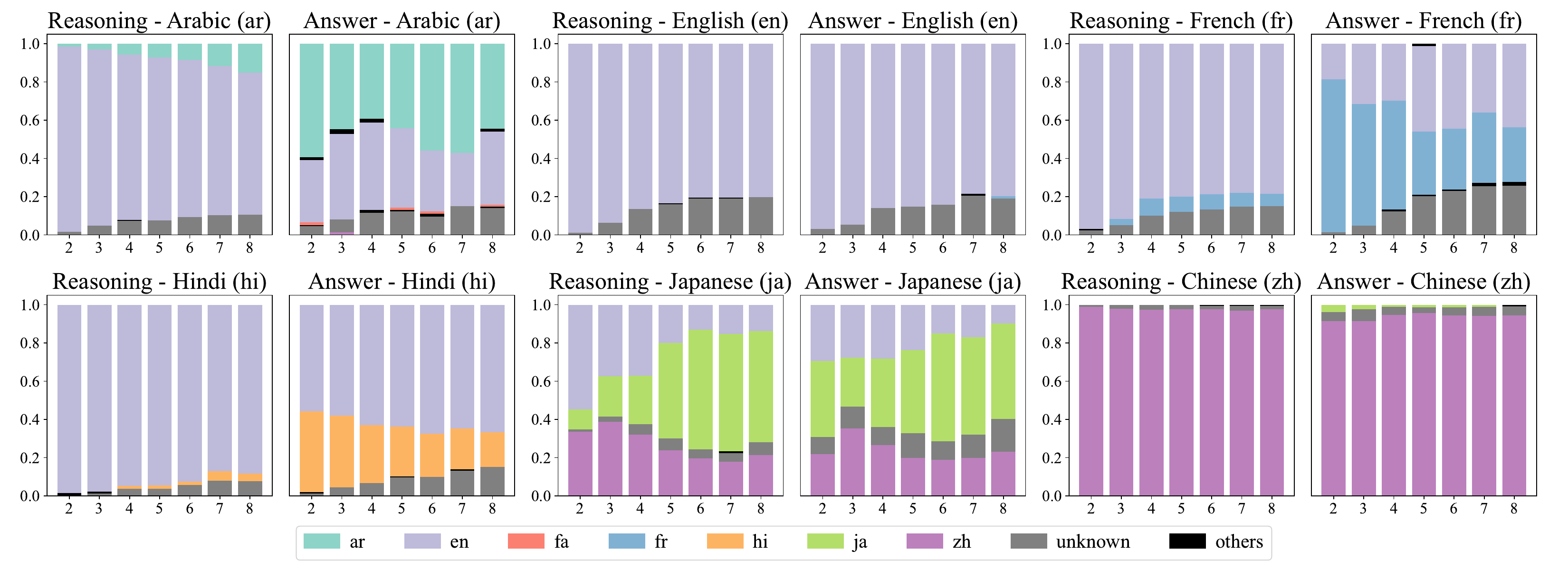}
    \caption{Language composition for \texttt{R1-1.5B} (K\&K).}
    \label{fig:r1_1_5b_kk}
\end{figure*}

\begin{figure*}[h]
    \centering
    \includegraphics[width=\textwidth]{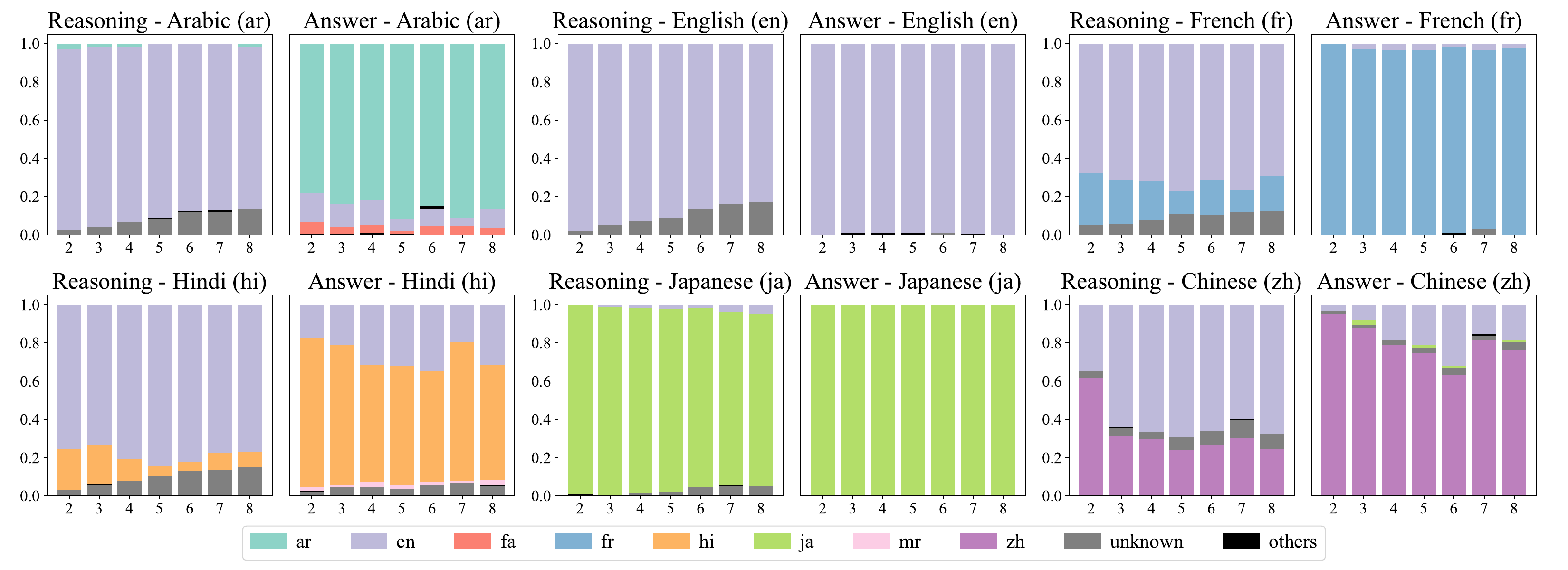}
    \caption{Language composition for \texttt{Gemini} (K\&K).}
    \label{fig:gemini_kk}
\end{figure*}

\begin{figure*}[h]
    \centering
    \includegraphics[width=\textwidth]{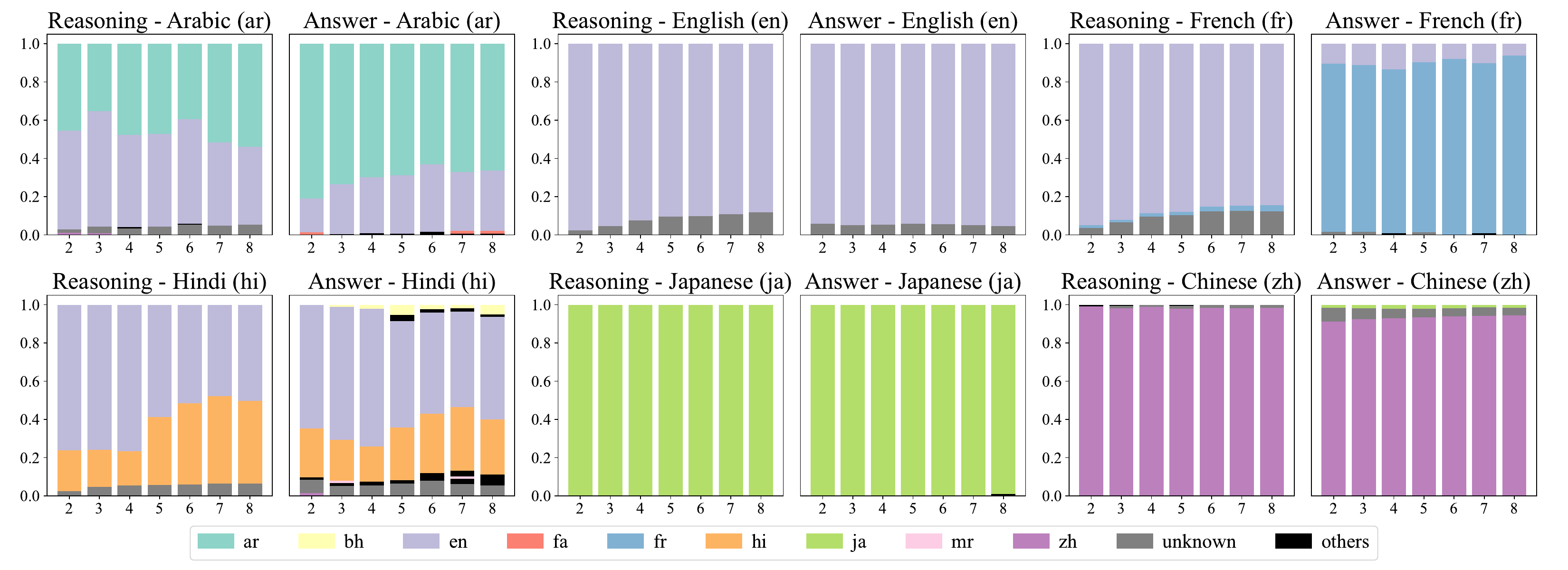}
    \caption{Language composition for \texttt{QwQ-32B} (K\&K).}
    \label{fig:qwq_32b_kk}
\end{figure*}

\begin{figure*}[h]
    \centering
    \includegraphics[width=\textwidth]{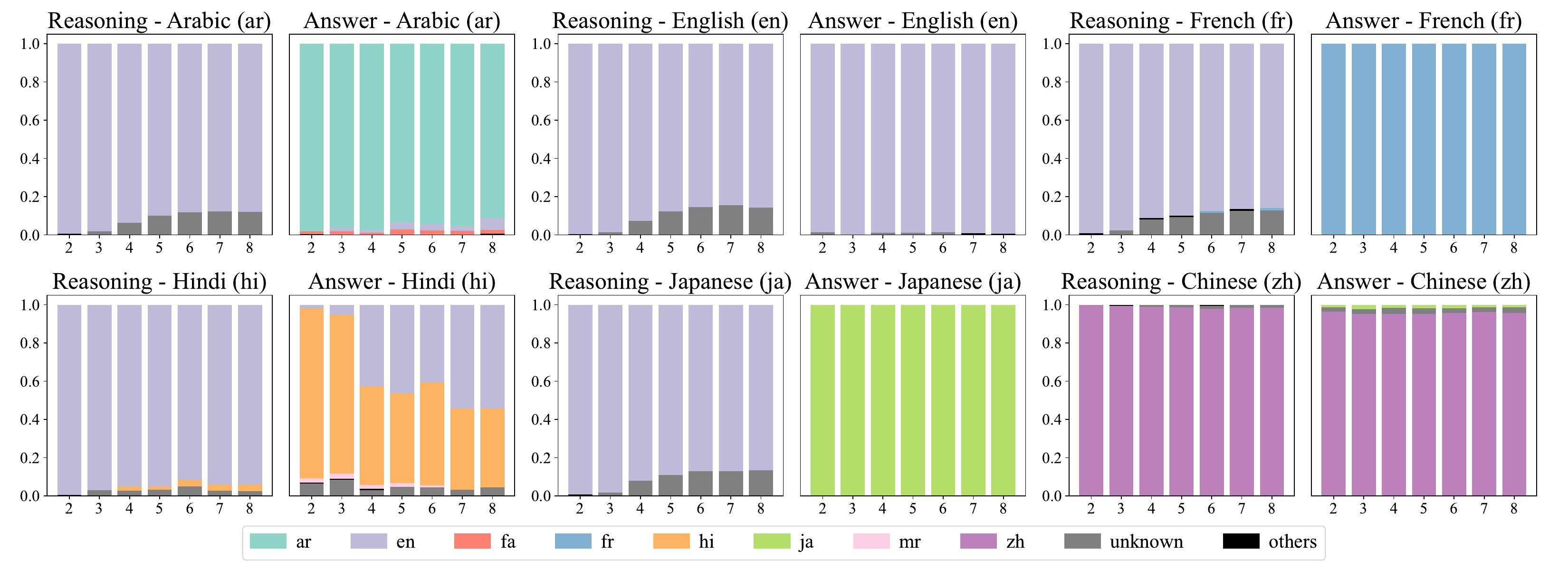}
    \caption{Language composition for \texttt{Qwen3-32B} (K\&K).}
    \label{fig:qwen3_32b_kk}
\end{figure*}

\begin{figure*}[h]
    \centering
    \includegraphics[width=\textwidth]{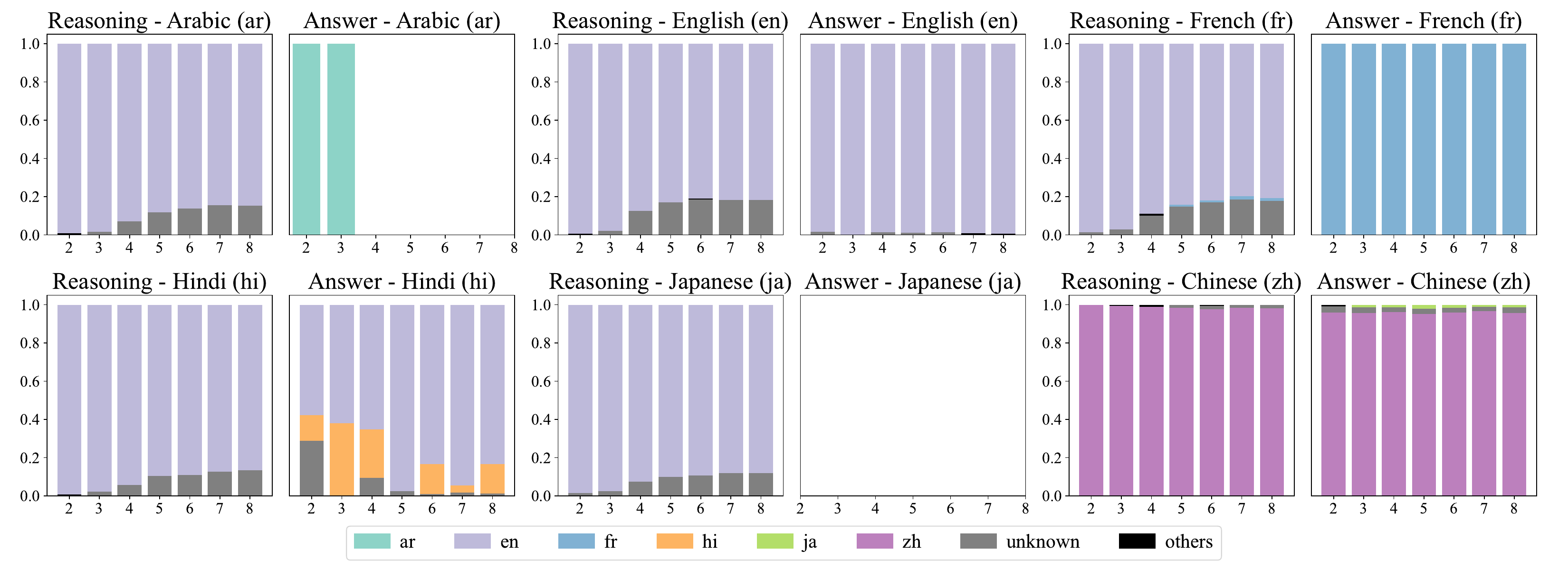}
    \caption{Language composition for \texttt{Qwen3-30B-A3B} (K\&K).}
    \label{fig:qwen3_30b_a3b_kk}
\end{figure*}

\begin{figure*}[h]
    \centering
    \includegraphics[width=\textwidth]{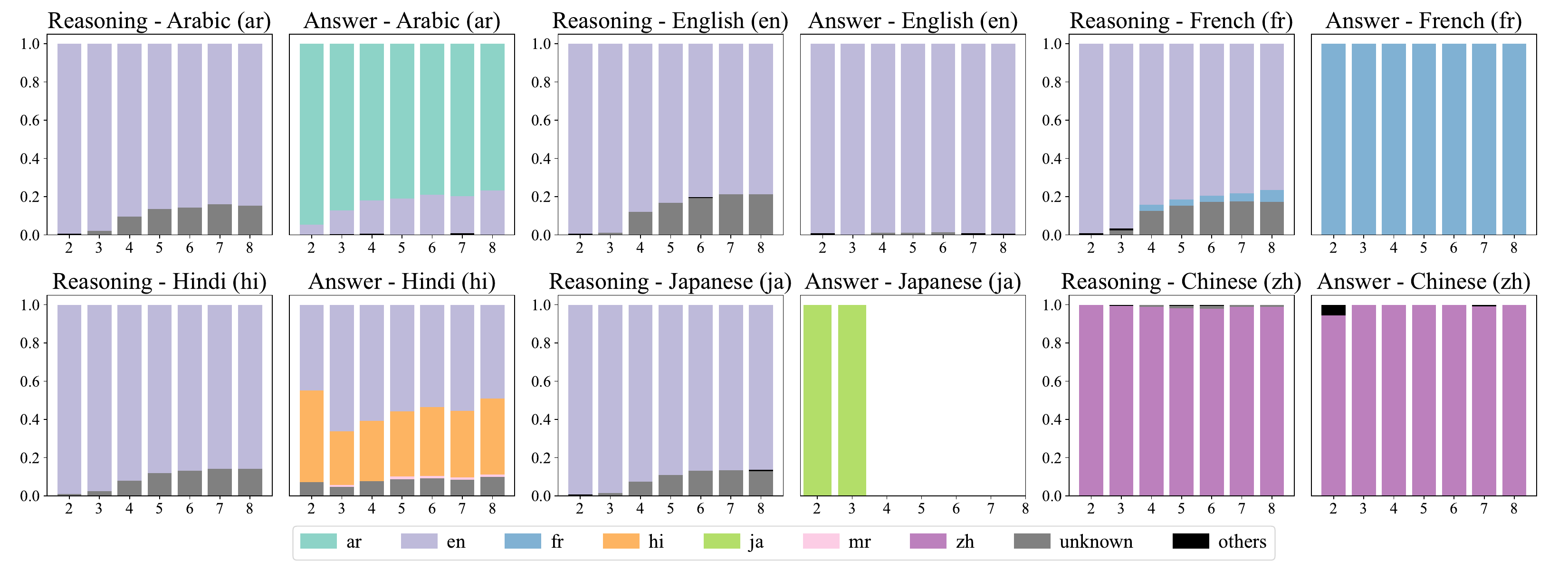}
    \caption{Language composition for \texttt{Qwen3-4B} (K\&K).}
    \label{fig:qwen3_4b_kk}
\end{figure*}

\begin{figure*}[h]
    \centering
    \includegraphics[width=\textwidth]{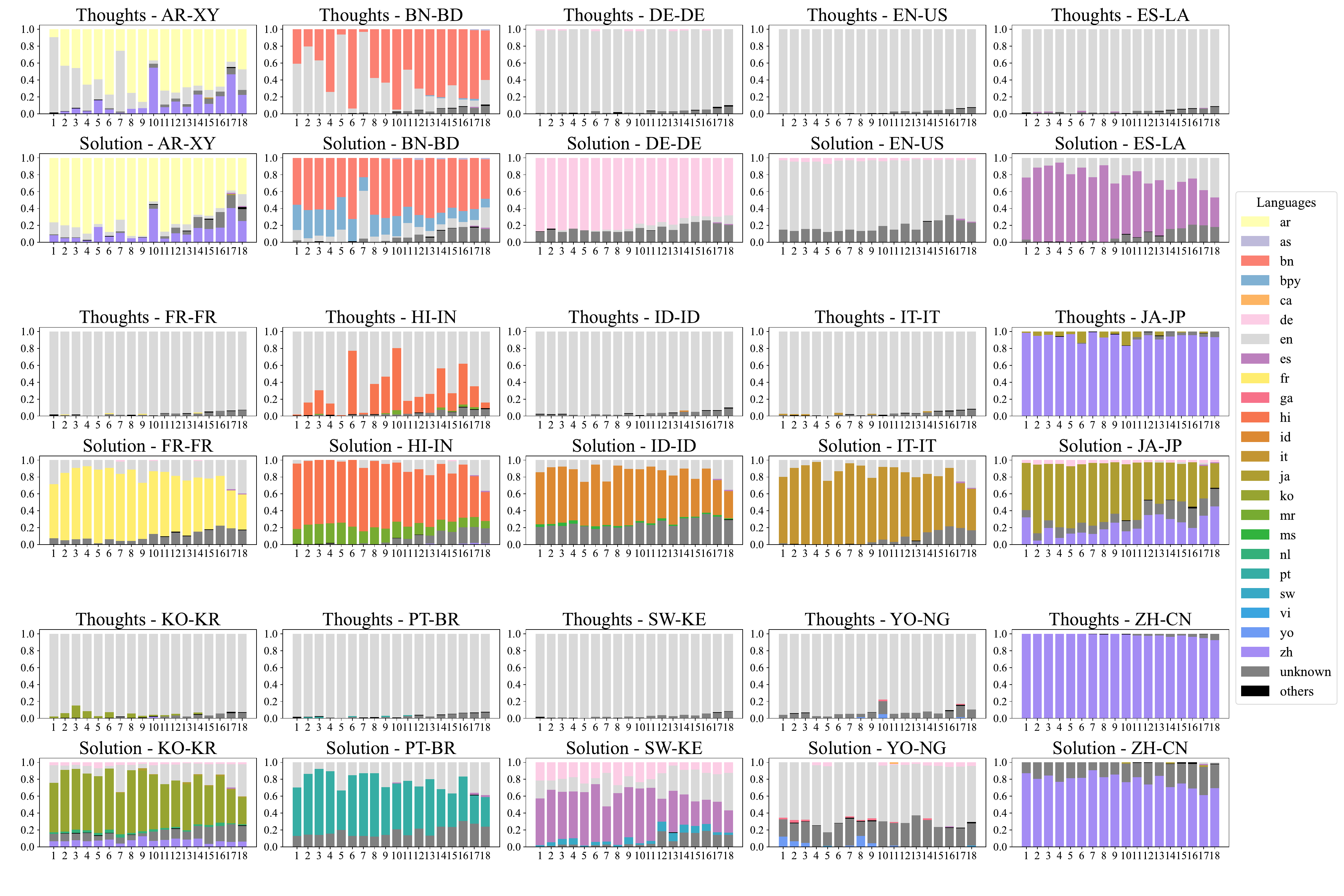}
    \caption{Language composition for \texttt{R1-70B} (m-MMLU).}
    \label{fig:r1_70b_mmlu}
\end{figure*}

\begin{figure*}[h]
    \centering
    \includegraphics[width=\textwidth]{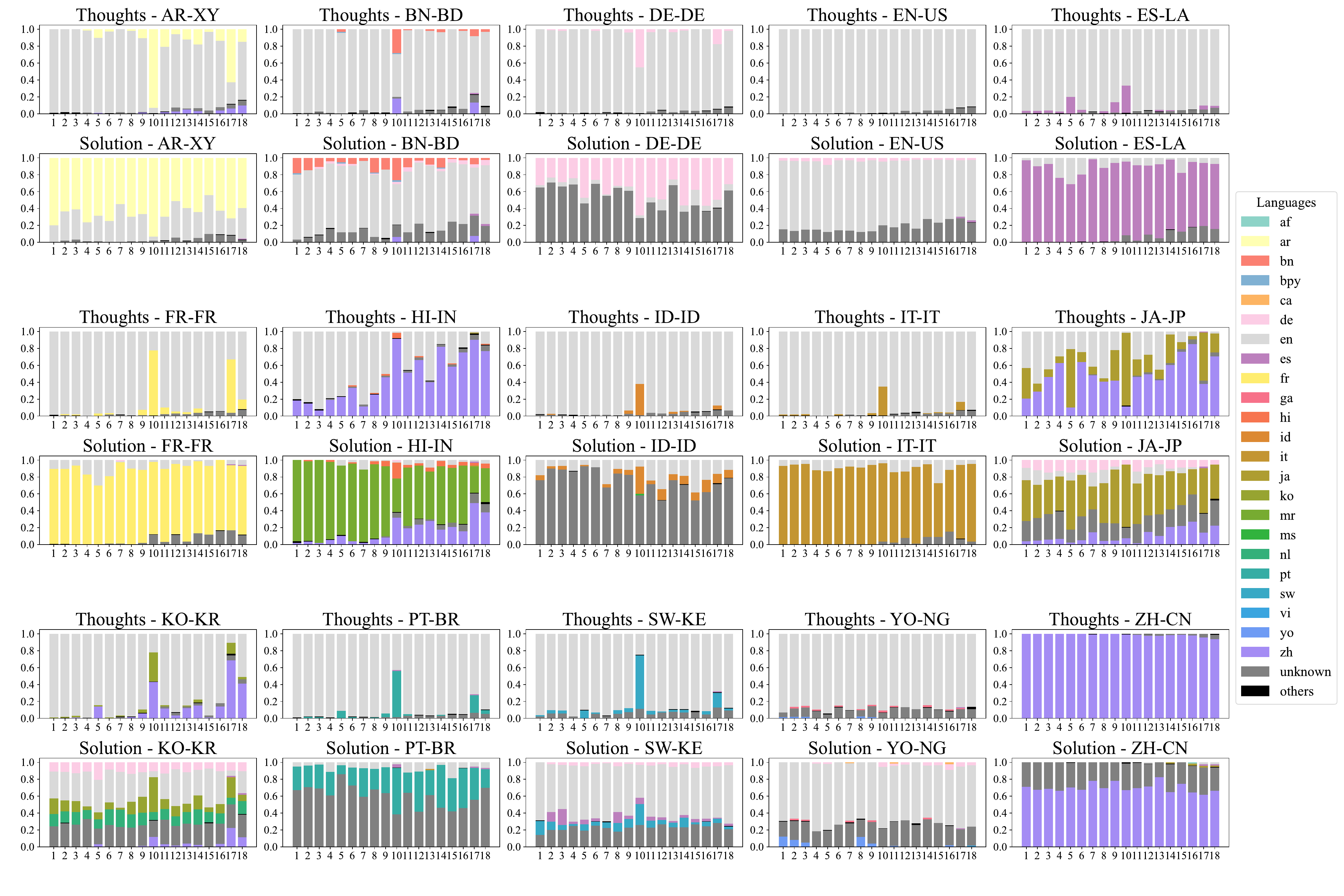}
    \caption{Language composition for \texttt{R1-8B} (m-MMLU).}
    \label{fig:r1_8b_mmlu}
\end{figure*}

\begin{figure*}[h]
    \centering
    \includegraphics[width=\textwidth]{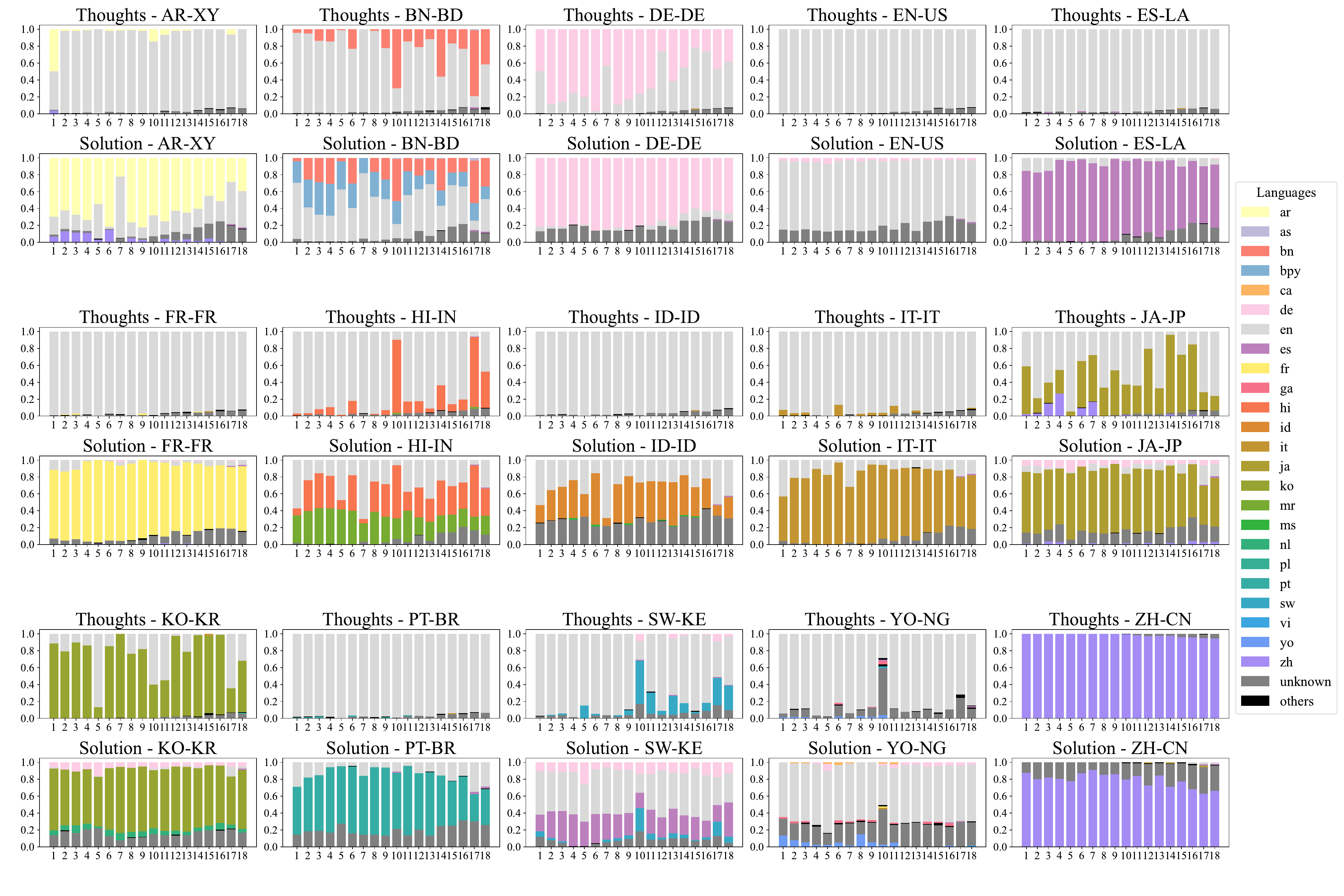}
    \caption{Language composition for \texttt{R1-32B} (m-MMLU).}
    \label{fig:r1_32b_mmlu}
\end{figure*}

\begin{figure*}[h]
    \centering
    \includegraphics[width=\textwidth]{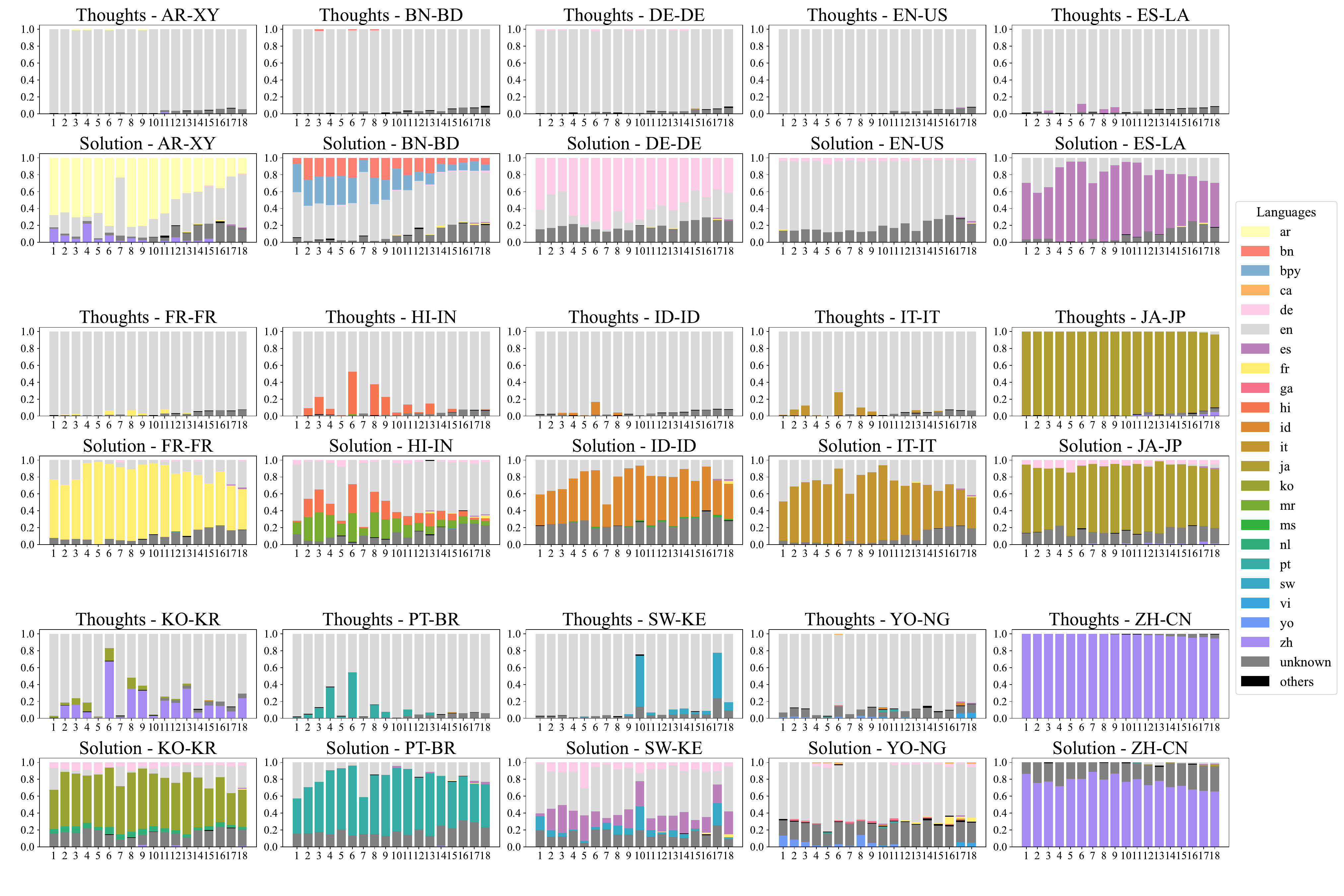}
    \caption{Language composition for \texttt{R1-14B} (m-MMLU).}
    \label{fig:r1_14b_mmlu}
\end{figure*}

\begin{figure*}[h]
    \centering
    \includegraphics[width=\textwidth]{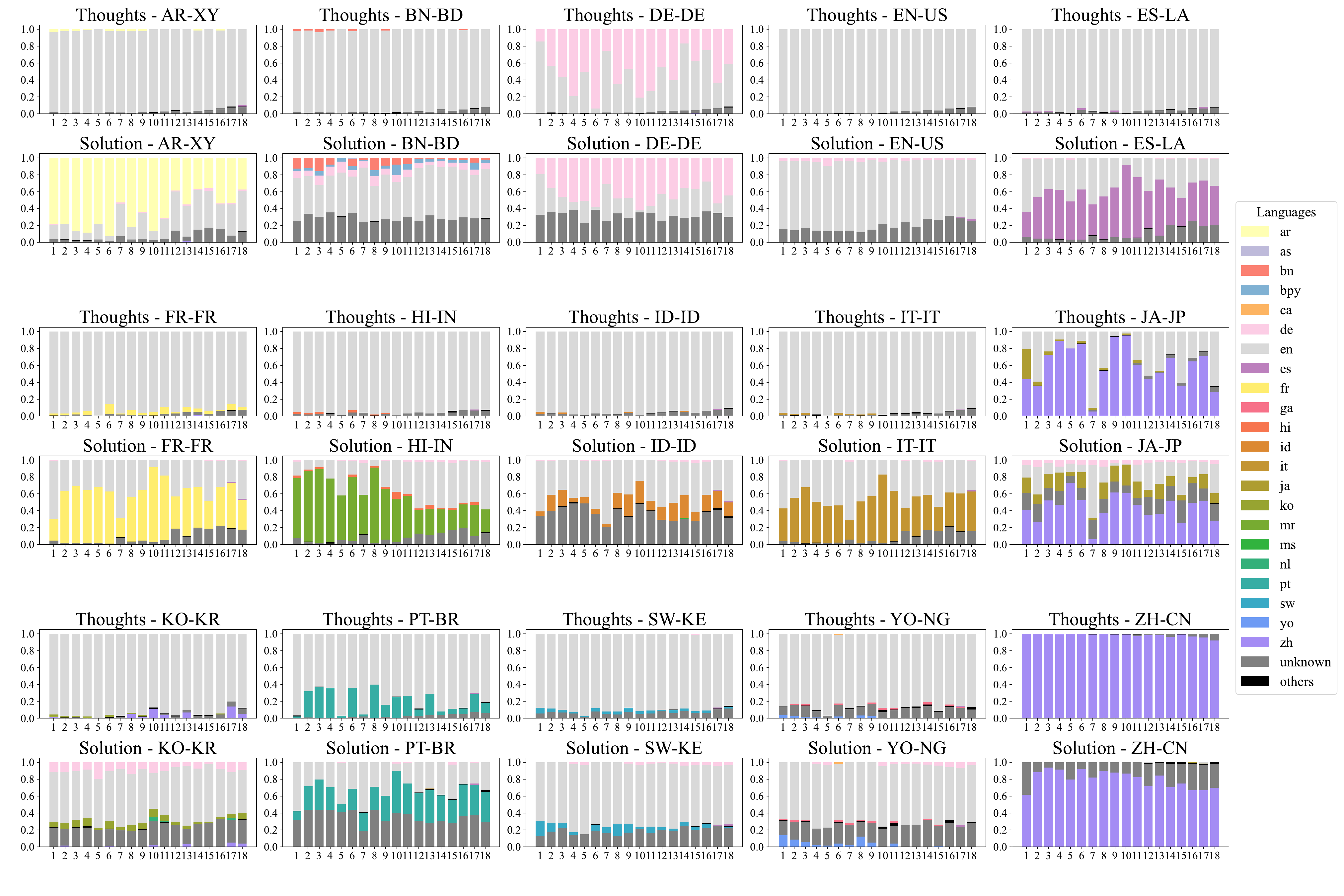}
    \caption{Language composition for \texttt{R1-7B} (m-MMLU).}
    \label{fig:r1_7b_mmlu}
\end{figure*}

\begin{figure*}[h]
    \centering
    \includegraphics[width=\textwidth]{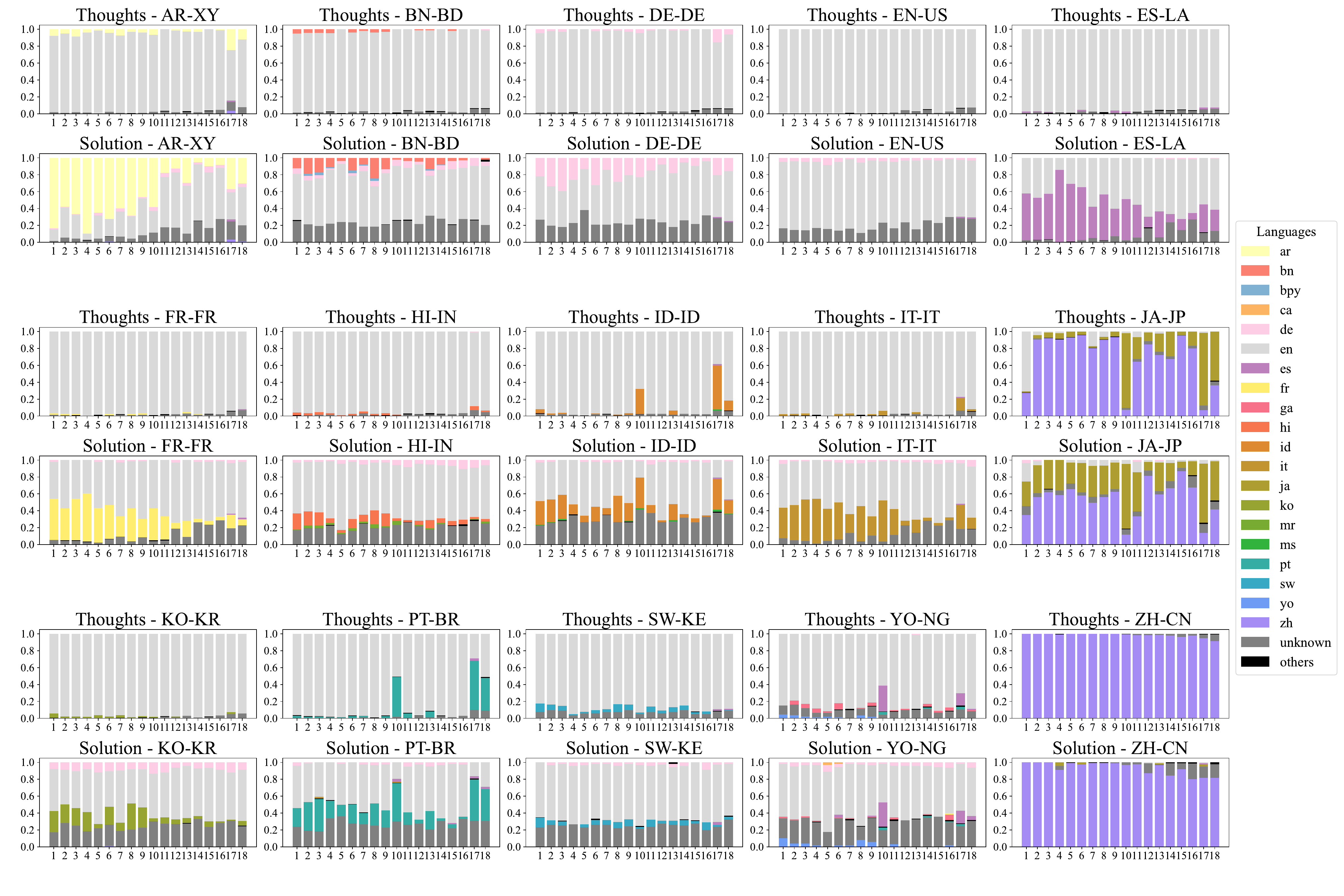}
    \caption{Language composition for \texttt{R1-1.5B} (m-MMLU).}
    \label{fig:r1_1_5b_mmlu}
\end{figure*}

\begin{figure*}[h]
    \centering
    \includegraphics[width=\textwidth]{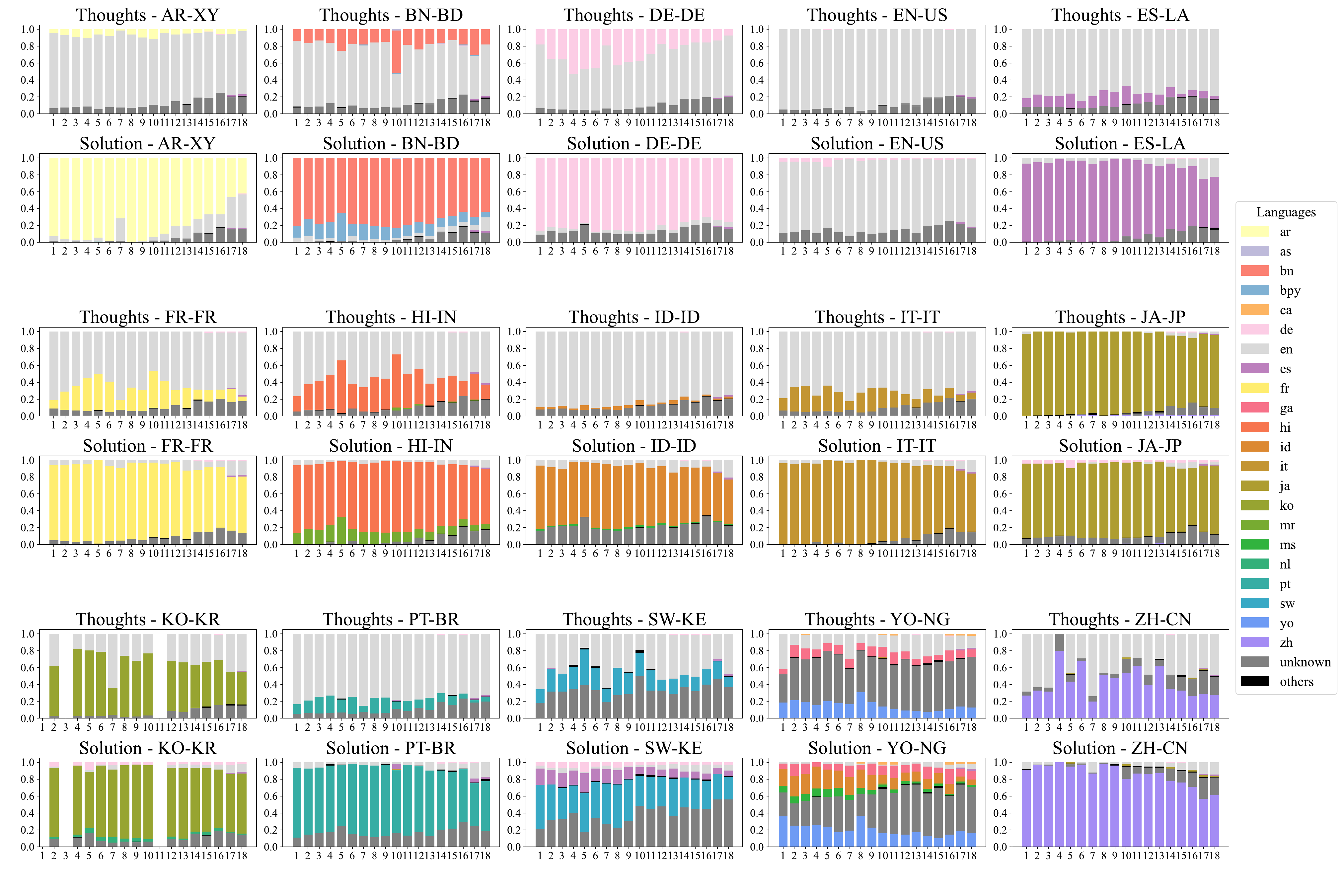}
    \caption{Language composition for \texttt{Gemini} (m-MMLU).}
    \label{fig:gemini_mmlu}
\end{figure*}

\begin{figure*}[h]
    \centering
    \includegraphics[width=\textwidth]{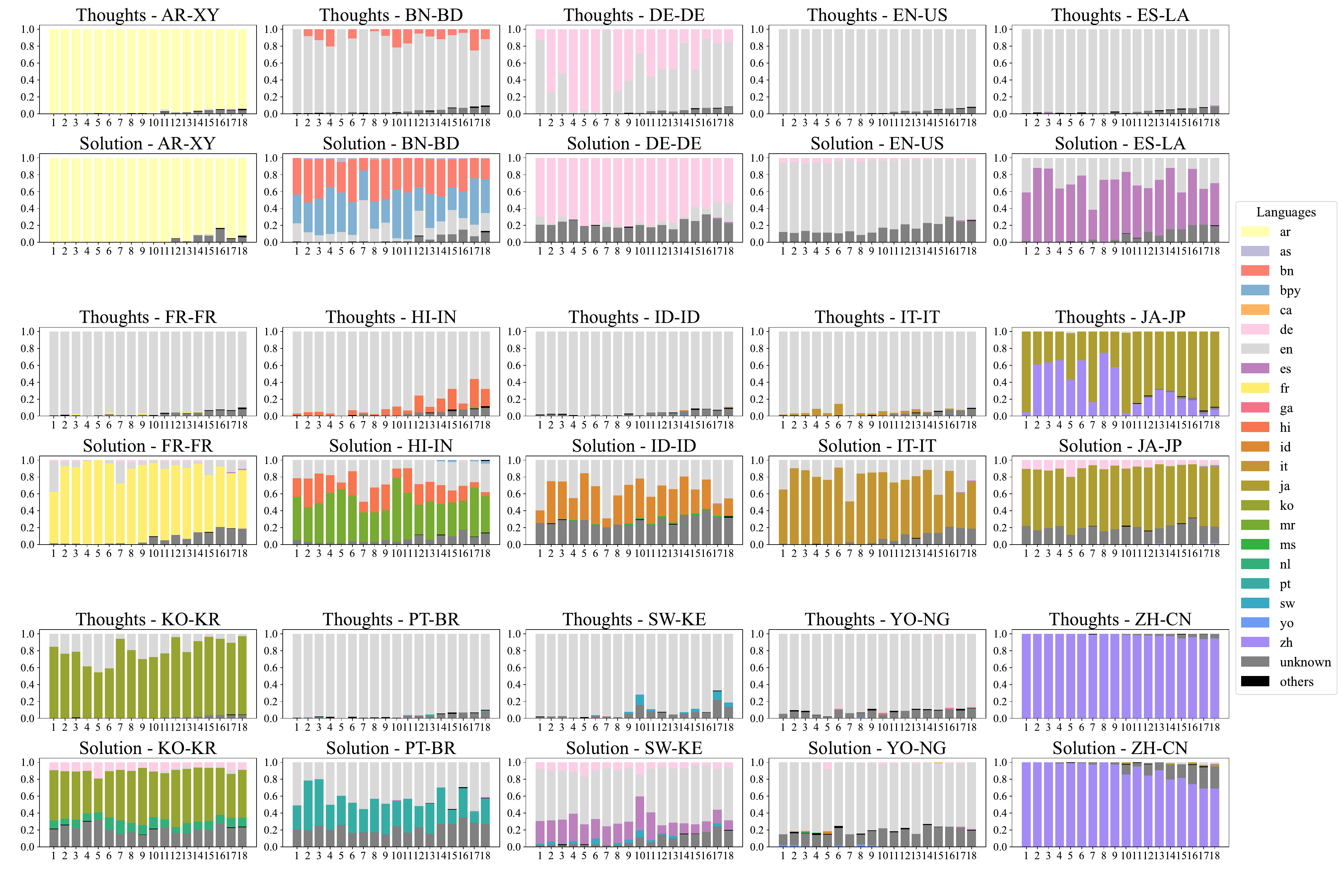}
    \caption{Language composition for \texttt{QwQ-32B} (m-MMLU).}
    \label{fig:qwq_32b_mmlu}
\end{figure*}

\begin{figure*}[h]
    \centering
    \includegraphics[width=\textwidth]{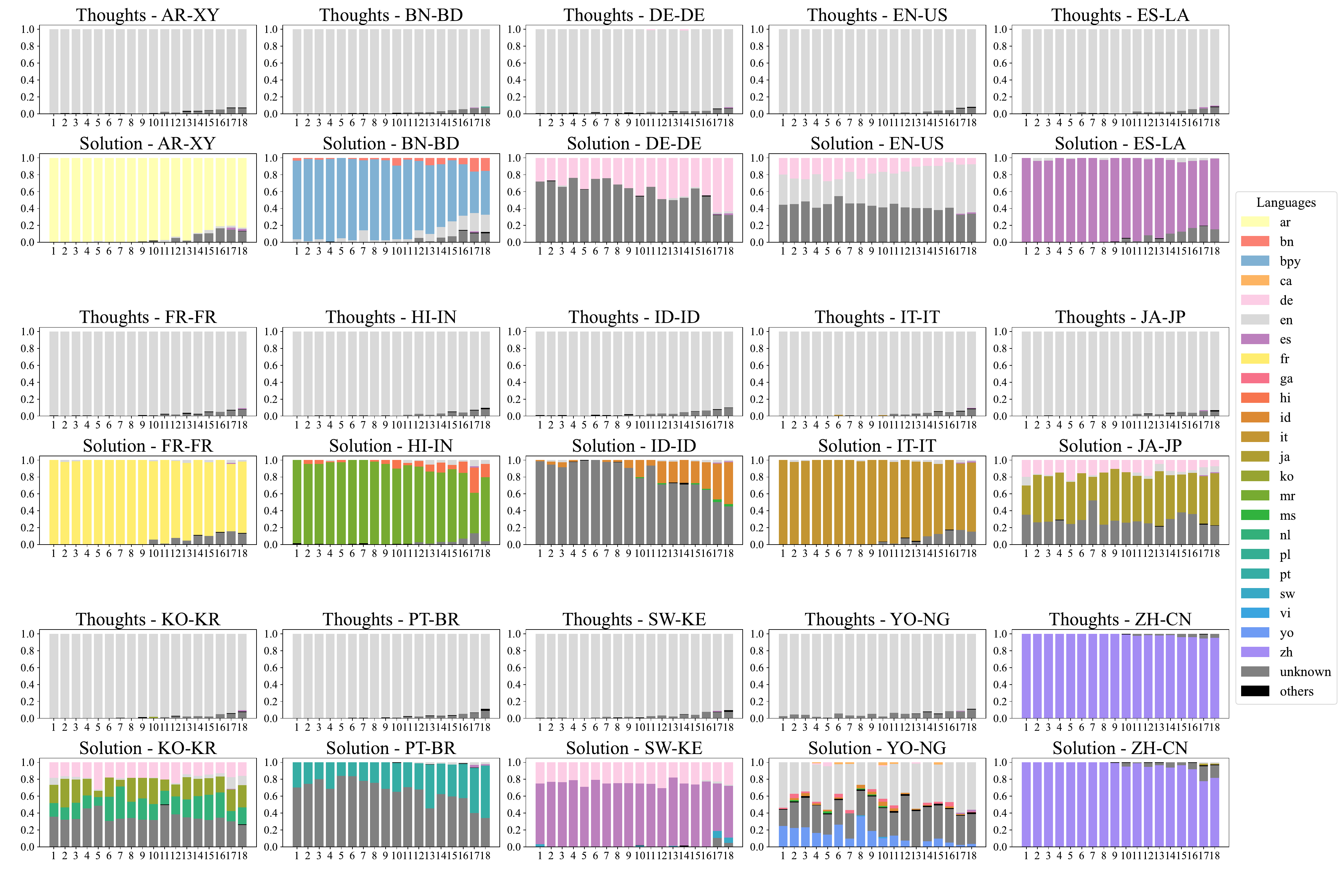}
    \caption{Language composition for \texttt{Qwen3-32B} (m-MMLU).}
    \label{fig:qwen3_32b_mmlu}
\end{figure*}

\begin{figure*}[h]
    \centering
    \includegraphics[width=\textwidth]{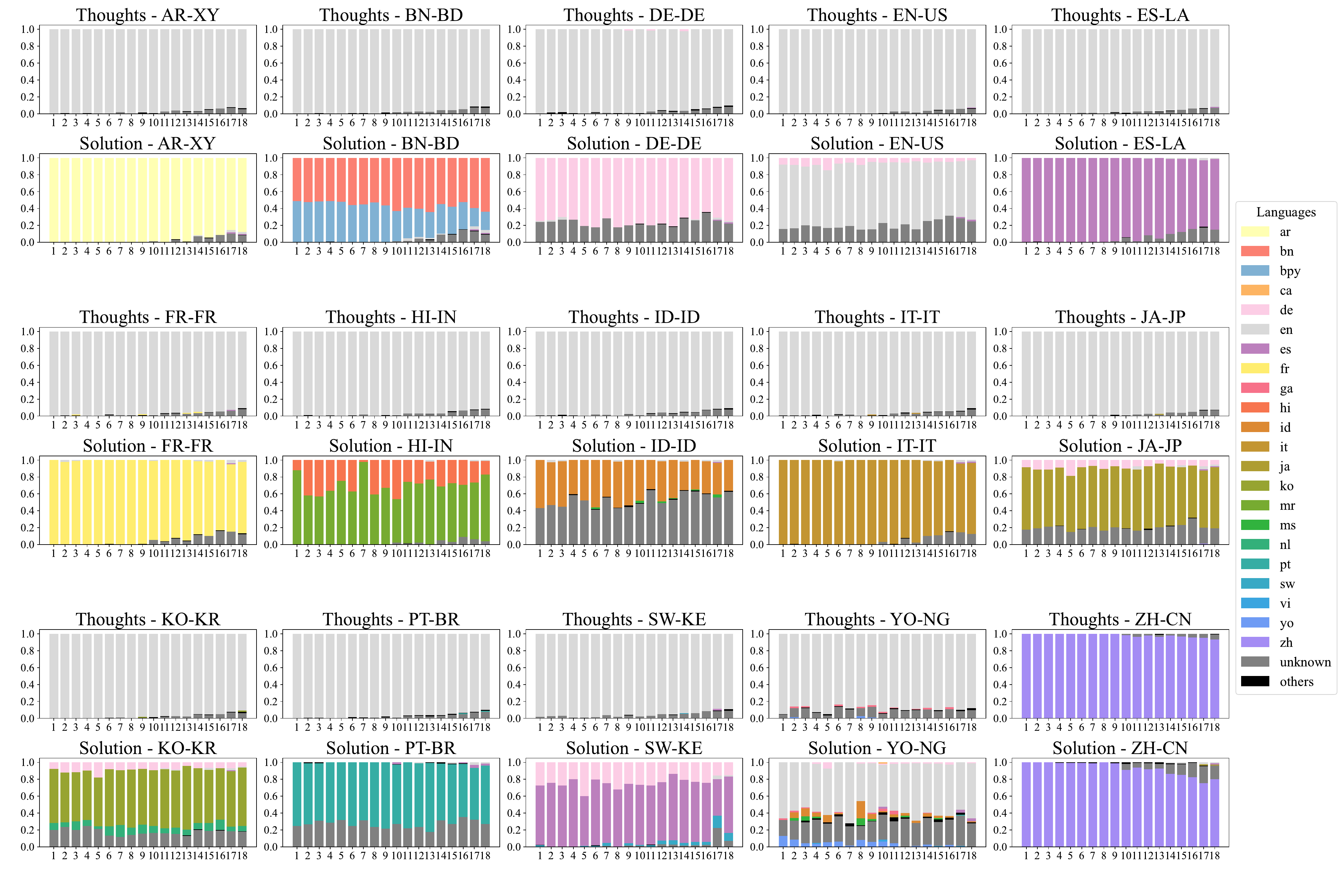}
    \caption{Language composition for \texttt{Qwen3-30B-A3B} (m-MMLU).}
    \label{fig:qwen3_30b_a3b_mmlu}
\end{figure*}

\begin{figure*}[h]
    \centering
    \includegraphics[width=\textwidth]{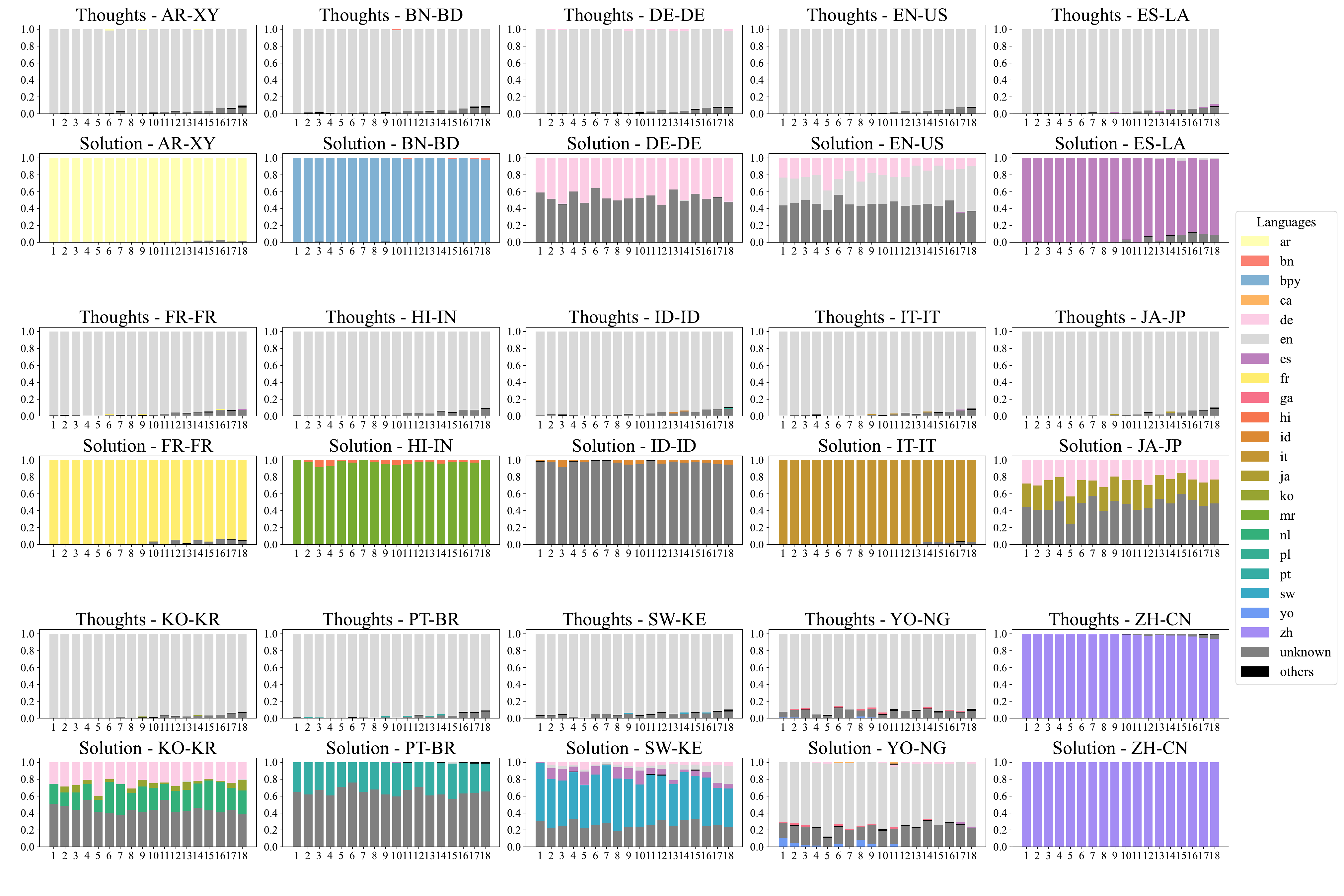}
    \caption{Language composition for \texttt{Qwen3-4B} (m-MMLU).}
    \label{fig:qwen3_4b_mmlu}
\end{figure*}

\paragraph{Language composition visualization.}
\label{sec:appendix-composition-vis}
In \Cref{fig:r1_kk,fig:r1_70b_kk,fig:r1_8b_kk,fig:r1_32b_kk,fig:r1_14b_kk,fig:r1_7b_kk,fig:r1_1_5b_kk,fig:gemini_kk,fig:qwq_32b_kk,fig:qwen3_32b_kk,fig:qwen3_30b_a3b_kk,fig:qwen3_4b_kk} and \Cref{fig:r1_70b_mmlu,fig:r1_8b_mmlu,fig:r1_32b_mmlu,fig:r1_14b_mmlu,fig:r1_7b_mmlu,fig:r1_1_5b_mmlu,fig:gemini_mmlu,fig:qwq_32b_mmlu,fig:qwen3_32b_mmlu,fig:qwen3_30b_a3b_mmlu,fig:qwen3_4b_mmlu}, we visualize the language composition of reasoning traces and final answers across models on the K\&K and m-MMLU datasets, respectively. These figures provide a detailed view of how different RLMs vary in their use of languages during the reasoning process across inputs. Missing bars in some plots correspond to rare failure cases where the model enters an endless reasoning loop and fails to produce a final answer. For more details on these invalid generations, see Appendix~\ref{sec:appendix-model-performance}.

\begin{table*}[h]
  \centering
  \footnotesize
  \scalebox{0.9}{
    \begin{tabular}{lcccccccc}
    \toprule
          & \multirow{2}[2]{*}{\textbf{No control}} & \multicolumn{3}{c}{\textbf{Single-script control}} & \multicolumn{4}{c}{\textbf{Multi-script control}} \\
          &       & Input script & Latin & Han & Latin+Han & Input+Latin & Input+Han & Input+Latin+Han \\
    \midrule
    \textit{\textbf{Arabic}} &       &       &       &       &       &       &       &  \\
    R1-70B & 0.34  & \celldarkred{0.26}  & \celldarkgreen{0.55}  & \cellequal{0.34}  & \celllightgreen{0.35}  & \cellmediumred{0.27}  & \cellmediumred{0.31}  & 0.33 \\
    R1-32B & 0.35  & \celllightgreen{0.37}  & \cellmediumgreen{0.49}  & \cellmediumgreen{0.45}  & \celldarkgreen{0.52}  & \celldarkgreen{0.68}  & \celldarkgreen{0.81}  & \celldarkgreen{0.66} \\
    R1-14B & 0.25  & \celllightred{0.23}  & \cellmediumgreen{0.32}  & \celldarkgreen{0.43}  & \cellmediumgreen{0.33}  & \cellmediumgreen{0.32}  & \celldarkgreen{0.38}  & \cellmediumgreen{0.32} \\
    AVG   & 0.31  & \celllightred{0.29}  & \cellmediumgreen{0.45}  & \cellmediumgreen{0.41}  & \cellmediumgreen{0.40}  & \cellmediumgreen{0.42}  & \celldarkgreen{0.50}  & \cellmediumgreen{0.44} \\
    \midrule
    \textit{\textbf{Hindi}} &       &       &       &       &       &       &       &  \\
    R1-70B & 0.33  & \celldarkred{0.03}  & \celldarkgreen{0.71}  & \celldarkgreen{0.49}  & \celldarkgreen{0.66}  & \celldarkgreen{0.68}  & \celldarkgreen{0.54}  & \celldarkgreen{0.66} \\
    R1-32B & 0.39  & \celldarkred{0.05}  & \celldarkgreen{0.60}  & \celldarkred{0.28}  & \celldarkgreen{0.58}  & \cellmediumred{0.32}  & \cellmediumred{0.29}  & \cellmediumred{0.32} \\
    R1-14B & 0.32  & \celldarkred{0.06}  & \cellmediumgreen{0.46}  & \cellmediumgreen{0.41}  & \celldarkgreen{0.48}  & \cellmediumgreen{0.47}  & \cellmediumgreen{0.44}  & \cellmediumgreen{0.47} \\
    AVG   & 0.35  & \celldarkred{0.05}  & \celldarkgreen{0.59}  & \celllightgreen{0.39}  & \celldarkgreen{0.57}  & \cellmediumgreen{0.49}  & \cellmediumgreen{0.42}  & \cellmediumgreen{0.48} \\
    \midrule
    \textit{\textbf{Japanese}} &       &       &       &       &       &       &       &  \\
    R1-70B & 0.31  & \celllightgreen{0.33}  & \celldarkgreen{0.64}  & \celldarkgreen{0.49}  & \celldarkgreen{0.65}  & \celllightred{0.30}  & \celldash  & \celldash \\
    R1-32B & 0.41  & \celllightgreen{0.42}  & \cellmediumgreen{0.56}  & \celldarkred{0.25}  & \celllightgreen{0.47}  & \cellmediumgreen{0.50}  & \celldash  & \celldash \\
    R1-14B & 0.25  & \celllightred{0.23}  & \celldarkgreen{0.38}  & \celldarkred{0.18}  & \celllightred{0.24}  & \cellequal{0.25}  & \celldash  & \celldash \\
    AVG   & 0.32  & \celllightgreen{0.33}  & \celldarkgreen{0.53}  & 0.31  & \cellmediumgreen{0.45}  & \celllightgreen{0.35}  & \celldash  & \celldash \\
    \bottomrule
    \end{tabular}%
    }
  \caption{Performance comparison across different script control strategies for Arabic, Hindi, and Japanese.\protect\footnotemark}
  \label{tab:script-control-all-1}%
\end{table*}%
\begin{table*}[h]
  \centering
  \footnotesize
    \begin{tabular}{lcccc}
    \toprule
          & \multirow{2}[2]{*}{\textbf{No control}} & \multicolumn{2}{c}{\textbf{Single-script control}} & \textbf{Multi-script control} \\
          &       & Latin & Han & Latin+Han \\
    \midrule
    \textit{\textbf{English}} &       &       &       &  \\
    R1-70B & 0.81  & 0.79  & \celldarkred{0.62}  & 0.79 \\
    R1-32B & 0.86  & 0.85  & \celldarkred{0.46}  & \cellequal{0.86} \\
    R1-14B & 0.81  & \cellequal{0.81}  & \celldarkred{0.64}  & 0.80 \\
    AVG   & 0.83  & 0.82  & \celldarkred{0.57}  & 0.82 \\
    \midrule
    \textit{\textbf{French}} &       &       &       &  \\
    R1-70B & 0.64  & \celllightgreen{0.65}  & \cellmediumred{0.59}  & \celllightgreen{0.65} \\
    R1-32B & 0.75  & \cellequal{0.75}  & \celldarkred{0.41}  & \celllightgreen{0.76} \\
    R1-14B & 0.72  & \cellequal{0.72}  & \celldarkred{0.53}  & \cellequal{0.72} \\
    AVG   & 0.70  & \celllightgreen{0.71}  & \celldarkred{0.51}  & \celllightgreen{0.71} \\
    \midrule
    \textit{\textbf{Chinese}} &       &       &       &  \\
    R1-70B & 0.67  & \celldarkred{0.55}  & \celllightgreen{0.69}  & \cellequal{0.67} \\
    R1-32B & 0.70  & \cellmediumred{0.61}  & \celllightred{0.67}  & \cellequal{0.70} \\
    R1-14B & 0.61  & \cellmediumred{0.52}  & 0.59  & 0.60 \\
    AVG   & 0.66  & \cellmediumred{0.56}  & 0.65  & \cellequal{0.66} \\
    \bottomrule
    \end{tabular}%
  \caption{Performance comparison across different script control strategies for English, French, and Chinese.}
  \label{tab:script-control-all-2}%
\end{table*}%

\subsubsection{Script-Controlled Generation}
\footnotetext{For Japanese, entries under “Input+Han” and “Input+Latin+Han” are omitted because Japanese inherently uses Han characters. Thus, “Input+Han” is equivalent to “Input”, and “Input+Latin+Han” is equivalent to “Input+Latin” in this case.}

\paragraph{Script control complete results.}
\label{sec:appendix-script-control-res}
Tables~\ref{tab:script-control-all-1} and~\ref{tab:script-control-all-2} present the detailed performance results of three DeepSeek-R1 models (\texttt{R1-70B}, \texttt{R1-32B}, and \texttt{R1-14B}) across various script control strategies for languages from the K\&K dataset. We compare model accuracy under three settings: no control (default decoding), single-script control (forcing reasoning in one script), and multi-script control (allowing reasoning in a combination of scripts). The languages are grouped into two sets based on their writing scripts: Arabic, Hindi, and Japanese use non-Latin/Han scripts; English, French, and Chinese use Latin or Han scripts. As shown, non-Latin/Han languages benefit substantially from being forced to reason in Latin or Han scripts, whereas native-script reasoning is optimal for Latin and Han-script languages. These findings highlight the importance of aligning reasoning scripts with the model’s internal preferences for achieving optimal multilingual reasoning performance.

\paragraph{Script control generation examples.}
\label{sec:appendix-script-control-examples}
We demonstrate the effects of script control on multilingual reasoning through examples using the same Knights and Knaves puzzle with Arabic input (\Cref{fig:script-control-exp-free,fig:script-control-exp-latin,fig:script-control-exp-han,fig:script-control-exp-arabic}). Figure~\ref{fig:script-control-exp-free} shows unconstrained reasoning with natural language mixing across English, Chinese, and Arabic. Figures~\ref{fig:script-control-exp-latin}, \ref{fig:script-control-exp-han}, and \ref{fig:script-control-exp-arabic} demonstrate controlled reasoning under Latin, Han, and Arabic script constraints respectively. These examples illustrate how script control can effectively guide language selection while maintaining the model's cross-lingual reasoning capabilities.

\subsubsection{Model Performance Details}
\label{sec:appendix-model-performance}
Here, we report the valid reasoning rate and accuracy across all reasoning models used in this work in \Cref{tab:acc-r1,tab:acc-r1-70b,tab:acc-r1-8b,tab:acc-r1-32b,tab:acc-r1-14b,tab:acc-r1-7b,tab:acc-r1-1.5b,tab:acc-gemini,tab:acc-qwq,tab:acc-q3-32b,tab:acc-q3-30b,tab:acc-q3-4b}. The valid reasoning rate refers to the proportion of generations that produce a complete reasoning trace followed by a final answer, as defined in Section~\ref{sec:exp-setup-models}. The accuracy is computed over all generations, both valid and invalid, and reflects the overall correctness of the final answers.

\begin{table*}[h]
  \centering
  \footnotesize
  \scalebox{0.85}{
    \begin{tabular}{c|cc|cc|cc|cc|cc|cc|cc}
    \toprule
    \textbf{\texttt{R1}} & \multicolumn{2}{c|}{\textbf{2ppl}} & \multicolumn{2}{c|}{\textbf{3ppl}} & \multicolumn{2}{c|}{\textbf{4ppl}} & \multicolumn{2}{c|}{\textbf{5ppl}} & \multicolumn{2}{c|}{\textbf{6ppl}} & \multicolumn{2}{c|}{\textbf{7ppl}} & \multicolumn{2}{c}{\textbf{8ppl}} \\
          & acc\%   & valid\% & acc\%   & valid\% & acc\%   & valid\% & acc\%   & valid\% & acc\%   & valid\% & acc\%   & valid\% & acc\%   & valid\% \\
    \midrule
    ar    & 96    & 100   & 100   & 100   & 89    & 100   & 92    & 100   & 92    & 100   & 91    & 100   & 92    & 100 \\
    en    & 100   & 100   & 100   & 100   & 98    & 100   & 96    & 100   & 96    & 100   & 98    & 100   & 96    & 100 \\
    fr    & 96    & 100   & 97    & 100   & 92    & 100   & 89    & 100   & 89    & 100   & 89    & 100   & 89    & 100 \\
    ja    & 98    & 100   & 94    & 100   & 86    & 100   & 89    & 100   & 89    & 100   & 81    & 100   & 89    & 100 \\
    zh    & 99    & 100   & 99    & 100   & 97    & 100   & 97    & 100   & 97    & 100   & 95    & 100   & 97    & 100 \\
    hi    & 100   & 100   & 99    & 100   & 91    & 100   & 92    & 100   & 92    & 100   & 88    & 100   & 92    & 100 \\
    AVG   & 98    & 100   & 98    & 100   & 92    & 100   & 93    & 100   & 93    & 100   & 90    & 100   & 93    & 100 \\
    \bottomrule
    \end{tabular}%
    }
  \caption{Accuracy and valid reasoning rate (\%) of \texttt{DeepSeek-R1} on the K\&K dataset across difficulty levels (2ppl to 8ppl) and input languages.}
  \label{tab:acc-r1}%
\end{table*}%

\begin{table*}[h]
  \centering
  \footnotesize
  \scalebox{0.85}{
    \begin{tabular}{c|cc|cc|cc|cc|cc|cc|cc}
    \toprule
    \textbf{\texttt{R1-70B}} & \multicolumn{2}{c|}{\textbf{2ppl}} & \multicolumn{2}{c|}{\textbf{3ppl}} & \multicolumn{2}{c|}{\textbf{4ppl}} & \multicolumn{2}{c|}{\textbf{5ppl}} & \multicolumn{2}{c|}{\textbf{6ppl}} & \multicolumn{2}{c|}{\textbf{7ppl}} & \multicolumn{2}{c}{\textbf{8ppl}} \\
          & acc\%   & valid\% & acc\%   & valid\% & acc\%   & valid\% & acc\%   & valid\% & acc\%   & valid\% & acc\%   & valid\% & acc\%   & valid\% \\
    \midrule
    ar    & 59    & 96    & 54    & 90    & 37    & 89    & 26    & 85    & 28    & 83    & 17    & 82    & 16    & 86 \\
    en    & 92    & 99    & 93    & 99    & 87    & 94    & 78    & 92    & 78    & 96    & 74    & 92    & 63    & 93 \\
    fr    & 76    & 92    & 72    & 87    & 67    & 84    & 61    & 85    & 61    & 87    & 59    & 91    & 52    & 88 \\
    ja    & 57    & 97    & 42    & 92    & 32    & 90    & 29    & 89    & 21    & 86    & 20    & 86    & 15    & 87 \\
    zh    & 88    & 100   & 81    & 99    & 75    & 99    & 64    & 97    & 66    & 97    & 50    & 91    & 44    & 99 \\
    hi    & 44    & 78    & 40    & 77    & 37    & 73    & 28    & 73    & 28    & 68    & 30    & 77    & 26    & 73 \\
    AVG   & 69    & 94    & 63    & 91    & 56    & 88    & 47    & 87    & 47    & 86    & 42    & 86    & 36    & 87 \\
    \bottomrule
    \end{tabular}%
    }
  \caption{Accuracy and valid reasoning rate (\%) of \texttt{DeepSeek-R1-Distill-Llama-70B} on the K\&K dataset across difficulty levels (2ppl to 8ppl) and input languages.}
  \label{tab:acc-r1-70b}%
\end{table*}%

\begin{table*}[h]
  \centering
  \footnotesize
  \scalebox{0.85}{
    \begin{tabular}{c|cc|cc|cc|cc|cc|cc|cc}
    \toprule
    \textbf{\texttt{R1-8B}} & \multicolumn{2}{c|}{\textbf{2ppl}} & \multicolumn{2}{c|}{\textbf{3ppl}} & \multicolumn{2}{c|}{\textbf{4ppl}} & \multicolumn{2}{c|}{\textbf{5ppl}} & \multicolumn{2}{c|}{\textbf{6ppl}} & \multicolumn{2}{c|}{\textbf{7ppl}} & \multicolumn{2}{c}{\textbf{8ppl}} \\
          & acc\%   & valid\% & acc\%   & valid\% & acc\%   & valid\% & acc\%   & valid\% & acc\%   & valid\% & acc\%   & valid\% & acc\%   & valid\% \\
    \midrule
    ar    & 10    & 99    & 5     & 93    & 3     & 94    & 3     & 96    & 1     & 92    & 1     & 93    & 0     & 89 \\
    en    & 68    & 100   & 66    & 100   & 56    & 99    & 42    & 99    & 38    & 99    & 35    & 100   & 20    & 98 \\
    fr    & 17    & 100   & 20    & 99    & 15    & 100   & 9     & 99    & 6     & 99    & 6     & 99    & 3     & 98 \\
    ja    & 14    & 100   & 9     & 98    & 7     & 98    & 1     & 99    & 0     & 98    & 1     & 96    & 1     & 99 \\
    zh    & 55    & 98    & 44    & 99    & 41    & 98    & 30    & 100   & 22    & 99    & 18    & 99    & 13    & 99 \\
    hi    & 24    & 99    & 23    & 99    & 11    & 100   & 8     & 99    & 6     & 98    & 2     & 98    & 2     & 97 \\
    AVG   & 31    & 99    & 28    & 98    & 22    & 98    & 15    & 99    & 12    & 98    & 10    & 98    & 6     & 97 \\
    \bottomrule
    \end{tabular}%
    }
  \caption{Accuracy and valid reasoning rate (\%) of \texttt{DeepSeek-R1-Distill-Llama-8B} on the K\&K dataset across difficulty levels (2ppl to 8ppl) and input languages.}
  \label{tab:acc-r1-8b}%
\end{table*}%

\begin{table*}[h]
  \centering
  \footnotesize
  \scalebox{0.85}{
    \begin{tabular}{c|cc|cc|cc|cc|cc|cc|cc}
    \toprule
    \textbf{\texttt{R1-32B}} & \multicolumn{2}{c|}{\textbf{2ppl}} & \multicolumn{2}{c|}{\textbf{3ppl}} & \multicolumn{2}{c|}{\textbf{4ppl}} & \multicolumn{2}{c|}{\textbf{5ppl}} & \multicolumn{2}{c|}{\textbf{6ppl}} & \multicolumn{2}{c|}{\textbf{7ppl}} & \multicolumn{2}{c}{\textbf{8ppl}} \\
          & acc\%   & valid\% & acc\%   & valid\% & acc\%   & valid\% & acc\%   & valid\% & acc\%   & valid\% & acc\%   & valid\% & acc\%   & valid\% \\
    \midrule
    ar    & 45    & 100   & 41    & 99    & 43    & 100   & 32    & 99    & 29    & 99    & 31    & 99    & 25    & 99 \\
    en    & 95    & 100   & 94    & 100   & 92    & 100   & 88    & 100   & 88    & 99    & 79    & 100   & 70    & 99 \\
    fr    & 81    & 100   & 84    & 100   & 82    & 100   & 76    & 100   & 75    & 100   & 67    & 100   & 60    & 100 \\
    ja    & 59    & 89    & 50    & 83    & 50    & 88    & 43    & 90    & 36    & 90    & 28    & 90    & 19    & 81 \\
    zh    & 87    & 97    & 78    & 93    & 78    & 93    & 71    & 96    & 68    & 96    & 59    & 99    & 52    & 95 \\
    hi    & 53    & 99    & 49    & 99    & 44    & 100   & 35    & 100   & 39    & 100   & 26    & 99    & 27    & 99 \\
    AVG   & 70    & 98    & 66    & 96    & 65    & 97    & 57    & 98    & 55    & 98    & 48    & 98    & 42    & 96 \\

    \bottomrule
    \end{tabular}%
    }
  \caption{Accuracy and valid reasoning rate (\%) of \texttt{DeepSeek-R1-Distill-Qwen-32B} on the K\&K dataset across difficulty levels (2ppl to 8ppl) and input languages.}
  \label{tab:acc-r1-32b}%
\end{table*}%

\begin{table*}[h]
  \centering
  \footnotesize
  \scalebox{0.85}{
    \begin{tabular}{c|cc|cc|cc|cc|cc|cc|cc}
    \toprule
    \textbf{\texttt{R1-14B}} & \multicolumn{2}{c|}{\textbf{2ppl}} & \multicolumn{2}{c|}{\textbf{3ppl}} & \multicolumn{2}{c|}{\textbf{4ppl}} & \multicolumn{2}{c|}{\textbf{5ppl}} & \multicolumn{2}{c|}{\textbf{6ppl}} & \multicolumn{2}{c|}{\textbf{7ppl}} & \multicolumn{2}{c}{\textbf{8ppl}} \\
          & acc\%   & valid\% & acc\%   & valid\% & acc\%   & valid\% & acc\%   & valid\% & acc\%   & valid\% & acc\%   & valid\% & acc\%   & valid\% \\
    \midrule
    ar    & 40    & 100   & 37    & 100   & 34    & 100   & 23    & 100   & 17    & 100   & 16    & 100   & 11    & 99 \\
    en    & 95    & 100   & 92    & 100   & 87    & 100   & 81    & 100   & 77    & 99    & 72    & 100   & 65    & 100 \\
    fr    & 84    & 100   & 77    & 100   & 83    & 100   & 72    & 99    & 70    & 99    & 62    & 99    & 55    & 99 \\
    ja    & 45    & 96    & 40    & 95    & 31    & 96    & 21    & 90    & 19    & 90    & 12    & 89    & 7     & 87 \\
    zh    & 83    & 97    & 78    & 95    & 65    & 94    & 60    & 93    & 56    & 95    & 48    & 93    & 41    & 93 \\
    hi    & 49    & 100   & 47    & 100   & 38    & 99    & 29    & 98    & 22    & 98    & 25    & 98    & 18    & 95 \\
    AVG   & 66    & 99    & 62    & 98    & 56    & 98    & 48    & 97    & 43    & 97    & 39    & 97    & 33    & 96 \\

    \bottomrule
    \end{tabular}%
    }
  \caption{Accuracy and valid reasoning rate (\%) of \texttt{DeepSeek-R1-Distill-Qwen-14B} on the K\&K dataset across difficulty levels (2ppl to 8ppl) and input languages.}
  \label{tab:acc-r1-14b}%
\end{table*}%

\begin{table*}[h]
  \centering
  \footnotesize
  \scalebox{0.85}{
    \begin{tabular}{c|cc|cc|cc|cc|cc|cc|cc}
    \toprule
    \textbf{\texttt{R1-7B}} & \multicolumn{2}{c|}{\textbf{2ppl}} & \multicolumn{2}{c|}{\textbf{3ppl}} & \multicolumn{2}{c|}{\textbf{4ppl}} & \multicolumn{2}{c|}{\textbf{5ppl}} & \multicolumn{2}{c|}{\textbf{6ppl}} & \multicolumn{2}{c|}{\textbf{7ppl}} & \multicolumn{2}{c}{\textbf{8ppl}} \\
          & acc\%   & valid\% & acc\%   & valid\% & acc\%   & valid\% & acc\%   & valid\% & acc\%   & valid\% & acc\%   & valid\% & acc\%   & valid\% \\
    \midrule
    ar    & 9     & 90    & 4     & 84    & 2     & 83    & 1     & 76    & 0     & 74    & 0     & 76    & 0     & 77 \\
    en    & 90    & 99    & 80    & 97    & 66    & 96    & 53    & 93    & 43    & 91    & 31    & 85    & 26    & 82 \\
    fr    & 40    & 98    & 23    & 98    & 14    & 98    & 10    & 99    & 3     & 97    & 4     & 97    & 2     & 95 \\
    ja    & 4     & 82    & 2     & 80    & 0     & 85    & 0     & 82    & 0     & 78    & 0     & 78    & 0     & 80 \\
    zh    & 61    & 98    & 52    & 97    & 45    & 95    & 40    & 97    & 26    & 95    & 21    & 95    & 10    & 97 \\
    hi    & 19    & 97    & 10    & 92    & 6     & 85    & 3     & 81    & 1     & 84    & 1     & 85    & 1     & 78 \\
    AVG   & 37    & 94    & 28    & 92    & 22    & 90    & 18    & 88    & 12    & 87    & 9     & 86    & 6     & 85 \\

    \bottomrule
    \end{tabular}%
    }
  \caption{Accuracy and valid reasoning rate (\%) of \texttt{DeepSeek-R1-Distill-Qwen-7B} on the K\&K dataset across difficulty levels (2ppl to 8ppl) and input languages.}
  \label{tab:acc-r1-7b}%
\end{table*}%

\begin{table*}[h]
  \centering
  \footnotesize
  \scalebox{0.85}{
    \begin{tabular}{c|cc|cc|cc|cc|cc|cc|cc}
    \toprule
    \textbf{\texttt{R1-1.5B}} & \multicolumn{2}{c|}{\textbf{2ppl}} & \multicolumn{2}{c|}{\textbf{3ppl}} & \multicolumn{2}{c|}{\textbf{4ppl}} & \multicolumn{2}{c|}{\textbf{5ppl}} & \multicolumn{2}{c|}{\textbf{6ppl}} & \multicolumn{2}{c|}{\textbf{7ppl}} & \multicolumn{2}{c}{\textbf{8ppl}} \\
          & acc\%   & valid\% & acc\%   & valid\% & acc\%   & valid\% & acc\%   & valid\% & acc\%   & valid\% & acc\%   & valid\% & acc\%   & valid\% \\
    \midrule
    ar    & 3     & 89    & 1     & 86    & 0     & 80    & 1     & 79    & 0     & 74    & 0     & 66    & 0     & 63 \\
    en    & 56    & 98    & 34    & 96    & 12    & 92    & 11    & 83    & 4     & 83    & 1     & 78    & 0     & 73 \\
    fr    & 27    & 92    & 12    & 92    & 2     & 83    & 1     & 76    & 0     & 75    & 0     & 66    & 0     & 67 \\
    ja    & 1     & 86    & 0     & 78    & 0     & 71    & 0     & 71    & 0     & 64    & 0     & 57    & 0     & 64 \\
    zh    & 21    & 84    & 17    & 75    & 12    & 70    & 7     & 78    & 2     & 76    & 0     & 73    & 0     & 77 \\
    hi    & 6     & 94    & 3     & 91    & 1     & 86    & 1     & 88    & 1     & 83    & 0     & 88    & 0     & 80 \\
    AVG   & 19    & 91    & 11    & 86    & 4     & 81    & 3     & 79    & 1     & 76    & 0     & 72    & 0     & 71 \\

    \bottomrule
    \end{tabular}%
    }
  \caption{Accuracy and valid reasoning rate (\%) of \texttt{DeepSeek-R1-Distill-Qwen-1.5B} on the K\&K dataset across difficulty levels (2ppl to 8ppl) and input languages.}
  \label{tab:acc-r1-1.5b}%
\end{table*}%

\begin{table*}[h]
  \centering
  \footnotesize
  \scalebox{0.85}{
    \begin{tabular}{c|cc|cc|cc|cc|cc|cc|cc}
    \toprule
    \textbf{\texttt{Gemini}} & \multicolumn{2}{c|}{\textbf{2ppl}} & \multicolumn{2}{c|}{\textbf{3ppl}} & \multicolumn{2}{c|}{\textbf{4ppl}} & \multicolumn{2}{c|}{\textbf{5ppl}} & \multicolumn{2}{c|}{\textbf{6ppl}} & \multicolumn{2}{c|}{\textbf{7ppl}} & \multicolumn{2}{c}{\textbf{8ppl}} \\
          & acc\%   & valid\% & acc\%   & valid\% & acc\%   & valid\% & acc\%   & valid\% & acc\%   & valid\% & acc\%   & valid\% & acc\%   & valid\% \\
    \midrule
    ar    & 45    & 59    & 45    & 59    & 38    & 52    & 45    & 50    & 39    & 47    & 49    & 56    & 45    & 52 \\
    en    & 53    & 63    & 60    & 75    & 67    & 76    & 59    & 71    & 69    & 74    & 60    & 62    & 53    & 57 \\
    fr    & 50    & 59    & 54    & 69    & 48    & 59    & 50    & 65    & 40    & 54    & 44    & 54    & 35    & 41 \\
    ja    & 35    & 51    & 44    & 62    & 42    & 61    & 58    & 75    & 45    & 62    & 46    & 64    & 41    & 56 \\
    zh    & 51    & 65    & 49    & 61    & 44    & 62    & 52    & 78    & 36    & 63    & 39    & 56    & 33    & 54 \\
    hi    & 41    & 92    & 31    & 82    & 25    & 85    & 29    & 89    & 36    & 86    & 31    & 83    & 34    & 84 \\
    AVG   & 46    & 65    & 47    & 68    & 44    & 66    & 49    & 71    & 44    & 64    & 45    & 62    & 40    & 57 \\
    \bottomrule
    \end{tabular}%
    }
  \caption{Accuracy and valid reasoning rate (\%) of \texttt{Gemini-Flash-Thinking} on the K\&K dataset across difficulty levels (2ppl to 8ppl) and input languages.}
  \label{tab:acc-gemini}%
\end{table*}%

\begin{table*}[h]
  \centering
  \footnotesize
  \scalebox{0.85}{
    \begin{tabular}{c|cc|cc|cc|cc|cc|cc|cc}
    \toprule
    \textbf{\texttt{QwQ}} & \multicolumn{2}{c|}{\textbf{2ppl}} & \multicolumn{2}{c|}{\textbf{3ppl}} & \multicolumn{2}{c|}{\textbf{4ppl}} & \multicolumn{2}{c|}{\textbf{5ppl}} & \multicolumn{2}{c|}{\textbf{6ppl}} & \multicolumn{2}{c|}{\textbf{7ppl}} & \multicolumn{2}{c}{\textbf{8ppl}} \\
          & acc\%   & valid\% & acc\%   & valid\% & acc\%   & valid\% & acc\%   & valid\% & acc\%   & valid\% & acc\%   & valid\% & acc\%   & valid\% \\
    \midrule
    ar    & 74    & 100   & 68    & 100   & 62    & 99    & 61    & 99    & 56    & 99    & 56    & 100   & 47    & 99 \\
    en    & 99    & 100   & 97    & 100   & 90    & 98    & 93    & 100   & 87    & 99    & 86    & 100   & 85    & 99 \\
    fr    & 89    & 99    & 87    & 98    & 88    & 100   & 87    & 99    & 85    & 99    & 83    & 99    & 81    & 98 \\
    ja    & 82    & 100   & 79    & 99    & 74    & 99    & 60    & 99    & 64    & 100   & 57    & 99    & 49    & 99 \\
    zh    & 92    & 100   & 86    & 100   & 80    & 100   & 72    & 100   & 61    & 100   & 60    & 100   & 59    & 100 \\
    hi    & 22    & 100   & 17    & 100   & 16    & 100   & 20    & 100   & 17    & 99    & 18    & 100   & 9     & 100 \\
    AVG   & 76    & 100   & 72    & 100   & 68    & 100   & 65    & 100   & 61    & 100   & 60    & 100   & 55    & 99 \\

    \bottomrule
    \end{tabular}%
    }
  \caption{Accuracy and valid reasoning rate (\%) of \texttt{QwQ-32B} on the K\&K dataset across difficulty levels (2ppl to 8ppl) and input languages.}
  \label{tab:acc-qwq}%
\end{table*}%

\begin{table*}[h]
  \centering
  \footnotesize
  \scalebox{0.85}{
    \begin{tabular}{c|cc|cc|cc|cc|cc|cc|cc}
    \toprule
    \textbf{\texttt{Qwen3-32B}} & \multicolumn{2}{c|}{\textbf{2ppl}} & \multicolumn{2}{c|}{\textbf{3ppl}} & \multicolumn{2}{c|}{\textbf{4ppl}} & \multicolumn{2}{c|}{\textbf{5ppl}} & \multicolumn{2}{c|}{\textbf{6ppl}} & \multicolumn{2}{c|}{\textbf{7ppl}} & \multicolumn{2}{c}{\textbf{8ppl}} \\
          & acc\%   & valid\% & acc\%   & valid\% & acc\%   & valid\% & acc\%   & valid\% & acc\%   & valid\% & acc\%   & valid\% & acc\%   & valid\% \\
    \midrule
    ar    & 95    & 99    & 92    & 97    & 89    & 95    & 86    & 99    & 87    & 98    & 84    & 98    & 81    & 99 \\
    en    & 100   & 100   & 100   & 100   & 100   & 100   & 100   & 100   & 98    & 100   & 99    & 100   & 97    & 99 \\
    fr    & 91    & 100   & 93    & 100   & 92    & 100   & 91    & 100   & 89    & 100   & 91    & 100   & 87    & 100 \\
    ja    & 94    & 99    & 94    & 100   & 92    & 100   & 83    & 100   & 88    & 100   & 87    & 100   & 85    & 100 \\
    zh    & 96    & 100   & 94    & 99    & 96    & 100   & 94    & 100   & 94    & 100   & 93    & 99    & 87    & 100 \\
    hi    & 78    & 95    & 72    & 85    & 7     & 62    & 5     & 59    & 11    & 62    & 5     & 61    & 2     & 80 \\
    AVG   & 92    & 99    & 91    & 97    & 79    & 93    & 76    & 93    & 78    & 93    & 76    & 93    & 73    & 96 \\
    \bottomrule
    \end{tabular}%
    }
  \caption{Accuracy and valid reasoning rate (\%) of \texttt{Qwen3-32B} on the K\&K dataset across difficulty levels (2ppl to 8ppl) and input languages.}
  \label{tab:acc-q3-32b}%
\end{table*}%

\begin{table*}[h]
  \centering
  \footnotesize
  \scalebox{0.85}{
    \begin{tabular}{c|cc|cc|cc|cc|cc|cc|cc}
    \toprule
    \textbf{\texttt{Qwen3-30B-A3B}} & \multicolumn{2}{c|}{\textbf{2ppl}} & \multicolumn{2}{c|}{\textbf{3ppl}} & \multicolumn{2}{c|}{\textbf{4ppl}} & \multicolumn{2}{c|}{\textbf{5ppl}} & \multicolumn{2}{c|}{\textbf{6ppl}} & \multicolumn{2}{c|}{\textbf{7ppl}} & \multicolumn{2}{c}{\textbf{8ppl}} \\
          & acc\%   & valid\% & acc\%   & valid\% & acc\%   & valid\% & acc\%   & valid\% & acc\%   & valid\% & acc\%   & valid\% & acc\%   & valid\% \\
    \midrule
    ar    & 0     & 0     & 0     & 0     & 0     & 100   & 0     & 0     & 0     & 0     & 2     & 2     & 1     & 98 \\
    en    & 99    & 98    & 96    & 95    & 98    & 100   & 97    & 98    & 99    & 99    & 99    & 99    & 97    & 99 \\
    fr    & 68    & 71    & 64    & 65    & 88    & 100   & 84    & 92    & 87    & 95    & 87    & 96    & 84    & 99 \\
    ja    & 0     & 0     & 0     & 0     & 0     & 100   & 0     & 0     & 0     & 0     & 0     & 0     & 0     & 99 \\
    zh    & 55    & 64    & 64    & 67    & 62    & 100   & 76    & 79    & 79    & 84    & 81    & 83    & 77    & 100 \\
    hi    & 0     & 3     & 0     & 4     & 0     & 89    & 0     & 2     & 0     & 4     & 0     & 4     & 0     & 92 \\
    AVG   & 37    & 40    & 37    & 39    & 41    & 98    & 43    & 45    & 44    & 47    & 45    & 48    & 43    & 98 \\
    \bottomrule
    \end{tabular}%
    }
  \caption{Accuracy and valid reasoning rate (\%) of \texttt{Qwen3-30B-A3B} on the K\&K dataset across difficulty levels (2ppl to 8ppl) and input languages.}
  \label{tab:acc-q3-30b}%
\end{table*}%

\begin{table*}[h]
  \centering
  \footnotesize
  \scalebox{0.85}{
    \begin{tabular}{c|cc|cc|cc|cc|cc|cc|cc}
    \toprule
    \textbf{\texttt{Qwen3-4B}} & \multicolumn{2}{c|}{\textbf{2ppl}} & \multicolumn{2}{c|}{\textbf{3ppl}} & \multicolumn{2}{c|}{\textbf{4ppl}} & \multicolumn{2}{c|}{\textbf{5ppl}} & \multicolumn{2}{c|}{\textbf{6ppl}} & \multicolumn{2}{c|}{\textbf{7ppl}} & \multicolumn{2}{c}{\textbf{8ppl}} \\
          & acc\%   & valid\% & acc\%   & valid\% & acc\%   & valid\% & acc\%   & valid\% & acc\%   & valid\% & acc\%   & valid\% & acc\%   & valid\% \\
    \midrule
    ar    & 81    & 100   & 73    & 100   & 59    & 100   & 51    & 99    & 40    & 100   & 37    & 100   & 28    & 100 \\
    en    & 97    & 96    & 96    & 96    & 95    & 94    & 97    & 96    & 95    & 95    & 93    & 93    & 89    & 90 \\
    fr    & 78    & 99    & 83    & 99    & 81    & 100   & 77    & 100   & 79    & 99    & 80    & 100   & 74    & 100 \\
    ja    & 0     & 0     & 0     & 0     & 0     & 0     & 0     & 0     & 0     & 0     & 0     & 0     & 0     & 0 \\
    zh    & 7     & 9     & 13    & 30    & 14    & 31    & 15    & 45    & 14    & 45    & 18    & 45    & 19    & 52 \\
    hi    & 71    & 98    & 46    & 97    & 39    & 97    & 36    & 93    & 31    & 98    & 30    & 100   & 24    & 97 \\
    AVG   & 56    & 67    & 52    & 70    & 48    & 70    & 46    & 72    & 43    & 73    & 43    & 73    & 39    & 73 \\
    \bottomrule
    \end{tabular}%
    }
  \caption{Accuracy and valid reasoning rate (\%) of \texttt{Qwen3-4B} on the K\&K dataset across difficulty levels (2ppl to 8ppl) and input languages.}
  \label{tab:acc-q3-4b}%
\end{table*}%

\end{document}